\documentclass[3p, times, onecolumn, authoryear, review, sort&compress,11pt]{elsarticle}
\usepackage{booktabs}
\usepackage{multirow}
\usepackage[table,xcdraw]{xcolor}

\usepackage{amsmath}  
\allowdisplaybreaks[4]  
\usepackage{amssymb}   

\usepackage{float}
\usepackage{caption}
\usepackage{subfigure}
\usepackage{verbatim}  

\usepackage{lipsum}      
\usepackage{stfloats}      

\usepackage[colorlinks,linkcolor=red]{hyperref}  

\usepackage{longtable}
\usepackage{lscape}     
\usepackage{footnote}   
\usepackage{threeparttable}  
\usepackage{hyperref}

\usepackage{verbatim}  

\usepackage{tikz}
\usepackage{pgfplots}
\usepackage{pgf}
\usetikzlibrary{arrows,snakes,backgrounds}
\usetikzlibrary{calc}
\usetikzlibrary{positioning, arrows.meta}  
\usepackage{graphicx}
\usepackage{color}

\usepackage{setspace}
\usepackage{xcolor}
\usepackage[framemethod=tikz]{mdframed}

\usepackage{algorithm}  
\usepackage{algorithmicx}
\usepackage{algpseudocode}  

\floatname{algorithm}{Algorithm}

\newcommand{\tabincell}[2]{\begin{tabular}{@{}#1@{}}#2\end{tabular}} 

\usepackage{color}
\usepackage{bm}
\usepackage{geometry}
\usepackage{pdflscape}
\usepackage{booktabs}
\usepackage{graphicx}

\usepackage{amssymb}
\usepackage{makecell}
\usepackage{enumitem}
\setenumerate[1]{itemsep=0pt,partopsep=0pt,parsep=\parskip,topsep=0pt}
\setitemize[1]{itemsep=0pt,partopsep=0pt,parsep=\parskip,topsep=0pt}

\journal{European Journal of Operational Research}

\begin{document}
	
	\begin{frontmatter}
		
		\title{Reinforcement learning for multi-item retrieval in the puzzle-based storage system}
		
		\author[label1,label2]{$^{\dagger}$Jing He}
		\author[label3]{$^{\dagger}$Xinglu Liu}
		\author[label3]{Qiyao Duan}
		\author[label3]{Wai Kin Victor Chan}
		\author[label2]{Mingyao Qi\corref{cor1}}
		\cortext[cor1]{Corresponding author}
		\ead{qimy@sz.tsinghua.edu.cn}

		\address[label1]{Department of Industrial Engineering, Tsinghua University, Beijing, 100084, China}
		\address[label2]{Research Center for Modern Logistics, Shenzhen International Graduate School, Tsinghua University, Shenzhen 518055, China} 
		\address[label3]{Intelligent Transportation and Logistics Systems Laboratory, Tsinghua-Berkeley Shenzhen Institute, Tsinghua University, Shenzhen 518055, China}

		\let\thefootnote\relax\footnotetext{${\dagger}$: These authors contributed equally to this work.}
		
		\begin{abstract}
	
Nowadays, fast delivery services have created the need for high-density warehouses. The puzzle-based storage system is a practical way to enhance the storage density, however, facing difficulties in the retrieval process. In this work, a deep reinforcement learning algorithm, specifically the Double\&Dueling Deep Q Network, is developed to solve the multi-item retrieval problem in the system with general settings, where multiple desired items, escorts, and I/O points are placed randomly. Additionally, we propose {\color{black}a general compact integer programming model to evaluate the solution quality.} Extensive numerical experiments demonstrate that the {\color{black}reinforcement learning approach can yield} high-quality solutions and outperforms three related state-of-the-art heuristic algorithms. Furthermore, a conversion algorithm and a decomposition framework are proposed to handle simultaneous movement and large-scale instances respectively, thus improving the applicability of the PBS system.
			
		\end{abstract}
		
		
	\end{frontmatter}
	
	


\section{Introduction}

Fast delivery services (e.g., same-day service, next-day service provided by FedEx, JD.com, Alibaba.com, and SF Express, etc.) contribute to enhancing the competitiveness of e-commerce and logistics firms. 
These services are achieved by first forecasting demand using historical data and then transporting inventory in advance
to front warehouses near demand areas (especially large cities, where demand is typically more concentrated). 
Based on such a strategy, it requires more storage space or a more compact storage policy to store enough inventory for near-large-city front warehouses. 
However, it may not be economical to expand warehouse space as land is extraordinarily scarce and costly in large cities. 
Therefore, developing more compact storage systems seems to be a practical way to relieve the space scarcity challenge.

In recent years, two main streams of {\color{black}compact storage systems}
are aisle-based shuttle storage systems and grid-based shuttle storage systems (\cite{Azadeh2019Robotized}). \cite{GueandKevin2006Veryhigh} showed that the storage density {\color{black}(the storage area occupied by pallets or shelves over the total area of the warehouse)} in the aisle-based shuttle storage systems cannot exceed $\frac{2k}{(2k + 1)}$ {\color{black}(achieved with only one aisle)}, where $k$ is the maximum shelves depth. 
Grid-based shuttle storage systems, in which all aisles are eliminated, seem to be 
a better proposal to alleviate the pressure of space scarcity. 
The dynamic version of grid-based shuttle storage systems (i.e., each item/SKU is stored on a shuttle and is individually movable) is called puzzle-based storage (PBS) system. The PBS system {{\color{black}was} originally proposed by \cite{Gue2007PuzzleBased} and has been implemented in warehouses and distribution centers, automated car parking systems (\cite{article}) and container terminals (\cite{Zaerpour2015Storing}). This system can be viewed as a grid with cells, each of which can be empty or occupied by an item, and thus its storage density limit is $\frac{(mn-1)}{mn}$, where $m$ and $n$ denote the size of the grid. As shown in Figure \ref{fig:introduction}, the density limits of these two systems are {\color{black} $\frac{2}{3}$ (note there are three aisles) and $\frac{35}{36}$}, respectively.

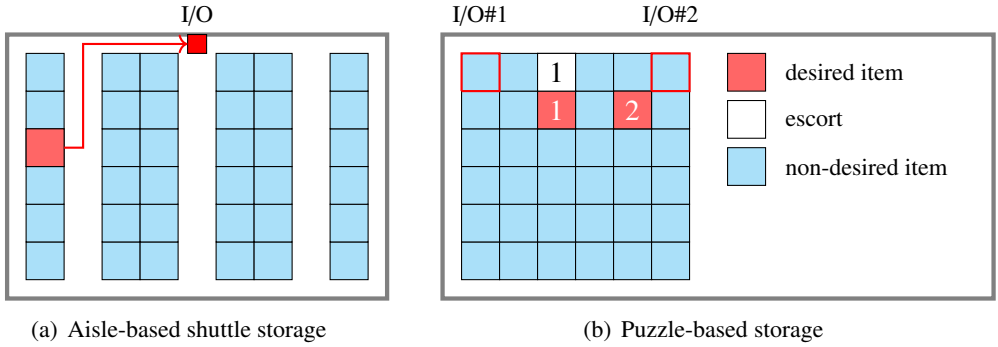
\begin{figure}[H]
\centering
\hspace{-7mm}
	\subfigure[Aisle-based shuttle storage]{
	\label{fig:Traditional layout}
	\tikzstyle{Point}=[circle,draw=black,fill=black,thick,
						inner sep=0pt,minimum size=2mm]
	\tikzstyle{arrow} = [thick,->,>=stealth]
	\tikzstyle{arrowonly} = [thick,-,>=stealth,line width = 2pt, color = cyan]
	\tikzstyle{arrowleft} = [thick,<-,>=stealth]
	\begin{tikzpicture}[auto, scale=0.5]
	 \foreach \x in {1,3,4,6,7,9}
	  \foreach \y in {1,...,6}
	  {
	  \filldraw[fill=cyan!30] (\x,\y) +(-0.5,-0.5) rectangle ++(0.5,0.5);
	  }
	 \filldraw[fill=red!60] (1,4) +(-0.5,-0.5) rectangle ++(0.5,0.5);
	 \draw[ultra thick, gray] (0,0) rectangle (10, 7);
	 \filldraw[fill=red] (4.75,6.5) rectangle ++(0.5,0.5); 
	 \draw (5, 7.5) node(name=I/O){\footnotesize{I/O}};
	 \draw[<-, thick, red] (4.75, 6.75) -- (2, 6.75) -- (2, 4) -- (1.5, 4);
	\end{tikzpicture}
	}
	\subfigure[Puzzle-based storage]{
	\label{fig:Puzzle-basedstorage}
	\tikzstyle{Point}=[circle,draw=black,fill=black,thick,
						inner sep=0pt,minimum size=2mm]
	\tikzstyle{arrow} = [thick,->,>=stealth]
	\tikzstyle{arrowonly} = [thick,-,>=stealth,line width = 2pt, color = cyan]
	\tikzstyle{arrowleft} = [thick,<-,>=stealth]
	\begin{tikzpicture}[auto, scale=0.5]
	 \foreach \x in {1,2,3,4,5,6}
	  \foreach \y in {1,...,6}
	  {
	   \filldraw[fill=cyan!30] (\x,\y) +(-0.5,-0.5) rectangle ++(0.5,0.5);
	  }
	 \filldraw[fill=red!60] (3,5) +(-0.5,-0.5) rectangle ++(0.5,0.5);
	 \draw [white] (3,5) node{1}; 
	 \filldraw[fill=red!60] (5,5) +(-0.5,-0.5) rectangle ++(0.5,0.5);
	 \draw [white] (5,5) node{2}; 
	 \filldraw[fill=white] (3,6) +(-0.5,-0.5) rectangle ++(0.5,0.5);
	 \draw [black] (3,6) node{1}; 

	 \draw[-, thick, red] (0.5, 5.5) -- (0.5, 6.5) -- (1.5, 6.5) -- (1.5, 5.5) -- (0.5, 5.5);
	 \draw[-, thick, red] (5.5, 5.5) -- (5.5, 6.5) -- (6.5, 6.5) -- (6.5, 5.5) -- (5.5, 5.5);
	 \draw[ultra thick, gray] (0,0) rectangle (14.5, 7);
	 \draw (1, 7.5) node(name=I/O){\footnotesize{I/O\#1}};
	 \draw (6, 7.5) node(name=I/O){\footnotesize{I/O\#2}};

	 \filldraw[fill=red!60] (8,6) +(-0.5,-0.5) rectangle ++(0.5,0.5);
	 \draw [black][right] (8.75,6) node{\footnotesize{desired item}}; 
	 \filldraw[fill=white] (8,4.75) +(-0.5,-0.5) rectangle ++(0.5,0.5);
	 \draw [black][right] (8.75,4.75) node{\footnotesize{escort}}; 
	 \filldraw[fill=cyan!30] (8,3.5) +(-0.5,-0.5) rectangle ++(0.5,0.5);
	 \draw [black][right] (8.75,3.5) node{\footnotesize{non-desired item}}; 

	\end{tikzpicture}
	}
	\caption{Example of aisle-based shuttle storage and puzzle-based storage}
	\label{fig:introduction}
\end{figure}

Although the PBS system achieves the highest space utilization, 
the retrieval process is significantly complicated. 
For aisle-based shuttle storage systems, retrieving items is easy as the desired items are moved directly along the empty aisles to reach the I/O points. 
However, the aisle-less configuration in the PBS system forces the items to move only to adjacent empty cells (escorts), requiring both desired and non-desired items to move corporately, thereby complicating the process. Retrieval paths with fewer movements can shorten response time and save energy consumption, thus improving the applicability of the PBS system.
Therefore, how to take fewer movements to finish the tasks remains the key issue in the PBS system.

The existing literature mainly focuses on single-item retrieval
{\color{black}(only one desired item is considered in each retrieval task)}. Nonetheless, according to \cite{Mirzaei2017modelling}, several picking stations work simultaneously in warehouses 
{\color{black}
and it is common to execute multiple retrieval requests parallelly (so-called multi-item retrieval) in real application environments.}
As demonstrated in Section \ref{Multi-item retrieval example}, joint optimization of multi-item retrieval is significantly more efficient than breaking it into several independent single-item retrieval tasks. 
Consequently, to fill this research gap, we concentrate on the multi-item retrieval problem in the general setting of the PBS system (i.e., multiple desired items, multiple randomly placed escorts, multiple I/O points 
are considered). {\color{black}Besides, unlike previous studies that use programming approaches (dynamic programming, integer programming), or search-based heuristic approaches, we aim at developing a learning approach (specifically, reinforcement learning) to solve the multi-item retrieval problem more efficiently.} 
In the past few years, reinforcement learning with a deep neural network has shown its effectiveness in numerous complex tasks.
{\color{black}
The main idea of reinforcement learning is that through a sequence of actions and rewards, the agent learns an action strategy from their interactions with the environment.} Compared to handcrafted heuristics, reinforcement learning can generate high-quality solutions with less CPU time based on the trained models.

This paper makes the following contributions.

 {\color{black}1) We propose a general compact integer programming (IP) formulation for the multi-item retrieval problems in PBS that dramatically reduces the number of decision variables and constraints when compared to the literature.} 
 
 2) We develop a reinforcement learning algorithm for multi-item retrieval problems in general settings under the \textit{single-load movement} 
{\color{black}(only one move is allowed at a timestamp)}
assumption. specifically, a \textit{semi-random and semi-guided} action selection mechanism is designed and integrated into the approach to address the convergence issue, which remains a typical challenge for reinforcement learning. Extensive numerical experiments demonstrate that the reinforcement learning solutions are quite close to optimum and outperform those of the existing heuristics. 

3) A \textit{conversion} algorithm is proposed to consider \textit{simultaneous movement} 
{\color{black}(multiple moves are allowed at a timestamp)}, thereby significantly reducing the retrieval time.

4) We propose a \textit{decomposition} framework to handle large-scale instances in seconds, which performs comparably to state-of-the-art algorithms {\color{black}while is capable of solving multi-item retrieval problems}.

The rest of this paper is structured as follows. In Section 2, we summarize the related literature about the PBS system and reinforcement learning. Section 3 describes the multi-item retrieval problem in detail and then provides an illustrative example to explain that the multi-item retrieval is indeed necessary. In Sections 4 and 5, an IP formulation and a reinforcement learning method are presented, respectively. Sections 6 and 7 introduce the decomposition framework and the conversion algorithm. Section 8 discusses the generation of benchmark instances and the experimental results. Section 9 summarizes this work and discusses its limitations and potential future directions.

\section{Related Work}

\subsection{Puzzle-based storage}
\label{sec:PBS}

Existing research on the PBS and retrieval system involves three major categories: 
system analysis, design optimization, 
and operations planning and control. Interested readers can refer to \cite{Azadeh2019Robotized} for more information about the PBS system and other types of compact storage systems. 
The system analysis-oriented literature mainly concerns about the retrieval performance of the PBS system, including
(i) evaluating the expected retrieval time and comparing it with that of traditional systems (\cite{Gue2007PuzzleBased}, \cite{Zaerpour2017aSmall}), 
(ii) analyzing the effect of escort location on retrieval time (\cite{Kota2015Retrieval}). 
Design optimization-related studies prefer to pay more attention to warehouse shape optimization under various storage policies (\cite{Zaerpour2017aSmall}, \cite{Zaerpour2017boptimal}, \cite{Zaerpour2017cResponse}) and emerging technologies on advanced systems
as well as smarter equipments (\cite{Gue2013Gridstore}). Operations planning and control problems are dedicated to optimizing the retrieval path (\cite{Gue2007PuzzleBased}, \cite{Rohit2010Retrieval}, \cite{Yu2016optimal}, \cite{Yalcin2019}, \cite{Alfieri2012Heuristics}, \cite{Gue2013Gridstore}, \cite{Mirzaei2017modelling}), or AGV dispatching plan (\cite{Alfieri2012Heuristics}). Since that the third type of research is very close to this work and the other two are not, we will not {{\color{black}go into the detail} of those categories. 
In this work, we assume that the resources (AGV, etc.) are well-equipped and thus we focus on the optimization of the retrieval path.

\begin{scriptsize}
		\begin{longtable}{m{3cm}m{4cm}m{3cm}m{5cm}}
		\caption{Item retrieval optimization literature in the PBS system}   
		\label{longtable:Item retrieval optimization literature}    \\  
		\hline 
			{\color{blue} Reference}  
			& {\color{blue} Topic} 
			& {\color{blue} Methodology} 
			& {\color{blue} 
			\tabincell{l}{Problem type \\ 
			\tiny Desired items $\times$ Escorts $\times$ I/O Points $\times$ Move type $\times$ Objective
			 }
			}  
			\\
			\hline
			\endfirsthead
			
			\hline 
			{\color{blue} Reference}  
			& {\color{blue} Topic} 
			& {\color{blue} Methodology} 
			& {\color{blue} 
			\tabincell{l}{Problem type \\ 
			\tiny Desired items $\times$ Escorts $\times$ I/O Points $\times$ Move type  \\ 
			\tiny $\times$ Objective
			 }
			} 
			\\
			\hline
			\endhead
			
		\hline
		\endfoot 	

		\cite{Gue2007PuzzleBased}
		& \tabincell{l}{Single-item retrieval path\\optimization with fixed escort\\locations (consider block move)}  
		& \tabincell{l}{Closed-form formular, \\ dynamic Programming, \\ heuristic  }
		& \tabincell{l}{S $\times$ S-F $\times$ S $\times$ lm $\times$ $T_{\max}$ \\
		S $\times$ S-F $\times$ S $\times$ bm $\times$ $T_{\max}$ 
		\\
		S $\times$ M-F $\times$ S $\times$ bm $\times$ $T_{\max}$}
		\\ \hline 

		\cite{Taylor2008Effects} 
		& Evaluate effect of escort locations   
		& \tabincell{l}{Discrete time simulation}
		& \tabincell{l}{S $\times$ M-R $\times$ S $\times$ lm $\times$ $T_{\max}$ }
		\\ \hline 			

		\cite{Alfieri2012Heuristics} 
		& Optimize the shelf movements and AGV dispatching    
		& \tabincell{l}{Heuristic}
		& \tabincell{l}{S $\times$ M-R $\times$ S $\times$ AGV $\times$ $T_{\max}$}
		\\ \hline 
				
		\cite{Gue2013Gridstore} 
		& Deadlock free decentralized control scheme, effect of WIP and escorts on the throughput rate    
		& \tabincell{l}{Heuristic}
		& \tabincell{l}{M $\times$ M-R $\times$ conveyor $\times$ GS\& sm $\times$ $ST$
		\\
		Multiple (arbitrary number) Desired items}
		\\ \hline 		
	
		\cite{Kota2015Retrieval} 
		& \tabincell{l}{Retrieval time estimation with\\randomly located escorts}    
		& \tabincell{l}{Closed-form formular\\
		Heuristics}
		& \tabincell{l}{S $\times$ S-R $\times$ M(2)-R $\times$ S $\times$ lm $\times$ $T_{\max}$
		\\
		S $\times$ M-R $\times$ S $\times$ lm $\times$ $T_{\max}$}
		\\ \hline

		\cite{Ma2021Efficient} 
        & \tabincell{l}{\color{black}Single-item retrieval with multiple \\\color{black}randomly located escorts}    
        & \tabincell{l}{\color{black}Heuristics}
        & \tabincell{l}{\color{black}S $\times$ M-R $\times$ S $\times$ lm $\times$ $N_{m}$}
        \\ \hline 

		\cite{Mirzaei2017modelling} 
		& Multi-item retrieval  
		& \tabincell{l}{Closed-form formular\\
		Heuristics}
		& \tabincell{l}{M(2) $\times$ S-R $\times$ S $\times$ lm $\times$ $T_{\max}$
		\\
		M(3) $\times$ S-R $\times$ S $\times$ lm $\times$ $T_{\max}$}
		\\ \hline 

		\cite{Yalcin2019} 
		& Single-item retrieval  
		& \tabincell{l}{Optimal\\
		Heuristics}
		& \tabincell{l}{S $\times$ M-R $\times$ M $\times$ lm $\times$ $N_m$}
		\\ \hline 
						
		\cite{Yu2016optimal} 
		& Single-item retrieval, effect of simultaneous and block movement of 
		items and escorts 
		& \tabincell{l}{Integer programming}
		& \tabincell{l}{S $\times$ M-R $\times$ S $\times$ sm, bm $\times$ $T_{\max}$}
		\\ \hline 

		\cite{Bukchin2020optimal} 
		& Single-item retrieval
		& \tabincell{l}{Heuristic}
		& \tabincell{l}{S $\times$ M-R $\times$ M $\times$ lm $\times$ $T_{\max}$
		\\
		S $\times$ M-R $\times$ M $\times$ bm $\times$ $T_{\max}$
		\\
		S $\times$ M-R $\times$ M $\times$ sm, lm $\times$ $T_{\max} + \alpha N_m$
		\\
		S $\times$ M-R $\times$ M $\times$ sm, bm $\times$ $T_{\max} + \alpha N_m$}
		\\ \hline 

		\cite{Rohit2010Retrieval} 
		& Single-item retrieval 
		& \tabincell{l}{Integer programming}
		& \tabincell{l}{S $\times$ M-R $\times$ S $\times$ lm $\times$ $N_{m}$}
		\\ \hline 

		\cite{zou2021heuristic}
		& Muti-item retrieval 
		& \tabincell{l}{Heuristic}
		& \tabincell{l}{M $\times$ M-R $\times$ M $\times$ lm $\times$ $N_{m}$}
		\\ \hline 

		This work
		& \tabincell{l}{Multi-item retrieval} 
		& \tabincell{l}{Integer programming \\
		Reinforcement Lerning}
		& \tabincell{l}{M $\times$ M-R $\times$ M $\times$ lm, sm $\times$ $N_{m}$}
		\\ \hline 
	    \end{longtable}	
	    \vspace{-0.6cm}
	        \begin{tablenotes}
	            \item S: single;
	            M: multiple, M($k$) denotes that the number equals to $k$;
	            S-R: single randomly placed escort; 
	            M-R: multiple randomly placed escorts;  
	            S-F: single fixed-position escort; 
	            M-F: multiple fixed-position escorts; 
	            lm: single-load movement;  
	            bm: block movement;
	            sm: simultaneous movement, different loads are allowed to move at a timestamp; 
	            $T_{\max}$: the completion time of the retrieval process;
	            $N_m$: the total number of moves; 
	            $ST$: the system throughput; 
	            AGV: refers to a system where automated guided vehicles (AGVs) are used to move the loads;  
	            GS: refers to a system based on GridStore technology. 
	        \end{tablenotes}
\end{scriptsize} 

We identify the relevant work for item retrieval {{\color{black}by the
\textit{desired item number}} (single, multiple), 
\textit{escort number} (single or multiple, fixed location or random location), 
\textit{I/O point number} (single or multiple), 
\textit{load movement type} (single-load movement, simultaneous movement, and block movement) and 
\textit{objective function} (minimize the total completion time of the retrieval process, the total number of moves and the system throughput). 
The relevant literature is summarized in Table \ref{longtable:Item retrieval optimization literature}. 

The majority of the existing literature studies single-item retrieval problem.
Optimizing the retrieval process and evaluating the performance of the PBS system are two major concerns in this field. \cite{Gue2007PuzzleBased} {\color{black}aim} to estimate the expected retrieval time under some special cases (e.g., considering a single, fixed-position escort, etc.). Besides, sufficient comparisons of retrieval performance between the PBS system and traditional systems are provided. 
To explore the more general cases, several follow-up literature has considered multiple randomly located escorts under the single-load movement assumption (\cite{Taylor2008Effects}, \cite{Kota2015Retrieval}, \cite{Yalcin2019}, \cite{Rohit2010Retrieval}), and multiple randomly located escorts under the simultaneous or block movement assumptions (\cite{Yu2016optimal}, \cite{Bukchin2020optimal}). 
Specifically, \cite{Kota2015Retrieval} {\color{black}extend} the analysis results to estimate retrieval performance, observing that retrieval time depends heavily on escorts' locations. Similar investigation see \cite{Taylor2008Effects}. 
\cite{Rohit2010Retrieval} {\color{black}present} a general IP formulation for single-item retrieval with multiple arbitrarily located escorts. It fails to solve medium- or large-scale instances because of the dramatic increase in the number of variables and constraints. \cite{Yalcin2019} {\color{black}focus} more on algorithm design and proposes a search-based optimal approach and a heuristic algorithm for small and large instances, respectively. {\color{black}\cite{Ma2021Efficient} develop an efficient hybrid heuristic algorithm that incorporates the state evaluation, neighborhood search, as well as beam search techniques to further improve the solution quality and reduce the computational complexity.} 
All the above studies make the single-load motion assumption, which simplifies the problem considerably but seems to be less practical, leading to a waste of retrieve time. Therefore, it is essential to consider simultaneous and block movement. \cite{Alfieri2012Heuristics} {\color{black}assume} that the retrieval actions are performed by a limited number of automated vehicles, and develops a heuristic approach to optimize the AGVs' scheduling and shelves' movement. \cite{Bukchin2020optimal} {\color{black}develop} an exact dynamic programming algorithm to solve the single-item retrieval problem that allows simultaneous or block movement.

However, the above studies only consider single-item retrieval, which is still far from the real warehouse operation.  
In practice, multiple picking stations work parallelly to pursue a shorter system-wide picking time, typically in manufacturing environments and e-commerce warehouses. Therefore, multi-item 
retrieval optimization tends to be essential and urgent. However, few research efforts have attempted to cope with this challenge so far. The only three closely relevant papers are \cite{Gue2013Gridstore}, \cite{Mirzaei2017modelling} and \cite{zou2021heuristic}. 
\cite{Gue2013Gridstore} {\color{black}discuss} the retrieval based on GridStore technology. Since the focus of their work is not on algorithm design, it is beyond our scope. 
\cite{Mirzaei2017modelling} {\color{black}propose} a closed-form formula for the two desired items retrieval case where two loads are first moved to a joining location and then brought together to the I/O point. Moreover, a heuristic is extended to retrieve three and more loads. But the method is valid only for systems with very high space utilization, i.e. only one escort, which is quite unrealistic. \cite{zou2021heuristic} {\color{black}propose} a heuristic algorithm to handle the multi-item retrieval problem with multiple randomly located escorts and I/O points. The method, however, has some incompleteness, resulting in a relatively high failure rate.

\subsection{Reinforcement learning and its applications}
\label{sec:RL}

Our work is closely related to reinforcement learning, {\color{black}specifically the Q-learning and Deep Q-Network (DQN) algorithms.}
{\color{black} Reinforcement learning (RL) is a learning paradigm tailored to solve sequential decision problems (e.g., consider sequencing the moves of the desired item in our problem). According to RL, an agent (an escort in the PBS system) learns to action based on the feedback it received from the environment (the PBS system). 
The value of taking an action on a state is evaluated by the state-action function, also called the Q function. Therefore, getting such a Q function is critical for solving the problem. 
Details on RL can be referred to \cite{sutton2018reinforcement}. Q-learning is a model-free algorithm for RL to learn the Q function without knowing a model of the environment (e.g., the state transition probabilities).} 
\cite{Watkins1992Technical} {\color{black}present} in detail that Q-learning converges to optimal action values based on his previous work (\cite{1989Learning}), but it is limited to domains with fully observed low-dimensional state spaces. {\color{black}DQN is a Deep Q-learning algorithm that overcomes this shortcoming by using a deep neural network to approximate the Q function.} \cite{2013Playing} {\color{black}apply} {\color{black}DQN} to learn control policies in Atari 2600 games, which was the first to introduce the deep learning model into reinforcement learning, making it possible for the agent to derive efficient representations of the environment from high-dimensional sensory inputs. Later enhanced {\color{black}DQN version in \cite{Mnih2015} }outperforms all previous approaches, showing the great potential of deep learning models with reinforcement learning. \cite{2015Deep} {\color{black}present} Double DQN to separate action selection from value estimation to avoid overestimation. Dueling DQN, proposed by \cite{wang2016dueling}, designs two separate estimators for the state value function and the action advantage function respectively, which leads to better policy evaluation, particularly when the Q values are close. Inspired by the human thought process, \cite{schaul2015prioritized} {\color{black}emphasize} the critical role of prioritized experience replay in DQN training, and how DQN with prioritized experience replay can improve sample utilization efficiency. More details of reinforcement learning can be found in \cite{sutton2018reinforcement}.

In recent years, a large number of applications of reinforcement learning exist in logistics and transportation related fields, 
e.g., manufacturing planning for material handling systems (\cite{Swetha2019Applying}, \cite{Li2018Simulation}),  
routing in baggage handling systems (\cite{mukhutdinov2019multi}), 
order dispatching in ride-hailing/ride-sharing systems (\cite{xu2018large}, \cite{tang2019deep}), 
rebalancing in bike-sharing system (\cite{pan2019deep}).
Reinforcement learning also shows promising potential for solving classical combinatorial optimization problems, e.g., travelling salesman problem (\cite{Vinyals2015Pointer}), 
vehicle routing problems (\cite{kool2018attention}, \cite{nazari2018reinforcement}) etc.  

To our knowledge, few studies have focused on the general multi-item retrieval problem in the PBS system. Additionally, no existing work has tried to address this problem in a deep reinforcement learning manner and further explores how to solve potential difficulties (e.g., convergence problems). This work attempts to fill this gap and extend existing research.

\section{Problem Description}

\subsection{Multi-item retrieval}
Consider a PBS system of size $m \times n$ with $d$ ($d \geqslant 1$) desired items, $e$ ($e \geqslant 1$) escorts and $d$ I/O points, where $m, n$ are the number of rows and columns respectively. 
The locations of I/O points can be chosen randomly but must be determined before training; this is reasonable given that picking stations in the warehouse do not generally move. 
{\color{black} Typically, the number of desired items can be more or less than the I/O points; however, we can always execute a pre-assignment step to assign a corresponding I/O point for each desired item. Therefore, we set the number of the desired items and the I/O points to be the same. We leave the optimization of the pre-assignment step for future work due to the complexity of the entire problem.}

The objective of this work is to minimize the total number of moves. To simplify the modeling, 
{\color{black}we first assume that each step allows for only one move, i.e., the single-load movement assumption.} Then, we offer a conversion algorithm to transfer the results to the simultaneous movement version.
\subsection{Retrieve jointly and retrieve separately}\label{Multi-item retrieval example}
\label{sec:32}
An obvious approximate solution to a multi-item retrieval problem is to break it down into several single-item retrieval problems and solve them sequentially. Note that this approach provides an upper bound for the original problem. Here we give an example (Figure \ref{fig:Non-optimal path} and \ref{fig:optimal path}) to illustrate the difference between joint retrieval and separate retrieval. Suppose there are two desired items ([2,2], [1,1]), two escorts ([0,1], [1,2]) and two I/O points ([0,0], [0,3]) in a $4 \times 4$ grid ([0,0] is at the left-up corner). 
Intuitively, it is easy to decompose the problem into two sequential single-item retrieval tasks (e.g., retrieve $d_1$ first, then $d_2$, or retrieve $d_2$ first, then $d_1$). At least 17 moves are needed when retrieving $d_1$ first, whereas only 15 moves are required {{\color{black}if we retrieve $d_2$ first}. Figure \ref{fig:Non-optimal path} shows the optimal path under the separate retrieval policy. While according to our proposed multi-item retrieval algorithm (described in section 5), the minimum number of moves is 13 under the joint retrieval policy, resulting in a saving of 15.4\% (as shown in Figure \ref{fig:optimal path}). Thus, it is not desirable to simply decompose multi-item retrieval problems into single-item retrieval ones and multi-item retrieval is worthy of dedicated optimization. 
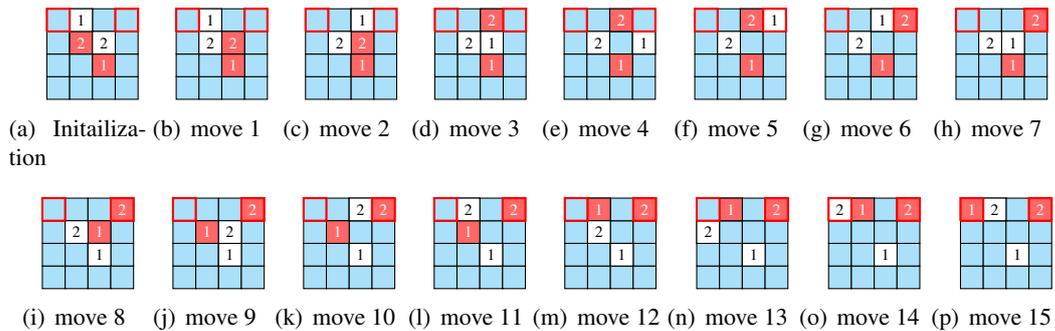
\begin{figure}[H]
\centering 
\hspace{-7mm}
	\subfigure[Initailization]{
	\tikzstyle{Point}=[circle,draw=black,fill=black,thick,
						inner sep=0pt,minimum size=2mm]
	\tikzstyle{arrow} = [thick,->,>=stealth]
	\tikzstyle{arrowonly} = [thick,-,>=stealth,line width = 2pt, color = cyan]
	\tikzstyle{arrowleft} = [thick,<-,>=stealth]
	\begin{tikzpicture}[auto, scale=0.3]
	 \foreach \x in {1,2,3,4}
	  \foreach \y in {1,...,4}
	  {
	   \filldraw[fill=cyan!30] (\x,\y) +(-0.5,-0.5) rectangle ++(0.5,0.5);
	  }
	 \filldraw[fill=red!60] (3,2) +(-0.5,-0.5) rectangle ++(0.5,0.5);
	 \draw [white] (3,2) node{\tiny{1}}; 
	 \filldraw[fill=red!60] (2,3) +(-0.5,-0.5) rectangle ++(0.5,0.5);
	 \draw [white] (2,3) node{\tiny{2}}; 
	 \filldraw[fill=white] (2,4) +(-0.5,-0.5) rectangle ++(0.5,0.5);
	 \draw [black] (2,4) node{\tiny{1}}; 
	 \filldraw[fill=white] (3,3) +(-0.5,-0.5) rectangle ++(0.5,0.5);
	 \draw [black] (3,3) node{\tiny{2}}; 

	 \draw[-, thick, red] (0.5, 3.5) -- (1.5, 3.5) -- (1.5, 4.5) -- (0.5, 4.5) -- (0.5, 3.5);
	 \draw[-, thick, red] (3.5, 3.5) -- (3.5, 4.5) -- (4.5, 4.5) -- (4.5, 3.5) -- (3.5, 3.5);
	\end{tikzpicture}
	}\hspace{-2mm}
	\subfigure[move 1]{
	\tikzstyle{Point}=[circle,draw=black,fill=black,thick,
						inner sep=0pt,minimum size=2mm]
	\tikzstyle{arrow} = [thick,->,>=stealth]
	\tikzstyle{arrowonly} = [thick,-,>=stealth,line width = 2pt, color = cyan]
	\tikzstyle{arrowleft} = [thick,<-,>=stealth]
	\begin{tikzpicture}[auto, scale=0.3]
	 \foreach \x in {1,2,3,4}
	  \foreach \y in {1,...,4}
	  {
	   \filldraw[fill=cyan!30] (\x,\y) +(-0.5,-0.5) rectangle ++(0.5,0.5);
	  }
	 \filldraw[fill=red!60] (3,2) +(-0.5,-0.5) rectangle ++(0.5,0.5);
	 \draw [white] (3,2) node{\tiny{1}}; 
	 \filldraw[fill=red!60] (3,3) +(-0.5,-0.5) rectangle ++(0.5,0.5);
	 \draw [white] (3,3) node{\tiny{2}}; 
	 \filldraw[fill=white] (2,4) +(-0.5,-0.5) rectangle ++(0.5,0.5);
	 \draw [black] (2,4) node{\tiny{1}}; 
	 \filldraw[fill=white] (2,3) +(-0.5,-0.5) rectangle ++(0.5,0.5);
	 \draw [black] (2,3) node{\tiny{2}};  

	 \draw[-, thick, red] (0.5, 3.5) -- (1.5, 3.5) -- (1.5, 4.5) -- (0.5, 4.5) -- (0.5, 3.5);
	 \draw[-, thick, red] (3.5, 3.5) -- (3.5, 4.5) -- (4.5, 4.5) -- (4.5, 3.5) -- (3.5, 3.5);
	\end{tikzpicture}
	}\hspace{-2mm}
	\subfigure[move 2]{
	\tikzstyle{Point}=[circle,draw=black,fill=black,thick,
						inner sep=0pt,minimum size=2mm]
	\tikzstyle{arrow} = [thick,->,>=stealth]
	\tikzstyle{arrowonly} = [thick,-,>=stealth,line width = 2pt, color = cyan]
	\tikzstyle{arrowleft} = [thick,<-,>=stealth]
	\begin{tikzpicture}[auto, scale=0.3]
	 \foreach \x in {1,2,3,4}
	  \foreach \y in {1,...,4}
	  {
	   \filldraw[fill=cyan!30] (\x,\y) +(-0.5,-0.5) rectangle ++(0.5,0.5);
	  }
	 \filldraw[fill=red!60] (3,2) +(-0.5,-0.5) rectangle ++(0.5,0.5);
	 \draw [white] (3,2) node{\tiny{1}}; 
	 \filldraw[fill=red!60] (3,3) +(-0.5,-0.5) rectangle ++(0.5,0.5);
	 \draw [white] (3,3) node{\tiny{2}}; 
	 \filldraw[fill=white] (3,4) +(-0.5,-0.5) rectangle ++(0.5,0.5);
	 \draw [black] (3,4) node{\tiny{1}}; 
	 \filldraw[fill=white] (2,3) +(-0.5,-0.5) rectangle ++(0.5,0.5);
	 \draw [black] (2,3) node{\tiny{2}};

	 \draw[-, thick, red] (0.5, 3.5) -- (1.5, 3.5) -- (1.5, 4.5) -- (0.5, 4.5) -- (0.5, 3.5);
	 \draw[-, thick, red] (3.5, 3.5) -- (3.5, 4.5) -- (4.5, 4.5) -- (4.5, 3.5) -- (3.5, 3.5);
	\end{tikzpicture}
	}\hspace{-2mm}
	\subfigure[move 3]{
	\tikzstyle{Point}=[circle,draw=black,fill=black,thick,
						inner sep=0pt,minimum size=2mm]
	\tikzstyle{arrow} = [thick,->,>=stealth]
	\tikzstyle{arrowonly} = [thick,-,>=stealth,line width = 2pt, color = cyan]
	\tikzstyle{arrowleft} = [thick,<-,>=stealth]
	\begin{tikzpicture}[auto, scale=0.3]
	 \foreach \x in {1,2,3,4}
	  \foreach \y in {1,...,4}
	  {
	   \filldraw[fill=cyan!30] (\x,\y) +(-0.5,-0.5) rectangle ++(0.5,0.5);
	  }
	 \filldraw[fill=red!60] (3,2) +(-0.5,-0.5) rectangle ++(0.5,0.5);
	 \draw [white] (3,2) node{\tiny{1}}; 
	 \filldraw[fill=red!60] (3,4) +(-0.5,-0.5) rectangle ++(0.5,0.5);
	 \draw [white] (3,4) node{\tiny{2}}; 
	 \filldraw[fill=white] (3,3) +(-0.5,-0.5) rectangle ++(0.5,0.5);
	 \draw [black] (3,3) node{\tiny{1}}; 
	 \filldraw[fill=white] (2,3) +(-0.5,-0.5) rectangle ++(0.5,0.5);
	 \draw [black] (2,3) node{\tiny{2}};

	 \draw[-, thick, red] (0.5, 3.5) -- (1.5, 3.5) -- (1.5, 4.5) -- (0.5, 4.5) -- (0.5, 3.5);
	 \draw[-, thick, red] (3.5, 3.5) -- (3.5, 4.5) -- (4.5, 4.5) -- (4.5, 3.5) -- (3.5, 3.5);
	\end{tikzpicture}
	}\hspace{-2mm}
	\subfigure[move 4]{
	\tikzstyle{Point}=[circle,draw=black,fill=black,thick,
						inner sep=0pt,minimum size=2mm]
	\tikzstyle{arrow} = [thick,->,>=stealth]
	\tikzstyle{arrowonly} = [thick,-,>=stealth,line width = 2pt, color = cyan]
	\tikzstyle{arrowleft} = [thick,<-,>=stealth]
	\begin{tikzpicture}[auto, scale=0.3]
	 \foreach \x in {1,2,3,4}
	  \foreach \y in {1,...,4}
	  {
	   \filldraw[fill=cyan!30] (\x,\y) +(-0.5,-0.5) rectangle ++(0.5,0.5);
	  }
	 \filldraw[fill=red!60] (3,2) +(-0.5,-0.5) rectangle ++(0.5,0.5);
	 \draw [white] (3,2) node{\tiny{1}}; 
	 \filldraw[fill=red!60] (3,4) +(-0.5,-0.5) rectangle ++(0.5,0.5);
	 \draw [white] (3,4) node{\tiny{2}}; 
	 \filldraw[fill=white] (4,3) +(-0.5,-0.5) rectangle ++(0.5,0.5);
	 \draw [black] (4,3) node{\tiny{1}}; 
	 \filldraw[fill=white] (2,3) +(-0.5,-0.5) rectangle ++(0.5,0.5);
	 \draw [black] (2,3) node{\tiny{2}};

	 \draw[-, thick, red] (0.5, 3.5) -- (1.5, 3.5) -- (1.5, 4.5) -- (0.5, 4.5) -- (0.5, 3.5);
	 \draw[-, thick, red] (3.5, 3.5) -- (3.5, 4.5) -- (4.5, 4.5) -- (4.5, 3.5) -- (3.5, 3.5);
	\end{tikzpicture}
	}\hspace{-2mm}
	\subfigure[move 5]{
	\tikzstyle{Point}=[circle,draw=black,fill=black,thick,
						inner sep=0pt,minimum size=2mm]
	\tikzstyle{arrow} = [thick,->,>=stealth]
	\tikzstyle{arrowonly} = [thick,-,>=stealth,line width = 2pt, color = cyan]
	\tikzstyle{arrowleft} = [thick,<-,>=stealth]
	\begin{tikzpicture}[auto, scale=0.3]
	 \foreach \x in {1,2,3,4}
	  \foreach \y in {1,...,4}
	  {
	   \filldraw[fill=cyan!30] (\x,\y) +(-0.5,-0.5) rectangle ++(0.5,0.5);
	  }
	 \filldraw[fill=red!60] (3,2) +(-0.5,-0.5) rectangle ++(0.5,0.5);
	 \draw [white] (3,2) node{\tiny{1}}; 
	 \filldraw[fill=red!60] (3,4) +(-0.5,-0.5) rectangle ++(0.5,0.5);
	 \draw [white] (3,4) node{\tiny{2}}; 
	 \filldraw[fill=white] (4,4) +(-0.5,-0.5) rectangle ++(0.5,0.5);
	 \draw [black] (4,4) node{\tiny{1}}; 
	 \filldraw[fill=white] (2,3) +(-0.5,-0.5) rectangle ++(0.5,0.5);
	 \draw [black] (2,3) node{\tiny{2}};

	 \draw[-, thick, red] (0.5, 3.5) -- (1.5, 3.5) -- (1.5, 4.5) -- (0.5, 4.5) -- (0.5, 3.5);
	 \draw[-, thick, red] (3.5, 3.5) -- (3.5, 4.5) -- (4.5, 4.5) -- (4.5, 3.5) -- (3.5, 3.5);
	\end{tikzpicture}
	}\hspace{-2mm}
	\subfigure[move 6]{
	\tikzstyle{Point}=[circle,draw=black,fill=black,thick,
						inner sep=0pt,minimum size=2mm]
	\tikzstyle{arrow} = [thick,->,>=stealth]
	\tikzstyle{arrowonly} = [thick,-,>=stealth,line width = 2pt, color = cyan]
	\tikzstyle{arrowleft} = [thick,<-,>=stealth]
	\begin{tikzpicture}[auto, scale=0.3]
	 \foreach \x in {1,2,3,4}
	  \foreach \y in {1,...,4}
	  {
	   \filldraw[fill=cyan!30] (\x,\y) +(-0.5,-0.5) rectangle ++(0.5,0.5);
	  }
	 \filldraw[fill=red!60] (3,2) +(-0.5,-0.5) rectangle ++(0.5,0.5);
	 \draw [white] (3,2) node{\tiny{1}}; 
	 \filldraw[fill=red!60] (4,4) +(-0.5,-0.5) rectangle ++(0.5,0.5);
	 \draw [white] (4,4) node{\tiny{2}}; 
	 \filldraw[fill=white] (3,4) +(-0.5,-0.5) rectangle ++(0.5,0.5);
	 \draw [black] (3,4) node{\tiny{1}}; 
	 \filldraw[fill=white] (2,3) +(-0.5,-0.5) rectangle ++(0.5,0.5);
	 \draw [black] (2,3) node{\tiny{2}};

	 \draw[-, thick, red] (0.5, 3.5) -- (1.5, 3.5) -- (1.5, 4.5) -- (0.5, 4.5) -- (0.5, 3.5);
	 \draw[-, thick, red] (3.5, 3.5) -- (3.5, 4.5) -- (4.5, 4.5) -- (4.5, 3.5) -- (3.5, 3.5);
	\end{tikzpicture}
	}\hspace{-2mm}
	\subfigure[move 7]{
	\tikzstyle{Point}=[circle,draw=black,fill=black,thick,
						inner sep=0pt,minimum size=2mm]
	\tikzstyle{arrow} = [thick,->,>=stealth]
	\tikzstyle{arrowonly} = [thick,-,>=stealth,line width = 2pt, color = cyan]
	\tikzstyle{arrowleft} = [thick,<-,>=stealth]
	\begin{tikzpicture}[auto, scale=0.3]
	 \foreach \x in {1,2,3,4}
	  \foreach \y in {1,...,4}
	  {
	   \filldraw[fill=cyan!30] (\x,\y) +(-0.5,-0.5) rectangle ++(0.5,0.5);
	  }
	 \filldraw[fill=red!60] (3,2) +(-0.5,-0.5) rectangle ++(0.5,0.5);
	 \draw [white] (3,2) node{\tiny{1}}; 
	 \filldraw[fill=red!60] (4,4) +(-0.5,-0.5) rectangle ++(0.5,0.5);
	 \draw [white] (4,4) node{\tiny{2}}; 
	 \filldraw[fill=white] (3,3) +(-0.5,-0.5) rectangle ++(0.5,0.5);
	 \draw [black] (3,3) node{\tiny{1}}; 
	 \filldraw[fill=white] (2,3) +(-0.5,-0.5) rectangle ++(0.5,0.5);
	 \draw [black] (2,3) node{\tiny{2}}; 

	 \draw[-, thick, red] (0.5, 3.5) -- (1.5, 3.5) -- (1.5, 4.5) -- (0.5, 4.5) -- (0.5, 3.5);
	 \draw[-, thick, red] (3.5, 3.5) -- (3.5, 4.5) -- (4.5, 4.5) -- (4.5, 3.5) -- (3.5, 3.5);
	\end{tikzpicture}
	}

	\hspace{-7mm}
	\subfigure[move 8]{
	\tikzstyle{Point}=[circle,draw=black,fill=black,thick,
						inner sep=0pt,minimum size=2mm]
	\tikzstyle{arrow} = [thick,->,>=stealth]
	\tikzstyle{arrowonly} = [thick,-,>=stealth,line width = 2pt, color = cyan]
	\tikzstyle{arrowleft} = [thick,<-,>=stealth]
	\begin{tikzpicture}[auto, scale=0.3]
	 \foreach \x in {1,2,3,4}
	  \foreach \y in {1,...,4}
	  {
	   \filldraw[fill=cyan!30] (\x,\y) +(-0.5,-0.5) rectangle ++(0.5,0.5);
	  }
	 \filldraw[fill=red!60] (3,3) +(-0.5,-0.5) rectangle ++(0.5,0.5);
	 \draw [white] (3,3) node{\tiny{1}}; 
	 \filldraw[fill=red!60] (4,4) +(-0.5,-0.5) rectangle ++(0.5,0.5);
	 \draw [white] (4,4) node{\tiny{2}}; 
	 \filldraw[fill=white] (3,2) +(-0.5,-0.5) rectangle ++(0.5,0.5);
	 \draw [black] (3,2) node{\tiny{1}}; 
	 \filldraw[fill=white] (2,3) +(-0.5,-0.5) rectangle ++(0.5,0.5);
	 \draw [black] (2,3) node{\tiny{2}};

	 \draw[-, thick, red] (0.5, 3.5) -- (1.5, 3.5) -- (1.5, 4.5) -- (0.5, 4.5) -- (0.5, 3.5);
	 \draw[-, thick, red] (3.5, 3.5) -- (3.5, 4.5) -- (4.5, 4.5) -- (4.5, 3.5) -- (3.5, 3.5);
	\end{tikzpicture}
	}\hspace{-2mm}
	\subfigure[move 9]{
	\tikzstyle{Point}=[circle,draw=black,fill=black,thick,
						inner sep=0pt,minimum size=2mm]
	\tikzstyle{arrow} = [thick,->,>=stealth]
	\tikzstyle{arrowonly} = [thick,-,>=stealth,line width = 2pt, color = cyan]
	\tikzstyle{arrowleft} = [thick,<-,>=stealth]
	\begin{tikzpicture}[auto, scale=0.3]
	 \foreach \x in {1,2,3,4}
	  \foreach \y in {1,...,4}
	  {
	   \filldraw[fill=cyan!30] (\x,\y) +(-0.5,-0.5) rectangle ++(0.5,0.5);
	  }
	 \filldraw[fill=red!60] (2,3) +(-0.5,-0.5) rectangle ++(0.5,0.5);
	 \draw [white] (2,3) node{\tiny{1}}; 
	 \filldraw[fill=red!60] (4,4) +(-0.5,-0.5) rectangle ++(0.5,0.5);
	 \draw [white] (4,4) node{\tiny{2}}; 
	 \filldraw[fill=white] (3,2) +(-0.5,-0.5) rectangle ++(0.5,0.5);
	 \draw [black] (3,2) node{\tiny{1}}; 
	 \filldraw[fill=white] (3,3) +(-0.5,-0.5) rectangle ++(0.5,0.5);
	 \draw [black] (3,3) node{\tiny{2}}; 

	 \draw[-, thick, red] (0.5, 3.5) -- (1.5, 3.5) -- (1.5, 4.5) -- (0.5, 4.5) -- (0.5, 3.5);
	 \draw[-, thick, red] (3.5, 3.5) -- (3.5, 4.5) -- (4.5, 4.5) -- (4.5, 3.5) -- (3.5, 3.5);
	\end{tikzpicture}
	}\hspace{-2mm}
	\subfigure[move 10]{
	\tikzstyle{Point}=[circle,draw=black,fill=black,thick,
						inner sep=0pt,minimum size=2mm]
	\tikzstyle{arrow} = [thick,->,>=stealth]
	\tikzstyle{arrowonly} = [thick,-,>=stealth,line width = 2pt, color = cyan]
	\tikzstyle{arrowleft} = [thick,<-,>=stealth]
	\begin{tikzpicture}[auto, scale=0.3]
	 \foreach \x in {1,2,3,4}
	  \foreach \y in {1,...,4}
	  {
	   \filldraw[fill=cyan!30] (\x,\y) +(-0.5,-0.5) rectangle ++(0.5,0.5);
	  }
	 \filldraw[fill=red!60] (2,3) +(-0.5,-0.5) rectangle ++(0.5,0.5);
	 \draw [white] (2,3) node{\tiny{1}}; 
	 \filldraw[fill=red!60] (4,4) +(-0.5,-0.5) rectangle ++(0.5,0.5);
	 \draw [white] (4,4) node{\tiny{2}}; 
	 \filldraw[fill=white] (3,2) +(-0.5,-0.5) rectangle ++(0.5,0.5);
	 \draw [black] (3,2) node{\tiny{1}}; 
	 \filldraw[fill=white] (3,4) +(-0.5,-0.5) rectangle ++(0.5,0.5);
	 \draw [black] (3,4) node{\tiny{2}};  

	 \draw[-, thick, red] (0.5, 3.5) -- (1.5, 3.5) -- (1.5, 4.5) -- (0.5, 4.5) -- (0.5, 3.5);
	 \draw[-, thick, red] (3.5, 3.5) -- (3.5, 4.5) -- (4.5, 4.5) -- (4.5, 3.5) -- (3.5, 3.5);
	\end{tikzpicture}
	}\hspace{-2mm}
	\subfigure[move 11]{
	\tikzstyle{Point}=[circle,draw=black,fill=black,thick,
						inner sep=0pt,minimum size=2mm]
	\tikzstyle{arrow} = [thick,->,>=stealth]
	\tikzstyle{arrowonly} = [thick,-,>=stealth,line width = 2pt, color = cyan]
	\tikzstyle{arrowleft} = [thick,<-,>=stealth]
	\begin{tikzpicture}[auto, scale=0.3]
	 \foreach \x in {1,2,3,4}
	  \foreach \y in {1,...,4}
	  {
	   \filldraw[fill=cyan!30] (\x,\y) +(-0.5,-0.5) rectangle ++(0.5,0.5);
	  }
	 \filldraw[fill=red!60] (2,3) +(-0.5,-0.5) rectangle ++(0.5,0.5);
	 \draw [white] (2,3) node{\tiny{1}}; 
	 \filldraw[fill=red!60] (4,4) +(-0.5,-0.5) rectangle ++(0.5,0.5);
	 \draw [white] (4,4) node{\tiny{2}}; 
	 \filldraw[fill=white] (3,2) +(-0.5,-0.5) rectangle ++(0.5,0.5);
	 \draw [black] (3,2) node{\tiny{1}}; 
	 \filldraw[fill=white] (2,4) +(-0.5,-0.5) rectangle ++(0.5,0.5);
	 \draw [black] (2,4) node{\tiny{2}}; 

	 \draw[-, thick, red] (0.5, 3.5) -- (1.5, 3.5) -- (1.5, 4.5) -- (0.5, 4.5) -- (0.5, 3.5);
	 \draw[-, thick, red] (3.5, 3.5) -- (3.5, 4.5) -- (4.5, 4.5) -- (4.5, 3.5) -- (3.5, 3.5);
	\end{tikzpicture}
	}\hspace{-2mm}
	\subfigure[move 12]{
	\tikzstyle{Point}=[circle,draw=black,fill=black,thick,
						inner sep=0pt,minimum size=2mm]
	\tikzstyle{arrow} = [thick,->,>=stealth]
	\tikzstyle{arrowonly} = [thick,-,>=stealth,line width = 2pt, color = cyan]
	\tikzstyle{arrowleft} = [thick,<-,>=stealth]
	\begin{tikzpicture}[auto, scale=0.3]
	 \foreach \x in {1,2,3,4}
	  \foreach \y in {1,...,4}
	  {
	   \filldraw[fill=cyan!30] (\x,\y) +(-0.5,-0.5) rectangle ++(0.5,0.5);
	  }
	 \filldraw[fill=red!60] (2,4) +(-0.5,-0.5) rectangle ++(0.5,0.5);
	 \draw [white] (2,4) node{\tiny{1}}; 
	 \filldraw[fill=red!60] (4,4) +(-0.5,-0.5) rectangle ++(0.5,0.5);
	 \draw [white] (4,4) node{\tiny{2}}; 
	 \filldraw[fill=white] (3,2) +(-0.5,-0.5) rectangle ++(0.5,0.5);
	 \draw [black] (3,2) node{\tiny{1}}; 
	 \filldraw[fill=white] (2,3) +(-0.5,-0.5) rectangle ++(0.5,0.5);
	 \draw [black] (2,3) node{\tiny{2}}; 

	 \draw[-, thick, red] (0.5, 3.5) -- (1.5, 3.5) -- (1.5, 4.5) -- (0.5, 4.5) -- (0.5, 3.5);
	 \draw[-, thick, red] (3.5, 3.5) -- (3.5, 4.5) -- (4.5, 4.5) -- (4.5, 3.5) -- (3.5, 3.5);
	\end{tikzpicture}
	}\hspace{-2mm}
	\subfigure[move 13]{
	\tikzstyle{Point}=[circle,draw=black,fill=black,thick,
						inner sep=0pt,minimum size=2mm]
	\tikzstyle{arrow} = [thick,->,>=stealth]
	\tikzstyle{arrowonly} = [thick,-,>=stealth,line width = 2pt, color = cyan]
	\tikzstyle{arrowleft} = [thick,<-,>=stealth]
	\begin{tikzpicture}[auto, scale=0.3]
	 \foreach \x in {1,2,3,4}
	  \foreach \y in {1,...,4}
	  {
	   \filldraw[fill=cyan!30] (\x,\y) +(-0.5,-0.5) rectangle ++(0.5,0.5);
	  }
	 \filldraw[fill=red!60] (2,4) +(-0.5,-0.5) rectangle ++(0.5,0.5);
	 \draw [white] (2,4) node{\tiny{1}}; 
	 \filldraw[fill=red!60] (4,4) +(-0.5,-0.5) rectangle ++(0.5,0.5);
	 \draw [white] (4,4) node{\tiny{2}}; 
	 \filldraw[fill=white] (3,2) +(-0.5,-0.5) rectangle ++(0.5,0.5);
	 \draw [black] (3,2) node{\tiny{1}}; 
	 \filldraw[fill=white] (1,3) +(-0.5,-0.5) rectangle ++(0.5,0.5);
	 \draw [black] (1,3) node{\tiny{2}}; 

	 \draw[-, thick, red] (0.5, 3.5) -- (1.5, 3.5) -- (1.5, 4.5) -- (0.5, 4.5) -- (0.5, 3.5);
	 \draw[-, thick, red] (3.5, 3.5) -- (3.5, 4.5) -- (4.5, 4.5) -- (4.5, 3.5) -- (3.5, 3.5);
	\end{tikzpicture}
	}\hspace{-2mm}
	\subfigure[move 14]{
	\tikzstyle{Point}=[circle,draw=black,fill=black,thick,
						inner sep=0pt,minimum size=2mm]
	\tikzstyle{arrow} = [thick,->,>=stealth]
	\tikzstyle{arrowonly} = [thick,-,>=stealth,line width = 2pt, color = cyan]
	\tikzstyle{arrowleft} = [thick,<-,>=stealth]
	\begin{tikzpicture}[auto, scale=0.3]
	 \foreach \x in {1,2,3,4}
	  \foreach \y in {1,...,4}
	  {
	   \filldraw[fill=cyan!30] (\x,\y) +(-0.5,-0.5) rectangle ++(0.5,0.5);
	  }
	 \filldraw[fill=red!60] (2,4) +(-0.5,-0.5) rectangle ++(0.5,0.5);
	 \draw [white] (2,4) node{\tiny{1}}; 
	 \filldraw[fill=red!60] (4,4) +(-0.5,-0.5) rectangle ++(0.5,0.5);
	 \draw [white] (4,4) node{\tiny{2}}; 
	 \filldraw[fill=white] (3,2) +(-0.5,-0.5) rectangle ++(0.5,0.5);
	 \draw [black] (3,2) node{\tiny{1}}; 
	 \filldraw[fill=white] (1,4) +(-0.5,-0.5) rectangle ++(0.5,0.5);
	 \draw [black] (1,4) node{\tiny{2}}; 

	 \draw[-, thick, red] (0.5, 3.5) -- (1.5, 3.5) -- (1.5, 4.5) -- (0.5, 4.5) -- (0.5, 3.5);
	 \draw[-, thick, red] (3.5, 3.5) -- (3.5, 4.5) -- (4.5, 4.5) -- (4.5, 3.5) -- (3.5, 3.5);
	\end{tikzpicture}
	}\hspace{-2mm}
	\subfigure[move 15]{
	\tikzstyle{Point}=[circle,draw=black,fill=black,thick,
						inner sep=0pt,minimum size=2mm]
	\tikzstyle{arrow} = [thick,->,>=stealth]
	\tikzstyle{arrowonly} = [thick,-,>=stealth,line width = 2pt, color = cyan]
	\tikzstyle{arrowleft} = [thick,<-,>=stealth]
	\begin{tikzpicture}[auto, scale=0.3]
	 \foreach \x in {1,2,3,4}
	  \foreach \y in {1,...,4}
	  {
	   \filldraw[fill=cyan!30] (\x,\y) +(-0.5,-0.5) rectangle ++(0.5,0.5);
	  }
	 \filldraw[fill=red!60] (1,4) +(-0.5,-0.5) rectangle ++(0.5,0.5);
	 \draw [white] (1,4) node{\tiny{1}}; 
	 \filldraw[fill=red!60] (4,4) +(-0.5,-0.5) rectangle ++(0.5,0.5);
	 \draw [white] (4,4) node{\tiny{2}}; 
	 \filldraw[fill=white] (3,2) +(-0.5,-0.5) rectangle ++(0.5,0.5);
	 \draw [black] (3,2) node{\tiny{1}}; 
	 \filldraw[fill=white] (2,4) +(-0.5,-0.5) rectangle ++(0.5,0.5);
	 \draw [black] (2,4) node{\tiny{2}}; 

	 \draw[-, thick, red] (0.5, 3.5) -- (1.5, 3.5) -- (1.5, 4.5) -- (0.5, 4.5) -- (0.5, 3.5);
	 \draw[-, thick, red] (3.5, 3.5) -- (3.5, 4.5) -- (4.5, 4.5) -- (4.5, 3.5) -- (3.5, 3.5);
	\end{tikzpicture}
	}
	\caption{Optimal path when retrieving separately}
	\label{fig:Non-optimal path}
\end{figure}

\begin{figure}[H]
\centering 
\hspace{-7mm}
	\subfigure[Initailization]{
	\tikzstyle{Point}=[circle,draw=black,fill=black,thick,
						inner sep=0pt,minimum size=2mm]
	\tikzstyle{arrow} = [thick,->,>=stealth]
	\tikzstyle{arrowonly} = [thick,-,>=stealth,line width = 2pt, color = cyan]
	\tikzstyle{arrowleft} = [thick,<-,>=stealth]
	\begin{tikzpicture}[auto, scale=0.3]
	 \foreach \x in {1,2,3,4}
	  \foreach \y in {1,...,4}
	  {
	   \filldraw[fill=cyan!30] (\x,\y) +(-0.5,-0.5) rectangle ++(0.5,0.5);
	  }
	 \filldraw[fill=red!60] (3,2) +(-0.5,-0.5) rectangle ++(0.5,0.5);
	 \draw [white] (3,2) node{\tiny{1}}; 
	 \filldraw[fill=red!60] (2,3) +(-0.5,-0.5) rectangle ++(0.5,0.5);
	 \draw [white] (2,3) node{\tiny{2}}; 
	 \filldraw[fill=white] (2,4) +(-0.5,-0.5) rectangle ++(0.5,0.5);
	 \draw [black] (2,4) node{\tiny{1}}; 
	 \filldraw[fill=white] (3,3) +(-0.5,-0.5) rectangle ++(0.5,0.5);
	 \draw [black] (3,3) node{\tiny{2}}; 

	 \draw[-, thick, red] (0.5, 3.5) -- (1.5, 3.5) -- (1.5, 4.5) -- (0.5, 4.5) -- (0.5, 3.5);
	 \draw[-, thick, red] (3.5, 3.5) -- (3.5, 4.5) -- (4.5, 4.5) -- (4.5, 3.5) -- (3.5, 3.5);
	\end{tikzpicture}
	}\hspace{-2mm}
	\subfigure[move 1]{
	\tikzstyle{Point}=[circle,draw=black,fill=black,thick,
						inner sep=0pt,minimum size=2mm]
	\tikzstyle{arrow} = [thick,->,>=stealth]
	\tikzstyle{arrowonly} = [thick,-,>=stealth,line width = 2pt, color = cyan]
	\tikzstyle{arrowleft} = [thick,<-,>=stealth]
	\begin{tikzpicture}[auto, scale=0.3]
	 \foreach \x in {1,2,3,4}
	  \foreach \y in {1,...,4}
	  {
	   \filldraw[fill=cyan!30] (\x,\y) +(-0.5,-0.5) rectangle ++(0.5,0.5);
	  }
	 \filldraw[fill=red!60] (3,3) +(-0.5,-0.5) rectangle ++(0.5,0.5);
	 \draw [white] (3,3) node{\tiny{1}}; 
	 \filldraw[fill=red!60] (2,3) +(-0.5,-0.5) rectangle ++(0.5,0.5);
	 \draw [white] (2,3) node{\tiny{2}}; 
	 \filldraw[fill=white] (2,4) +(-0.5,-0.5) rectangle ++(0.5,0.5);
	 \draw [black] (2,4) node{\tiny{1}}; 
	 \filldraw[fill=white] (3,2) +(-0.5,-0.5) rectangle ++(0.5,0.5);
	 \draw [black] (3,2) node{\tiny{2}}; 

	 \draw[-, thick, red] (0.5, 3.5) -- (1.5, 3.5) -- (1.5, 4.5) -- (0.5, 4.5) -- (0.5, 3.5);
	 \draw[-, thick, red] (3.5, 3.5) -- (3.5, 4.5) -- (4.5, 4.5) -- (4.5, 3.5) -- (3.5, 3.5);
	\end{tikzpicture}
	}\hspace{-2mm}
	\subfigure[move 2]{
	\tikzstyle{Point}=[circle,draw=black,fill=black,thick,
						inner sep=0pt,minimum size=2mm]
	\tikzstyle{arrow} = [thick,->,>=stealth]
	\tikzstyle{arrowonly} = [thick,-,>=stealth,line width = 2pt, color = cyan]
	\tikzstyle{arrowleft} = [thick,<-,>=stealth]
	\begin{tikzpicture}[auto, scale=0.3]
	 \foreach \x in {1,2,3,4}
	  \foreach \y in {1,...,4}
	  {
	   \filldraw[fill=cyan!30] (\x,\y) +(-0.5,-0.5) rectangle ++(0.5,0.5);
	  }
	 \filldraw[fill=red!60] (3,3) +(-0.5,-0.5) rectangle ++(0.5,0.5);
	 \draw [white] (3,3) node{\tiny{1}}; 
	 \filldraw[fill=red!60] (2,4) +(-0.5,-0.5) rectangle ++(0.5,0.5);
	 \draw [white] (2,4) node{\tiny{2}}; 
	 \filldraw[fill=white] (2,3) +(-0.5,-0.5) rectangle ++(0.5,0.5);
	 \draw [black] (2,3) node{\tiny{1}}; 
	 \filldraw[fill=white] (3,2) +(-0.5,-0.5) rectangle ++(0.5,0.5);
	 \draw [black] (3,2) node{\tiny{2}}; 

	 \draw[-, thick, red] (0.5, 3.5) -- (1.5, 3.5) -- (1.5, 4.5) -- (0.5, 4.5) -- (0.5, 3.5);
	 \draw[-, thick, red] (3.5, 3.5) -- (3.5, 4.5) -- (4.5, 4.5) -- (4.5, 3.5) -- (3.5, 3.5);
	\end{tikzpicture}
	}\hspace{-2mm}
	\subfigure[move 3]{
	\tikzstyle{Point}=[circle,draw=black,fill=black,thick,
						inner sep=0pt,minimum size=2mm]
	\tikzstyle{arrow} = [thick,->,>=stealth]
	\tikzstyle{arrowonly} = [thick,-,>=stealth,line width = 2pt, color = cyan]
	\tikzstyle{arrowleft} = [thick,<-,>=stealth]
	\begin{tikzpicture}[auto, scale=0.3]
	 \foreach \x in {1,2,3,4}
	  \foreach \y in {1,...,4}
	  {
	   \filldraw[fill=cyan!30] (\x,\y) +(-0.5,-0.5) rectangle ++(0.5,0.5);
	  }
	 \filldraw[fill=red!60] (2,3) +(-0.5,-0.5) rectangle ++(0.5,0.5);
	 \draw [white] (2,3) node{\tiny{1}}; 
	 \filldraw[fill=red!60] (2,4) +(-0.5,-0.5) rectangle ++(0.5,0.5);
	 \draw [white] (2,4) node{\tiny{2}}; 
	 \filldraw[fill=white] (3,3) +(-0.5,-0.5) rectangle ++(0.5,0.5);
	 \draw [black] (3,3) node{\tiny{1}}; 
	 \filldraw[fill=white] (3,2) +(-0.5,-0.5) rectangle ++(0.5,0.5);
	 \draw [black] (3,2) node{\tiny{2}};  

	 \draw[-, thick, red] (0.5, 3.5) -- (1.5, 3.5) -- (1.5, 4.5) -- (0.5, 4.5) -- (0.5, 3.5);
	 \draw[-, thick, red] (3.5, 3.5) -- (3.5, 4.5) -- (4.5, 4.5) -- (4.5, 3.5) -- (3.5, 3.5);
	\end{tikzpicture}
	}\hspace{-2mm}
	\subfigure[move 4]{
	\tikzstyle{Point}=[circle,draw=black,fill=black,thick,
						inner sep=0pt,minimum size=2mm]
	\tikzstyle{arrow} = [thick,->,>=stealth]
	\tikzstyle{arrowonly} = [thick,-,>=stealth,line width = 2pt, color = cyan]
	\tikzstyle{arrowleft} = [thick,<-,>=stealth]
	\begin{tikzpicture}[auto, scale=0.3]
	 \foreach \x in {1,2,3,4}
	  \foreach \y in {1,...,4}
	  {
	   \filldraw[fill=cyan!30] (\x,\y) +(-0.5,-0.5) rectangle ++(0.5,0.5);
	  }
	 \filldraw[fill=red!60] (2,3) +(-0.5,-0.5) rectangle ++(0.5,0.5);
	 \draw [white] (2,3) node{\tiny{1}}; 
	 \filldraw[fill=red!60] (2,4) +(-0.5,-0.5) rectangle ++(0.5,0.5);
	 \draw [white] (2,4) node{\tiny{2}}; 
	 \filldraw[fill=white] (3,4) +(-0.5,-0.5) rectangle ++(0.5,0.5);
	 \draw [black] (3,4) node{\tiny{1}}; 
	 \filldraw[fill=white] (3,2) +(-0.5,-0.5) rectangle ++(0.5,0.5);
	 \draw [black] (3,2) node{\tiny{2}};

	 \draw[-, thick, red] (0.5, 3.5) -- (1.5, 3.5) -- (1.5, 4.5) -- (0.5, 4.5) -- (0.5, 3.5);
	 \draw[-, thick, red] (3.5, 3.5) -- (3.5, 4.5) -- (4.5, 4.5) -- (4.5, 3.5) -- (3.5, 3.5);
	\end{tikzpicture}
	}\hspace{-2mm}
	\subfigure[move 5]{
	\tikzstyle{Point}=[circle,draw=black,fill=black,thick,
						inner sep=0pt,minimum size=2mm]
	\tikzstyle{arrow} = [thick,->,>=stealth]
	\tikzstyle{arrowonly} = [thick,-,>=stealth,line width = 2pt, color = cyan]
	\tikzstyle{arrowleft} = [thick,<-,>=stealth]
	\begin{tikzpicture}[auto, scale=0.3]
	 \foreach \x in {1,2,3,4}
	  \foreach \y in {1,...,4}
	  {
	   \filldraw[fill=cyan!30] (\x,\y) +(-0.5,-0.5) rectangle ++(0.5,0.5);
	  }
	 \filldraw[fill=red!60] (2,3) +(-0.5,-0.5) rectangle ++(0.5,0.5);
	 \draw [white] (2,3) node{\tiny{1}}; 
	 \filldraw[fill=red!60] (2,4) +(-0.5,-0.5) rectangle ++(0.5,0.5);
	 \draw [white] (2,4) node{\tiny{2}}; 
	 \filldraw[fill=white] (3,4) +(-0.5,-0.5) rectangle ++(0.5,0.5);
	 \draw [black] (3,4) node{\tiny{1}}; 
	 \filldraw[fill=white] (4,2) +(-0.5,-0.5) rectangle ++(0.5,0.5);
	 \draw [black] (4,2) node{\tiny{2}};  

	 \draw[-, thick, red] (0.5, 3.5) -- (1.5, 3.5) -- (1.5, 4.5) -- (0.5, 4.5) -- (0.5, 3.5);
	 \draw[-, thick, red] (3.5, 3.5) -- (3.5, 4.5) -- (4.5, 4.5) -- (4.5, 3.5) -- (3.5, 3.5);
	\end{tikzpicture}
	}\hspace{-2mm}
	\subfigure[move 6]{
	\tikzstyle{Point}=[circle,draw=black,fill=black,thick,
						inner sep=0pt,minimum size=2mm]
	\tikzstyle{arrow} = [thick,->,>=stealth]
	\tikzstyle{arrowonly} = [thick,-,>=stealth,line width = 2pt, color = cyan]
	\tikzstyle{arrowleft} = [thick,<-,>=stealth]
	\begin{tikzpicture}[auto, scale=0.3]
	 \foreach \x in {1,2,3,4}
	  \foreach \y in {1,...,4}
	  {
	   \filldraw[fill=cyan!30] (\x,\y) +(-0.5,-0.5) rectangle ++(0.5,0.5);
	  }
	 \filldraw[fill=red!60] (2,3) +(-0.5,-0.5) rectangle ++(0.5,0.5);
	 \draw [white] (2,3) node{\tiny{1}}; 
	 \filldraw[fill=red!60] (2,4) +(-0.5,-0.5) rectangle ++(0.5,0.5);
	 \draw [white] (2,4) node{\tiny{2}}; 
	 \filldraw[fill=white] (3,4) +(-0.5,-0.5) rectangle ++(0.5,0.5);
	 \draw [black] (3,4) node{\tiny{1}}; 
	 \filldraw[fill=white] (4,3) +(-0.5,-0.5) rectangle ++(0.5,0.5);
	 \draw [black] (4,3) node{\tiny{2}};  

	 \draw[-, thick, red] (0.5, 3.5) -- (1.5, 3.5) -- (1.5, 4.5) -- (0.5, 4.5) -- (0.5, 3.5);
	 \draw[-, thick, red] (3.5, 3.5) -- (3.5, 4.5) -- (4.5, 4.5) -- (4.5, 3.5) -- (3.5, 3.5);
	\end{tikzpicture}
	}

	\hspace{-7mm}
	\subfigure[move 7]{
	\tikzstyle{Point}=[circle,draw=black,fill=black,thick,
						inner sep=0pt,minimum size=2mm]
	\tikzstyle{arrow} = [thick,->,>=stealth]
	\tikzstyle{arrowonly} = [thick,-,>=stealth,line width = 2pt, color = cyan]
	\tikzstyle{arrowleft} = [thick,<-,>=stealth]
	\begin{tikzpicture}[auto, scale=0.3]
	 \foreach \x in {1,2,3,4}
	  \foreach \y in {1,...,4}
	  {
	   \filldraw[fill=cyan!30] (\x,\y) +(-0.5,-0.5) rectangle ++(0.5,0.5);
	  }
	 \filldraw[fill=red!60] (2,3) +(-0.5,-0.5) rectangle ++(0.5,0.5);
	 \draw [white] (2,3) node{\tiny{1}}; 
	 \filldraw[fill=red!60] (2,4) +(-0.5,-0.5) rectangle ++(0.5,0.5);
	 \draw [white] (2,4) node{\tiny{2}}; 
	 \filldraw[fill=white] (3,4) +(-0.5,-0.5) rectangle ++(0.5,0.5);
	 \draw [black] (3,4) node{\tiny{1}}; 
	 \filldraw[fill=white] (4,4) +(-0.5,-0.5) rectangle ++(0.5,0.5);
	 \draw [black] (4,4) node{\tiny{2}};  

	 \draw[-, thick, red] (0.5, 3.5) -- (1.5, 3.5) -- (1.5, 4.5) -- (0.5, 4.5) -- (0.5, 3.5);
	 \draw[-, thick, red] (3.5, 3.5) -- (3.5, 4.5) -- (4.5, 4.5) -- (4.5, 3.5) -- (3.5, 3.5);
	\end{tikzpicture}
	}\hspace{-2mm}
	\subfigure[move 8]{
	\tikzstyle{Point}=[circle,draw=black,fill=black,thick,
						inner sep=0pt,minimum size=2mm]
	\tikzstyle{arrow} = [thick,->,>=stealth]
	\tikzstyle{arrowonly} = [thick,-,>=stealth,line width = 2pt, color = cyan]
	\tikzstyle{arrowleft} = [thick,<-,>=stealth]
	\begin{tikzpicture}[auto, scale=0.3]
	 \foreach \x in {1,2,3,4}
	  \foreach \y in {1,...,4}
	  {
	   \filldraw[fill=cyan!30] (\x,\y) +(-0.5,-0.5) rectangle ++(0.5,0.5);
	  }
	 \filldraw[fill=red!60] (2,3) +(-0.5,-0.5) rectangle ++(0.5,0.5);
	 \draw [white] (2,3) node{\tiny{1}}; 
	 \filldraw[fill=red!60] (3,4) +(-0.5,-0.5) rectangle ++(0.5,0.5);
	 \draw [white] (3,4) node{\tiny{2}}; 
	 \filldraw[fill=white] (2,4) +(-0.5,-0.5) rectangle ++(0.5,0.5);
	 \draw [black] (2,4) node{\tiny{1}}; 
	 \filldraw[fill=white] (4,4) +(-0.5,-0.5) rectangle ++(0.5,0.5);
	 \draw [black] (4,4) node{\tiny{2}};  

	 \draw[-, thick, red] (0.5, 3.5) -- (1.5, 3.5) -- (1.5, 4.5) -- (0.5, 4.5) -- (0.5, 3.5);
	 \draw[-, thick, red] (3.5, 3.5) -- (3.5, 4.5) -- (4.5, 4.5) -- (4.5, 3.5) -- (3.5, 3.5);
	\end{tikzpicture}
	}\hspace{-2mm}
	\subfigure[move 9]{
	\tikzstyle{Point}=[circle,draw=black,fill=black,thick,
						inner sep=0pt,minimum size=2mm]
	\tikzstyle{arrow} = [thick,->,>=stealth]
	\tikzstyle{arrowonly} = [thick,-,>=stealth,line width = 2pt, color = cyan]
	\tikzstyle{arrowleft} = [thick,<-,>=stealth]
	\begin{tikzpicture}[auto, scale=0.3]
	 \foreach \x in {1,2,3,4}
	  \foreach \y in {1,...,4}
	  {
	   \filldraw[fill=cyan!30] (\x,\y) +(-0.5,-0.5) rectangle ++(0.5,0.5);
	  }
	 \filldraw[fill=red!60] (2,3) +(-0.5,-0.5) rectangle ++(0.5,0.5);
	 \draw [white] (2,3) node{\tiny{1}}; 
	 \filldraw[fill=red!60] (4,4) +(-0.5,-0.5) rectangle ++(0.5,0.5);
	 \draw [white] (4,4) node{\tiny{2}}; 
	 \filldraw[fill=white] (2,4) +(-0.5,-0.5) rectangle ++(0.5,0.5);
	 \draw [black] (2,4) node{\tiny{1}}; 
	 \filldraw[fill=white] (3,4) +(-0.5,-0.5) rectangle ++(0.5,0.5);
	 \draw [black] (3,4) node{\tiny{2}};  

	 \draw[-, thick, red] (0.5, 3.5) -- (1.5, 3.5) -- (1.5, 4.5) -- (0.5, 4.5) -- (0.5, 3.5);
	 \draw[-, thick, red] (3.5, 3.5) -- (3.5, 4.5) -- (4.5, 4.5) -- (4.5, 3.5) -- (3.5, 3.5);
	\end{tikzpicture}
	}\hspace{-2mm}
	\subfigure[move 10]{
	\tikzstyle{Point}=[circle,draw=black,fill=black,thick,
						inner sep=0pt,minimum size=2mm]
	\tikzstyle{arrow} = [thick,->,>=stealth]
	\tikzstyle{arrowonly} = [thick,-,>=stealth,line width = 2pt, color = cyan]
	\tikzstyle{arrowleft} = [thick,<-,>=stealth]
	\begin{tikzpicture}[auto, scale=0.3]
	 \foreach \x in {1,2,3,4}
	  \foreach \y in {1,...,4}
	  {
	   \filldraw[fill=cyan!30] (\x,\y) +(-0.5,-0.5) rectangle ++(0.5,0.5);
	  }
	 \filldraw[fill=red!60] (2,4) +(-0.5,-0.5) rectangle ++(0.5,0.5);
	 \draw [white] (2,4) node{\tiny{1}}; 
	 \filldraw[fill=red!60] (4,4) +(-0.5,-0.5) rectangle ++(0.5,0.5);
	 \draw [white] (4,4) node{\tiny{2}}; 
	 \filldraw[fill=white] (2,3) +(-0.5,-0.5) rectangle ++(0.5,0.5);
	 \draw [black] (2,3) node{\tiny{1}}; 
	 \filldraw[fill=white] (3,4) +(-0.5,-0.5) rectangle ++(0.5,0.5);
	 \draw [black] (3,4) node{\tiny{2}};  

	 \draw[-, thick, red] (0.5, 3.5) -- (1.5, 3.5) -- (1.5, 4.5) -- (0.5, 4.5) -- (0.5, 3.5);
	 \draw[-, thick, red] (3.5, 3.5) -- (3.5, 4.5) -- (4.5, 4.5) -- (4.5, 3.5) -- (3.5, 3.5);
	\end{tikzpicture}
	}\hspace{-2mm}
	\subfigure[move 11]{
	\tikzstyle{Point}=[circle,draw=black,fill=black,thick,
						inner sep=0pt,minimum size=2mm]
	\tikzstyle{arrow} = [thick,->,>=stealth]
	\tikzstyle{arrowonly} = [thick,-,>=stealth,line width = 2pt, color = cyan]
	\tikzstyle{arrowleft} = [thick,<-,>=stealth]
	\begin{tikzpicture}[auto, scale=0.3]
	 \foreach \x in {1,2,3,4}
	  \foreach \y in {1,...,4}
	  {
	   \filldraw[fill=cyan!30] (\x,\y) +(-0.5,-0.5) rectangle ++(0.5,0.5);
	  }
	 \filldraw[fill=red!60] (2,4) +(-0.5,-0.5) rectangle ++(0.5,0.5);
	 \draw [white] (2,4) node{\tiny{1}}; 
	 \filldraw[fill=red!60] (4,4) +(-0.5,-0.5) rectangle ++(0.5,0.5);
	 \draw [white] (4,4) node{\tiny{2}}; 
	 \filldraw[fill=white] (1,3) +(-0.5,-0.5) rectangle ++(0.5,0.5);
	 \draw [black] (1,3) node{\tiny{1}}; 
	 \filldraw[fill=white] (3,4) +(-0.5,-0.5) rectangle ++(0.5,0.5);
	 \draw [black] (3,4) node{\tiny{2}};   

	 \draw[-, thick, red] (0.5, 3.5) -- (1.5, 3.5) -- (1.5, 4.5) -- (0.5, 4.5) -- (0.5, 3.5);
	 \draw[-, thick, red] (3.5, 3.5) -- (3.5, 4.5) -- (4.5, 4.5) -- (4.5, 3.5) -- (3.5, 3.5);
	\end{tikzpicture}
	}\hspace{-2mm}
	\subfigure[move 12]{
	\tikzstyle{Point}=[circle,draw=black,fill=black,thick,
						inner sep=0pt,minimum size=2mm]
	\tikzstyle{arrow} = [thick,->,>=stealth]
	\tikzstyle{arrowonly} = [thick,-,>=stealth,line width = 2pt, color = cyan]
	\tikzstyle{arrowleft} = [thick,<-,>=stealth]
	\begin{tikzpicture}[auto, scale=0.3]
	 \foreach \x in {1,2,3,4}
	  \foreach \y in {1,...,4}
	  {
	   \filldraw[fill=cyan!30] (\x,\y) +(-0.5,-0.5) rectangle ++(0.5,0.5);
	  }
	 \filldraw[fill=red!60] (2,4) +(-0.5,-0.5) rectangle ++(0.5,0.5);
	 \draw [white] (2,4) node{\tiny{1}}; 
	 \filldraw[fill=red!60] (4,4) +(-0.5,-0.5) rectangle ++(0.5,0.5);
	 \draw [white] (4,4) node{\tiny{2}}; 
	 \filldraw[fill=white] (1,4) +(-0.5,-0.5) rectangle ++(0.5,0.5);
	 \draw [black] (1,4) node{\tiny{1}}; 
	 \filldraw[fill=white] (3,4) +(-0.5,-0.5) rectangle ++(0.5,0.5);
	 \draw [black] (3,4) node{\tiny{2}};  

	 \draw[-, thick, red] (0.5, 3.5) -- (1.5, 3.5) -- (1.5, 4.5) -- (0.5, 4.5) -- (0.5, 3.5);
	 \draw[-, thick, red] (3.5, 3.5) -- (3.5, 4.5) -- (4.5, 4.5) -- (4.5, 3.5) -- (3.5, 3.5);
	\end{tikzpicture}
	}\hspace{-2mm}
	\subfigure[move 13]{
	\tikzstyle{Point}=[circle,draw=black,fill=black,thick,
						inner sep=0pt,minimum size=2mm]
	\tikzstyle{arrow} = [thick,->,>=stealth]
	\tikzstyle{arrowonly} = [thick,-,>=stealth,line width = 2pt, color = cyan]
	\tikzstyle{arrowleft} = [thick,<-,>=stealth]
	\begin{tikzpicture}[auto, scale=0.3]
	 \foreach \x in {1,2,3,4}
	  \foreach \y in {1,...,4}
	  {
	   \filldraw[fill=cyan!30] (\x,\y) +(-0.5,-0.5) rectangle ++(0.5,0.5);
	  }
	 \filldraw[fill=red!60] (1,4) +(-0.5,-0.5) rectangle ++(0.5,0.5);
	 \draw [white] (1,4) node{\tiny{1}}; 
	 \filldraw[fill=red!60] (4,4) +(-0.5,-0.5) rectangle ++(0.5,0.5);
	 \draw [white] (4,4) node{\tiny{2}}; 
	 \filldraw[fill=white] (2,4) +(-0.5,-0.5) rectangle ++(0.5,0.5);
	 \draw [black] (2,4) node{\tiny{1}}; 
	 \filldraw[fill=white] (3,4) +(-0.5,-0.5) rectangle ++(0.5,0.5);
	 \draw [black] (3,4) node{\tiny{2}};  

	 \draw[-, thick, red] (0.5, 3.5) -- (1.5, 3.5) -- (1.5, 4.5) -- (0.5, 4.5) -- (0.5, 3.5);
	 \draw[-, thick, red] (3.5, 3.5) -- (3.5, 4.5) -- (4.5, 4.5) -- (4.5, 3.5) -- (3.5, 3.5);
	\end{tikzpicture}
	}

	\caption{Optimal path when retrieving jointly}
	\label{fig:optimal path}
\end{figure}
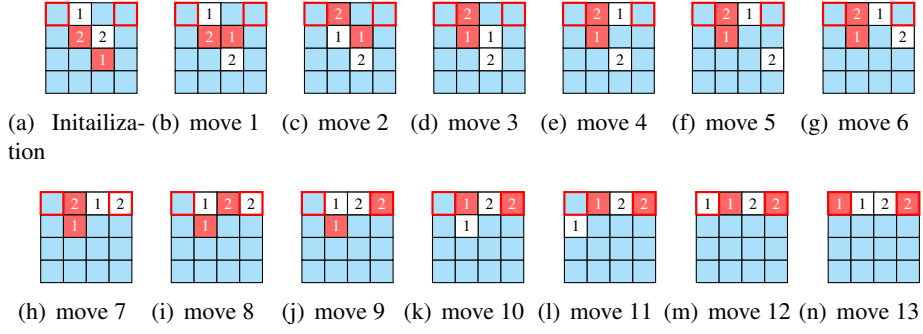

\section{Integer Programming Formulation }
\label{sec:IPformulationnew}

In this section, we provide a generic IP formulation to cover a variety of problem settings, including (i) multiple desired items, multiple escorts, and multiple I/O points are all placed randomly, (ii) the number of desired items/escorts can be set to any feasible values. 
{\color{black}
In what follows, we first present brief descriptions about state transition that motivate us to design decision variables properly and then present the complete mathematical formulation.
}

Suppose a grid of size $m \times n$ with $e$ escorts and the position ID of each cell is denoted by $p, p=0, 1,\cdots, mn - 1$. There are $d$ desired items in the grid, which are {\color{black}indexed by $r$.}
{\color{black}
Intuitively, the state of the grid at each timestamp, denoted as \textit{occupancy state}, is determined by whether each cell is occupied and the positions of the desired items. 
To better illustrate \textit{occupancy state} and its changes, we introduce following four groups of binary variables:
\begin{itemize}
\item $x_{p}^{k}$: if position $p$ is occupied after move $k$, then $x_{p}^{k}=1$; otherwise, $x_{p}^{k} = 0$.
\item $y_{p,r}^{k}$: if position $p$ is occupied by desired item $r$ after move $k$, then $y_{p,r}^{k}=1$; otherwise, $y_{p,r}^{k} = 0$.
\item $z_{p,q}^{k}$: if an item moves from position $p$ to $q$ at move $k$, then $z_{p,q}^{k} = 1$; otherwise, $z_{p,q}^{k} = 0$. 
\item $w_{p,q,r}^{k}$: if desired item $r$ occupies position $p$ and moves from position $p$ to $q$ at move $k$, then $w_{p,q,r}^{k} = 1$; otherwise, $w_{p,q,r}^{k} = 0$.
\end{itemize}
}

As shown in Figure \ref{fig:movestateSeqNew}, we move desired item 1 from position 8 to 7 during move $k$, and the occupancy state changed accordingly. 
The whole retrieval process can be interpreted as a sequence of changing occupancy states, i.e., the process starts with an initial occupancy state and terminates when and only when all desired items reach their pre-assigned I/O points.

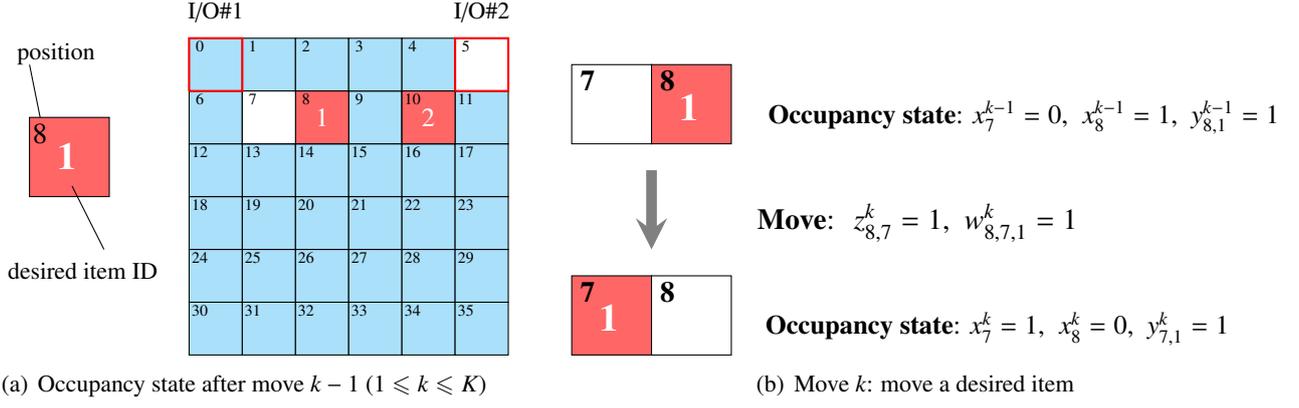
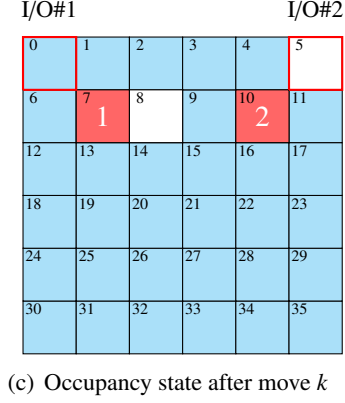
\begin{figure}[H]
\centering
\hspace{-7mm}
	\subfigure[Occupancy state after move $k-1$ ($1 \leqslant k \leqslant K$)]{
	\label{fig:beginningState}
	\tikzstyle{Point}=[circle,draw=black,fill=black,thick,
						inner sep=0pt,minimum size=2mm]
	\tikzstyle{arrow} = [thick,->,>=stealth]
	\tikzstyle{arrowonly} = [thick,-,>=stealth,line width = 2pt, color = cyan]
	\tikzstyle{arrowleft} = [thick,<-,>=stealth]
	\begin{tikzpicture}[auto, scale=0.7]
	 \foreach \x in {1,2,3,4,5,6}
	  \foreach \y in {1,...,6}
	  {
	   \filldraw[fill=cyan!30] (\x,\y) +(-0.5,-0.5) rectangle ++(0.5,0.5);
	  }
	 \filldraw[fill=red!60] (3,5) +(-0.5,-0.5) rectangle ++(0.5,0.5);
	 \draw [white] (3,5) node{\small 1}; 
	 \filldraw[fill=red!60] (5,5) +(-0.5,-0.5) rectangle ++(0.5,0.5);
	 \draw [white] (5,5) node{\small 2}; 
	 \filldraw[fill=white] (2,5) +(-0.5,-0.5) rectangle ++(0.5,0.5);
	 \draw [black] (3,6) node{}; 
	 \filldraw[fill=white] (6,6) +(-0.5,-0.5) rectangle ++(0.5,0.5);
	 \draw [black] (6,6) node{}; 
	 \draw[-, thick, red] (0.5, 5.5) -- (0.5, 6.5) -- (1.5, 6.5) -- (1.5, 5.5) -- (0.5, 5.5);
	 \draw[-, thick, red] (5.5, 5.5) -- (5.5, 6.5) -- (6.5, 6.5) -- (6.5, 5.5) -- (5.5, 5.5);
	 \draw (1, 7) node(name=I/0){\footnotesize{I/O\#1}};
	 \draw (6, 7) node(name=I/0){\footnotesize{I/O\#2}};
	  \foreach \x/\y in {1/6,2/6,3/6,4/6,5/6,6/6,
	  1/5,2/5,3/5,4/5,5/5,6/5,
	  1/4,2/4,3/4,4/4,5/4,6/4,
	  1/3,2/3,3/3,4/3,5/3,6/3,
	  1/2,2/2,3/2,4/2,5/2,6/2,
	  1/1,2/1,3/1,4/1,5/1,6/1} {
	  	\draw (\x-0.3,7-\y+0.35) node{\tiny \color{black} \pgfmathparse{int(\x + 6*(\y -1) - 1)}\pgfmathresult};
	  }
	  \foreach \x/\y in {1/6} {
	    \filldraw[fill=red!60,xshift=-3cm,yshift=-2cm] (\x,\y) +(-0.5,-0.5) rectangle ++(1,1);
	  	\draw [xshift=-2cm,yshift=-2cm] (\x-1.3,\y+0.7) node{ \color{black} 8}; 
	      \draw [xshift=-2cm,yshift=-2cm] (\x-0.8,\y+0.275) node{\color{white} \textbf{\Large 1}}; 
	  }	
	 \draw[thin, black] (-2.3, 5) -- (-2.5, 6);
	 \draw (-2, 6.2) node(name=example){\footnotesize{position}};
	 \draw[thin, black] (-1.7, 3.7) -- (-1.1, 2.5);
	 \draw (-1.5, 2.1) node(name=example){\footnotesize{desired item ID}};

	\end{tikzpicture}
	} 
	\subfigure[Move $k$: move a desired item]{
	\label{fig:move}
	\tikzstyle{Point}=[circle,draw=black,fill=black,thick,
						inner sep=0pt,minimum size=2mm]
	\tikzstyle{arrow} = [thick,->,>=stealth, line width = 4pt]
	\tikzstyle{arrowonly} = [thick,-,>=stealth,line width = 2pt, color = black]
	\tikzstyle{arrowleft} = [thick,<-,>=stealth]
	\begin{tikzpicture}[auto, scale=0.7]
	 \filldraw[fill=white] (0,0) +(-0.5,-0.5) rectangle ++(1,1);
	 \draw (0-0.2,0+0.7) node{ \color{black} \textbf{7}}; 
	 \filldraw[fill=red!60,xshift=1.5cm] (0,0) +(-0.5,-0.5) rectangle ++(1,1);
	 \draw[xshift=1.5cm] (0-0.2,0+0.7) node{ \color{black} \textbf{8}};
	 \draw [white] (1.7,0.2) node{\textbf{\Large 1}}; 
	 \draw (8, 0) node{\small \color{black} \textbf{Occupancy state}: $x_{7}^{k-1} = 0, \,\,x_{8}^{k-1} = 1, \,\, y_{8, 1}^{k-1} = 1$}; 	 
	 \draw[arrow, gray] (1, -1) -- (1, -2.5);
	 \draw (6, -2) node{ \color{black} \textbf{Move}:  $ \,\,z_{8, 7}^{k} = 1, \,\, w_{8, 7, 1}^{k} = 1$}; 	 
	 \filldraw[fill=red!60,yshift=-4cm] (0,0) +(-0.5,-0.5) rectangle ++(1,1);
	 \filldraw[fill=white,xshift=1.5cm, yshift=-4cm] (0,0) +(-0.5,-0.5) rectangle ++(1,1);
	 \draw[yshift=-4cm] (0-0.2,0+0.7) node{ \color{black} \textbf{7}}; 
	 \draw [white] (0.2, -3.8) node{\textbf{\Large 1}}; 
	 \draw[xshift=1.5cm, yshift=-4cm] (0-0.2,0+0.7) node{ \color{black} \textbf{8}}; 
	 \draw (7.5, -4) node{\small  \color{black} \textbf{Occupancy state}:  $x_{7}^{k} = 1, \,\,x_{8}^{k}= 0, \,\, y_{7, 1}^{k} = 1$}; 	 
	\end{tikzpicture}
	}
	\subfigure[Occupancy state after move $k$]{
	\label{fig:resultedState}
	\tikzstyle{Point}=[circle,draw=black,fill=black,thick,
						inner sep=0pt,minimum size=2mm]
	\tikzstyle{arrow} = [thick,->,>=stealth]
	\tikzstyle{arrowonly} = [thick,-,>=stealth,line width = 2pt, color = cyan]
	\tikzstyle{arrowleft} = [thick,<-,>=stealth]
	\begin{tikzpicture}[auto, scale=0.7]
	 \foreach \x in {1,2,3,4,5,6}
	  \foreach \y in {1,...,6}
	  {
	   \filldraw[fill=cyan!30] (\x,\y) +(-0.5,-0.5) rectangle ++(0.5,0.5);
	  }
	 \filldraw[fill=red!60] (2,5) +(-0.5,-0.5) rectangle ++(0.5,0.5);
	 \draw [white] (2,5) node{1}; 
	 \filldraw[fill=red!60] (5,5) +(-0.5,-0.5) rectangle ++(0.5,0.5);
	 \draw [white] (5,5) node{2}; 
	 \filldraw[fill=white] (3,5) +(-0.5,-0.5) rectangle ++(0.5,0.5);
	 \draw [black] (3,5) node{}; 
	 \filldraw[fill=white] (6,6) +(-0.5,-0.5) rectangle ++(0.5,0.5);
	 \draw [black] (6,6) node{}; 
	 \draw[-, thick, red] (0.5, 5.5) -- (0.5, 6.5) -- (1.5, 6.5) -- (1.5, 5.5) -- (0.5, 5.5);
	 \draw[-, thick, red] (5.5, 5.5) -- (5.5, 6.5) -- (6.5, 6.5) -- (6.5, 5.5) -- (5.5, 5.5);
	 \draw (1, 7) node(name=I/0){\footnotesize{I/O\#1}};
	 \draw (6, 7) node(name=I/0){\footnotesize{I/O\#2}};
	  \foreach \x/\y in {1/6,2/6,3/6,4/6,5/6,6/6,
	  1/5,2/5,3/5,4/5,5/5,6/5,
	  1/4,2/4,3/4,4/4,5/4,6/4,
	  1/3,2/3,3/3,4/3,5/3,6/3,
	  1/2,2/2,3/2,4/2,5/2,6/2,
	  1/1,2/1,3/1,4/1,5/1,6/1} {
	  	\draw (\x-0.3,7-\y+0.35) node{\tiny \color{black} \pgfmathparse{int(\x + 6*(\y -1) - 1)}\pgfmathresult};
	  }

	\end{tikzpicture}
	}
	\caption{Move-occupancy state sequence}
	\label{fig:movestateSeqNew}
\end{figure}

Based on the above observations, we attempt to formulate the problem in the form of IP. In order to make the formulation tractable, we introduce an input parameter, the maximum number of moves $K$, which implies that the retrieval task must be completed in at most $K$ moves. $K$ can be set to a sufficiently large number or can be estimated by a regression model (see Section \ref{sec:regressionModel} for details).
{\color{black}And we use a parameter $A_{p, q}$ to represent if the position $p$ and $q$ are adjacent, i.e., the legality of the move.} If it is possible to move from 
position $p$ to $q$, then $A_{p, q}  = 1$, otherwise $A_{p, q}= 0$. 
{\color{black}Let $\mathcal{P}$ be the set of all the positions and $\mathcal{D}$ the set of all desired items, where $\mathcal{P} = \{0, 1, 2, \cdots, mn-1 \}, \mathcal{D}= \{1, \cdots, d\}$.
The multi-item retrieval problem can be formulated as follows:

\vspace{-1cm}
\begin{align}
\min \quad  &\sum_{k=0}^{K}{\sum_{p \in \mathcal{P}}{\sum_{q \in \mathcal{P}}{z_{p,q}^{k}}}}  \label{eq:newobj}&&
\\
s.t. \,\, &x_{p}^{0}=\left\{
\begin{matrix}1,&\text{if\,\,position\,\,$p$\,\,is\,\,occupied\,\,initially}\\0,&\text{otherwise}
\\
\end{matrix} \right. && \forall p\in \mathcal{P},
\label{eq:new_c1}
\\
&y_{p,r}^{0}=\left\{ \begin{matrix}1,&\text{if\,\,desired\,\,item\,\,$r$\,\,is\,\,at\,\,position\,\,$p$\,\,initially}\\
0,&\text{otherwise}\\
\end{matrix} \right. && \forall p\in \mathcal{P},\forall r\in \mathcal{D},  \label{eq:new_c2}
\\
&z_{p,q}^{0}=0  &&\forall p, q \in \mathcal{P},\label{eq:new_c3}
\\
&\sum_{p\in \mathcal{P}}{x_{p}^{k}}=|\mathcal{P}|-e && \forall k=0,1,\cdots, K,  \label{eq:new_c4}
\\
&x_{p}^{k-1}-\sum_{q\in \mathcal{P}}{z_{p,q}^{k}}+\sum_{q\in \mathcal{P}}{z_{q,p}^{k}}=x_{p}^{k} && \forall k=1,2,\cdots, K,\forall p\in \mathcal{P},  \label{eq:new_c5}
\\
&y_{p,r}^{k}\leqslant x_{p}^{k} && \forall k=0,1,\cdots, K,\forall p\in \mathcal{P},\forall r\in \mathcal{D},  \label{eq:new_c6}
\\
&\sum_{p \in \mathcal{P}}{y_{p,r}^{k}}=1 &&\forall k=0,1,\cdots, K,  \forall r\in \mathcal{D},  \label{eq:new_c7}
\\
&\sum_{r \in \mathcal{D}}{y_{p,r}^{k}}\leqslant 1 && \forall k\in k=0,1,\cdots, K,\forall p\in \mathcal{P},  \label{eq:new_c8}
\\
&y_{p,r}^{k-1}-\sum_{q \in \mathcal{P}}{w_{p,q, r}^{k}}+\sum_{q \in \mathcal{P}}{w_{q,p,r}^{k}}=y_{p,r}^{k}\,\, && \forall k=1,2,\cdots, K,\forall p\in \mathcal{P},\forall r\in \mathcal{D},  \label{eq:new_c9}
\\
&y_{IO_r,r}^{K}=1 && \forall r\in \mathcal{D},  \label{eq:new_c10}
\\
&\sum_{k=1}^{K}z_{p,q}^{k}\leqslant A_{p,q}K && \forall p,q\in \mathcal{P}, \label{eq:new_c11}
\\
&z_{p,q}^{k}\leqslant 1-x_{q}^{k-1} && \forall k=1,2,\cdots, K,\forall p,q\in \mathcal{P},  \label{eq:new_c12}
\\
&\sum_{p \in \mathcal{P}}{\sum_{q \in \mathcal{P}}{z_{p,q}^{k}}}\leqslant 1 && \forall k=0,1,\cdots, K,   \label{eq:new_c13}
\\
&\sum_{r \in \mathcal{D}}{w_{p,q,r}^{k}}\leqslant z_{p,q}^{k}   && \forall k=0,1,\cdots, K,\forall p,q\in \mathcal{P}, \label{eq:new_c14}
\\
&\text{if } y_{p,r}^{k-1}=1 \text{ and } z_{p,q}^{k}=1, \text{ then }  w_{p,q, r}^{k}=1   && \forall k=0,1,\cdots, K,\forall p,q\in \mathcal{P},\forall r\in \mathcal{D},  \label{eq:new_c15}
\\
&
x_{p}^{k}, y_{p,r}^{k}, w_{q,p,r}^{k} \in \{0, 1 \} && \forall k=0,1,\cdots, K,\forall p, q \in \mathcal{P},\forall r\in \mathcal{D}.  \label{eq:new_c16}
\end{align}

The objective function equation (\ref{eq:newobj}) is to minimize the total number of moves. Constraints (\ref{eq:new_c1})-(\ref{eq:new_c3}) initialize the value of 
the decision variables at step 0.
Constraints (\ref{eq:new_c4}) restrict that the number of occupied positions is equal to the number of items (includes desired items and normal items) at every step. 
Constraints (\ref{eq:new_c5}) denote the occupancy state transition between move $k-1$ and $k$.
Constraints (\ref{eq:new_c6}) imply that desired item $r$ can not locate at position $p$
if $p$ is an escort. Constraints (\ref{eq:new_c7}) and 
(\ref{eq:new_c8}) guarantee that each desired item must locate at one position at each step, and each position can be occupied by at most one desired item as well. 
Similar to constraints (\ref{eq:new_c5}), constraints (\ref{eq:new_c9}) denote the occupancy state transition of each desired item between move $k-1$ and $k$. 
Constraints (\ref{eq:new_c10}) ensure that all desired items reach the corresponding I/O points at the final move.
Constraints (\ref{eq:new_c11})-(\ref{eq:new_c15}) are move relevant constraints.  
Constraints (\ref{eq:new_c11}) restrict that an item can only move to its neighbor positions over $K$ timestamps. Constraints (\ref{eq:new_c12}) imply that a move is allowed if the targeted position is an escort. Constraints (\ref{eq:new_c13}) ensure that at most one move is allowed at each step. Constraints (\ref{eq:new_c14}) impose that the desired items can move from position $p$ to $q$ only if an action from position $p$ to $q$ is executed. Constraints (\ref{eq:new_c15}) clarify the relationship among $y_{p,r}^{k-1}$, $z_{p,q}^{k}$ and $w_{p,q, r}^{k}$ to specify the move condition of the desired items, and they can be easily implemented by existing commerical optimization solvers (e.g., Gurobi). 
Finally, the variables' type is defined by constraints (\ref{eq:new_c16}).

We note that this IP model is quite compact compared to the IP model in \cite{Rohit2010Retrieval} which is dedicated to single-item retrieval problems. Specifically, we remove subscript $i$ from decision variables $x$ and $y$ of the previous IP model and add variables $z$ and $w$. As a result, the number of decision variables and constraints is significantly reduced. 

Even though the proposed IP formulation is based on the single-movement assumption, it can be easily extended to simultaneous movement schemes by making the following three revisions: (i) remove constraints (\ref{eq:new_c13}), (ii) introduce binary decision variables $f_k$ to indicate whether any move is executed at timestamp $k$, and add one more group of constraints to ensure the relationship between $f_k$ and $z_{p,q}^{k}$ (more specifically, the detailed equations are $f_{k} \geqslant z_{p,q}^{k}, \forall p, q \in \mathcal{P}, \forall k = 0,1, \cdots, K$), and (iii) change the objective function to $\min \sum_{k=1}^{K}{f_k}$. 
}

\section{Reinforcement Learning Formulation}
\label{sec:RLformulation}

We characterize this research issue as a sequential decision-making problem and structure it as a Markov Decision Process based on its characteristics. The proposed model is then solved via reinforcement learning, using a well-designed environment, dueling DQN, and double DQN for the PBS system.
Finally, rather than using traditional random exploration, we develop a semi-random and semi-guided action selection mechanism to accelerate convergence. We will offer a detailed description of the proposed techniques in this section.

\subsection{Environment}

An environment serves as the foundation for training a reinforcement learning model, in which an agent interacts with the environment and learns from its experiences. The agent receives specific feedback (rewards) based on the current state of the environment and the actions performed. If the agent receives a positive reward, the appearance of the action will be enhanced; otherwise, the appearance of the action will be weakened. Through the learning procedure, {\color{black}the agent tends to pick the best action (not 100\%, given the effects of possible noise, exploration, possibility to converge to a suboptimal policy, etc.), i.e. the one with the highest cumulative reward.}

The iterative process of reinforcement learning for the PBS system is described as follows. To begin, prepare the environment by placing $d$ desired items and $e$ escorts in random order. Then perform a move, update the environment, and receive the corresponding reward. As shown in Figure \ref{fig:Types of moves}, the environment changes when a normal move is performed, and it remains unchanged with an illegal move (when the escort moves out of the grid). Continue until all desired items have arrived at their I/O points, then terminate the round and reset the environment.

\begin{figure}[H]
\centering
\hspace{-7mm}
	\subfigure[normal move]{
	\label{fig:normal move}
	\tikzstyle{Point}=[circle,draw=red!60,fill=red!60,thick,
						inner sep=0pt,minimum size=2mm]
	\tikzstyle{arrow} = [thick,->,>=stealth]
	\tikzstyle{arrowonly} = [thick,-,>=stealth,line width = 2pt, color = cyan]
	\tikzstyle{arrowleft} = [thick,<-,>=stealth]
	\begin{tikzpicture}[auto, scale=0.5]
	 \foreach \x in {1,2,3,4}
	  \foreach \y in {1,...,4}
	  {
	   \filldraw[fill=cyan!30] (\x,\y) +(-0.5,-0.5) rectangle ++(0.5,0.5);
	  }
	 \filldraw[fill=red!60] (4,3) +(-0.5,-0.5) rectangle ++(0.5,0.5);
	 \filldraw[fill=white] (3,2) +(-0.5,-0.5) rectangle ++(0.5,0.5);
	 \draw[-, thick, red] (0.5, 3.5) -- (0.5, 4.5) -- (1.5, 4.5) -- (1.5, 3.5) -- (0.5, 3.5);
	 \draw[->, thick, red] (3, 2) -- (3, 3);
	\end{tikzpicture}
	}
	\subfigure[illegal move]{
	\label{fig:illegal move}
	\tikzstyle{Point}=[circle,draw=red!60,fill=red!60,thick,
						inner sep=0pt,minimum size=2mm]
	\tikzstyle{arrow} = [thick,->,>=stealth]
	\tikzstyle{arrowonly} = [thick,-,>=stealth,line width = 2pt, color = cyan]
	\tikzstyle{arrowleft} = [thick,<-,>=stealth]
	\begin{tikzpicture}[auto, scale=0.5]
	 \foreach \x in {1,2,3,4}
	  \foreach \y in {1,...,4}
	  {
	   \filldraw[fill=cyan!30] (\x,\y) +(-0.5,-0.5) rectangle ++(0.5,0.5);
	  }
	 \filldraw[fill=red!60] (4,3) +(-0.5,-0.5) rectangle ++(0.5,0.5);
	 \filldraw[fill=white] (3,4) +(-0.5,-0.5) rectangle ++(0.5,0.5);
	 \draw[-, thick, red] (0.5, 3.5) -- (0.5, 4.5) -- (1.5, 4.5) -- (1.5, 3.5) -- (0.5, 3.5);
	 \draw[->, thick, red] (3, 4) -- (3, 5);
	\end{tikzpicture}
	}
	\caption{Types of moves}
	\label{fig:Types of moves}
\end{figure}
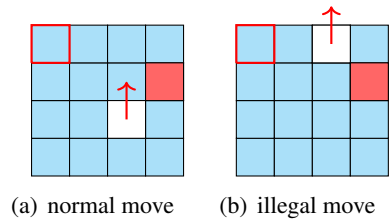

This section will focus on creating an environment suitable for the PBS system, including state, action, and reward.

\textbf{State}
To continue the learning procedure, a reinforcement learning agent needs environment-relevant information, which implies the environment should feedback state after each step. Rather than using local data to generate a distributed decision (\cite{mukhutdinov2019multi}), the state in this paper should reflect information about the entire system. As a result, the state is represented by the coordinates of the desired items and escorts. If there are $d$ desired items, $e$ escorts, and $d$ I/O points in a $m\times n$ grid, the size of the state space is
\begin{align}
A_{m\times n}^{d+e}-A_{m\times n-d}^{e} \label{eq:state_space}
\end{align}
where $A_{m\times n}^{d+e} = \frac{(m \times n)!}{(m \times n - d - e)!}$ and $A_{m\times n-d}^{e} = \frac{(m\times n-d)!}{(m\times n - d - e)!}$, standing for the number of all permutations of the corresponding setting. {\color{black}The second subtracted term represents the cases where the desired items are already at the corresponding I/O points and should be excluded.}

\textbf{Action}
Each escort has four possible actions {\color{black}correspongding to moving up, down, left, and right}, and hence there are $4e$ optional actions in total. 

\textbf{Reward}
{\color{black}The environment provides a positive reward (e.g., 1)} if the desired items reach their I/O points; otherwise, the reward is zero. This encourages the agent to discover the termination as soon as possible. In fact, before implementing the semi-random and semi-guided action selection mechanism, we tried to design a reward rule to tackle the sparse reward problem, but it was too sophisticated to guarantee the global optimum. Therefore, sophisticated incentive design is no longer an option.

\subsection{Semi-random and semi-guided action selection mechanism}\label{Semi-random and semi-guided action selection mechanism}

{\color{black}In reinforcement learning, it is essential to strike a balance between exploration and exploitation when selecting actions. Traditionally, pick actions randomly to explore the environment; relatively, choose actions based on the predicted action values from the neural network to exploit the stored information. With exploitation remaining unchanged, we find the random action selection for exploration} is not suitable for this problem as it is difficult to reach the final state due to the large state space and the non-directional escort motions. The experiments indicate that when the grid size is $6\times 6$ and there are 2 desired items and 2 escorts, it generally takes tens of thousands of moves to reach the final state; that is, there are only a few meaningful pieces among the tens of thousands of experiences. Sparse reward means that the agent will not receive enough effective feedbacks, which is detrimental to learning. {\color{black}To increase the rate of receiving non-zero rewards, we introduce a guiding method to help the agent choose reasonable actions and reach the final state quickly. Therefore, a semi-random and semi-guided action selection mechanism is designed to replace the random one.} 

Semi-random and semi-guided action selection mechanism implies that the chance of adopting a random action selection mechanism is {\color{black}$1-\eta$} and the likelihood of using a guided action selection mechanism is {\color{black}$\eta$}, where {\color{black}$\eta$} is a parameter and can take any value ranging from 0 to 1. During the early stages of training, the neural network has very limited knowledge. Thus, a large {\color{black}$\eta$} should be used to aid the agents in locating the final state more quickly. In the later stages, as the neural network matures and stores more data, the value of {\color{black}$\eta$} should be decreased to encourage the agent to explore new information.

\subsubsection{Random action selection mechanism}
It is self-evident that the random action selection mechanism chooses one action randomly among the available actions. This is not a subject that we address extensively here. 

\subsubsection{Guided action selection mechanism}
\label{sec:usefulPoint}
As the name implies, the guided action selection mechanism is a heuristic algorithm that guides escorts to move effectively to enable the agent to complete the task sooner. 
{\color{black}Note that exploration is inherently part of reinforcement learning, thus this approach does not need to be a complete heuristic algorithm as long as it is instructive.}
There are two essential evaluation indices: the distance between the desired item and the corresponding I/O point $d_{item\_manhattan}$ and the distance between the escort and its useful point (which may be considered as the I/O point for the escort) $d_{escort\_manhattan}$. Let $\left(x_p,y_p \right)$ be the location of the desired item, $\left(x_{io},y_{io}\right)$ the location of its corresponding I/O point. 

Thus, the distance between the desired item and the corresponding I/O point is
\begin{align}
d_{item\_manhattan}=\left| x_{io}-x_p \right|+\left| y_{io}-y_p \right|
\end{align}

Likewise, let $\left(x_{escort},y_{escort}\right)$ be the location of the escort, $\left(x_{useful},y_{useful}\right)$ the location of its useful point. Points that satisfy the following two conditions are useful. (i) The point is adjacent to the desired item, at which  $|x_{useful}-x_{p}|+|y_{useful}-y_{p}|=1$ holds.
(ii) $d_{item\_manhattan}$ decreases if the desired item reaches this point, i.e., $\left| x_{io}-x_{p} \right|+\left| y_{io}-y_{p} \right|>\left| x_{io}-x_{useful} \right|+\left| y_{io}-y_{useful} \right|$. 
Thus, the distance between the escort and its useful point is 
\begin{align}
d_{escort\_manhattan}=|x_{useful}-x_{escort}|+|y_{useful}-y_{escort}|+2\times \theta
\end{align}
{\color{black}where $\theta$ is a binary indicator of whether the desired item is between the useful point and the escort.}

Take Figure \ref{fig:different theta}\subref{fig:theta=0} as an example, the desired item [1,1] is not between the useful point [0,1] and the escort [2,2], thus $\theta=0$ and $d_{escort\_manhattan}=|0-2|+|1-2|+2\times 0=3$.
In Figure \ref{fig:different theta}\subref{fig:theta=1}, the desired item [1,1] is located between the useful point [0,1] and the escort [2,1], thus $\theta=1$ and $d_{escort\_manhattan}=|0-2|+|1-1|+2\times 1=4$.

\begin{figure}[H]
\centering
\hspace{-7mm}
	\subfigure[$\theta=0$]{
	\label{fig:theta=0}
	\tikzstyle{Point}=[circle,draw=red!60,fill=red!60,thick,
						inner sep=0pt,minimum size=2mm]
	\tikzstyle{arrow} = [thick,->,>=stealth]
	\tikzstyle{arrowonly} = [thick,-,>=stealth,line width = 2pt, color = cyan]
	\tikzstyle{arrowleft} = [thick,<-,>=stealth]
	\begin{tikzpicture}[auto, scale=0.5]
	 \foreach \x in {1,2,3,4}
	  \foreach \y in {1,...,4}
	  {
	   \filldraw[fill=cyan!30] (\x,\y) +(-0.5,-0.5) rectangle ++(0.5,0.5);
	  }
	 \filldraw[fill=red!60] (2,3) +(-0.5,-0.5) rectangle ++(0.5,0.5);
	 \draw [white] (2,3) node{1}; 
	 \filldraw[fill=white] (3,2) +(-0.5,-0.5) rectangle ++(0.5,0.5);
	 \draw [red!60] (3,2) node{1}; 

	 \filldraw[fill=green] (2,4) +(-0.5,-0.5) rectangle ++(0.5,0.5);
	 \filldraw[fill=green] (1,3) +(-0.5,-0.5) rectangle ++(0.5,0.5);

	 \draw[-, thick, red] (0.5, 3.5) -- (0.5, 4.5) -- (1.5, 4.5) -- (1.5, 3.5) -- (0.5, 3.5);
	 \draw[->, thick, red] (3, 2.5) -- (3, 4) -- (2, 4);
	\end{tikzpicture}
	}
	\subfigure[$\theta=1$]{
	\label{fig:theta=1}
	\tikzstyle{Point}=[circle,draw=red!60,fill=red!60,thick,
						inner sep=0pt,minimum size=2mm]
	\tikzstyle{arrow} = [thick,->,>=stealth]
	\tikzstyle{arrowonly} = [thick,-,>=stealth,line width = 2pt, color = cyan]
	\tikzstyle{arrowleft} = [thick,<-,>=stealth]
	\begin{tikzpicture}[auto, scale=0.5]
	 \foreach \x in {1,2,3,4}
	  \foreach \y in {1,...,4}
	  {
	   \filldraw[fill=cyan!30] (\x,\y) +(-0.5,-0.5) rectangle ++(0.5,0.5);
	  }
	 \filldraw[fill=red!60] (2,3) +(-0.5,-0.5) rectangle ++(0.5,0.5);
	 \draw [white] (2,3) node{1}; 
	 \filldraw[fill=white] (2,2) +(-0.5,-0.5) rectangle ++(0.5,0.5);
	 \draw [red!60] (2,2) node{1}; 

	 \filldraw[fill=green] (2,4) +(-0.5,-0.5) rectangle ++(0.5,0.5);
	 \filldraw[fill=green] (1,3) +(-0.5,-0.5) rectangle ++(0.5,0.5);

	 \draw[-, thick, red] (0.5, 3.5) -- (0.5, 4.5) -- (1.5, 4.5) -- (1.5, 3.5) -- (0.5, 3.5);
	 \draw[->, thick, red] (2.5, 2) -- (3,2) -- (3, 4) -- (2, 4);

	\end{tikzpicture}
	}
	\subfigure{
	\label{fig:tuli}
	\tikzstyle{Point}=[circle,draw=red!60,fill=red!60,thick,
						inner sep=0pt,minimum size=2mm]
	\tikzstyle{arrow} = [thick,->,>=stealth]
	\tikzstyle{arrowonly} = [thick,-,>=stealth,line width = 2pt, color = cyan]
	\tikzstyle{arrowleft} = [thick,<-,>=stealth]
	\begin{tikzpicture}[auto, scale=0.5]

	 \filldraw[fill=red!60] (1,3.5) +(-0.5,-0.5) rectangle ++(0.5,0.5);
	 \draw [black][right] (1.75,3.5) node{\footnotesize{desired item}}; 
	 \filldraw[fill=white] (1,2.25) +(-0.5,-0.5) rectangle ++(0.5,0.5);
	 \draw [black][right] (1.75,2.25) node{\footnotesize{escort}}; 
	 \filldraw[fill=green] (1,1) +(-0.5,-0.5) rectangle ++(0.5,0.5);
	 \draw [black][right] (1.75,1) node{\footnotesize{useful point}}; 

	 \filldraw[fill=white] (1,0.25) +(-0.25,-0.25);

	\end{tikzpicture}
	}
	\caption{Example of different $\theta$}
	\label{fig:different theta}
\end{figure}
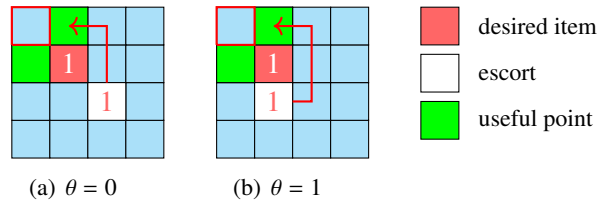

As shown in Figure \ref{fig:different theta}, one desired item can be searched both horizontally and vertically to obtain two useful points. For a particular escort and $d$ desired items, a matrix of shape $d\times 2$ (denoted as
$d_{escort\_manhattan\_matrix}$) can be used to store $d_{escort\_manhattan}$. While the desired item may occasionally lack a vertical or horizontal useful point, we set $d_{escort\_manhattan}=\infty$ accordingly. 

The guided action selection mechanism is described in detail below.

\begin{algorithm}[H]
	\caption{Guided action selection mechanism}
	\begin{algorithmic}[1]
		\Require
		State
		\Ensure
		The selected action
		\State Randomly choose an escort
		\State ({\color{blue}$\bm{d_{escort\_manhattan\_matrix}}$})
		\For {item in desired items}
			\State Get the vertical useful point
			\State Compute $d_{escort\_manhattan}$ between the escort and the vertical useful point
			\State Get the horizontal useful point
			\State Compute $d_{escort\_manhattan}$ between the escort and the horizontal useful point
			\State Compute $d_{item\_manhattan}$
		\EndFor
		\State Find the $min\ d_{escort\_manhattan}$, corresponding $d_{item\_manhattan}$, desired item and useful point
		\State (\textbf{\color{blue} Find a useful move})
		\For {action in shuffled(up,down,left,right)}
			\State Take action
			\State Compute $d_{item\_manhattan\_new}$
			\If {$d_{item\_manhattan\_new} < d_{item\_manhattan}$}
				\State Return action
			\EndIf
			\If {$d_{item\_manhattan\_new} = d_{item\_manhattan}$}
				\State Compute $d_{escort\_manhattan\_new}$ between the escort and the useful point
				\If {$d_{escort\_manhattan\_new} < d_{escort\_manhattan}$}
				\State Return action
				\EndIf
			\EndIf
		\EndFor
	\end{algorithmic}
\end{algorithm}

\subsection{Dueling deep neural network}\label{Dueling deep neural network}
{\color{black}The neural network, which acts as an approximation function $Q(s,a)$, determines 
the value of a state-action pair $(s,a)$.} It appears natural to represent the PBS system as a picture, with each pixel representing a grid cell, and feed the image data into a convolutional neural network for learning. Experiments, however, demonstrate that this technique is unfavorable. The convolutional neural network tends to obliterate details in favor of capturing the primary features. While in this case, the action option is influenced by the ``image details''. Once the desired items or escorts are relocated, the action will alter completely. 

Considering the characteristics of the problem, {\color{black} we use an input layer, several hidden layers and an output layer, which are all fully connected layers, to build the deep neural network (DNN) and choose Relu as the activation function.} This structure can efficiently compute the Q-values for all actions in a given state S (\cite{Mnih2015}). {\color{black}The input of the neural network is represented by an array of the desired items' coordinates followed by the escorts’ coordinates, in the form of $(x_1, y_1, x_2, y_2, \cdots, x_d, y_d, x^\prime_1, y^\prime_1, x^\prime_2, y^\prime_2, \cdots, x^\prime_e, y^\prime_e)$ where $(x, y)$ are items' coordinates and $(x', y')$ are escorts' coordinates.}
Naturally, the numbers of hidden layers and neurons are up to the complexity of the specific problem. The more complex the problem, the higher the numbers; conversely, the lower the numbers. Moreover, we observe that states with a small $d_{item\_manhattan}$ generally have a higher {\color{black}state value} $V(s)$ than those with a larger $d_{item\_manhattan}$ in the PBS system. Scilicet, the state has an anticipated $V(s)$ independent of the actions available. As a result, we adopt the dueling DNN as our neural network architecture, {\color{black} as shown in Figure \ref{fig:DNN_Structure}.
According to \cite{wang2016dueling}, the dueling DNN can significantly improve the learning performance by separating the state-action value function $Q(s,a)$ into two parts: state value function $V(s)$ and advantage value function $A(s,a)$. Then, two deep neural networks, as shown in the top and bottom of Figure \ref{fig:DNN_Structure}, are used to estimate $V(s)$ and $A(s,a)$, respectively. After that, the Q value is calculated by $Q(s, a)=V(s)+(A(s, a)-\frac{1}{|\mathcal{A}|} \sum\limits_{a^{\prime} \in \mathcal{A}} A(s, a^{\prime}))$, where $\mathcal{A}$ is the set of all actions. For more details of Dueling DNN, see \cite{wang2016dueling}.}

\begin{figure}[H]
\centering
	\includegraphics[width=0.8\textwidth]{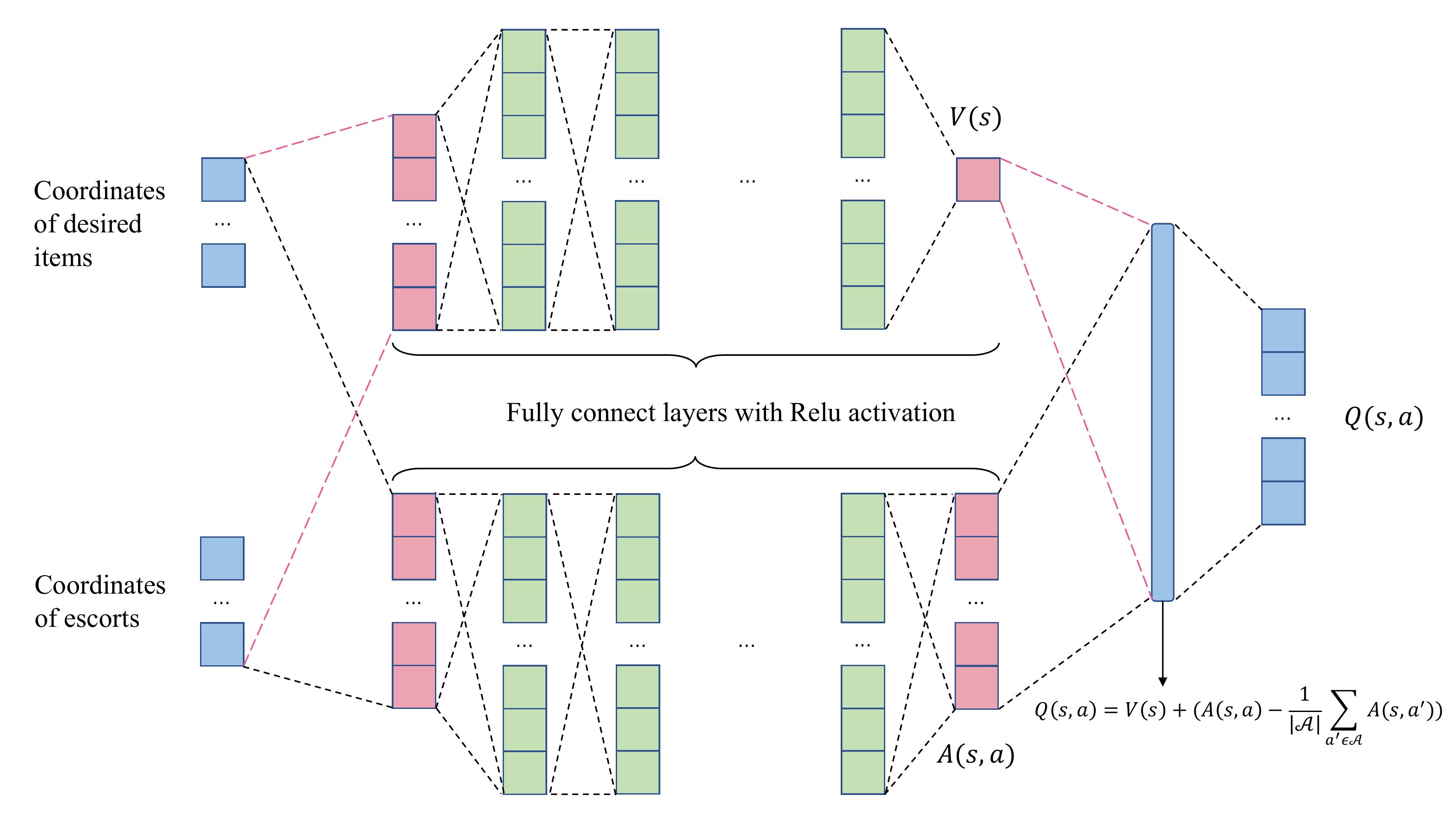}
\caption{Structure of dueling DNN}
\label{fig:DNN_Structure}
\end{figure}{}

\subsection{Double \& Dueling DQN algorithm for the PBS system}

We utilize DQN (\cite{Mnih2015}) as the framework for reinforcement learning{\color{black}, that is, to deduce the parameters $\theta$ (the weights of the connections) of the state-action value function $Q(s,a;\theta)$.} Furthermore, since overestimations of the value functions commonly occur in DQN, we adopt the Double DQN (\cite{2015Deep}) to solve our problem. {\color{black}Due to the existence of \textit{sumTree}, the data structure for the implementation of prioritized experience replay,} the priority DQN (\cite{schaul2015prioritized}) {\color{black}(an alternative improvement to DQN)} is somewhat time-consuming and hence not adopted.


When choosing an action, the agent has a probability of $\varepsilon$ to explore using a semi-random and semi-guided mechanism, {\color{black}and a  probability of $1-\varepsilon$ to exploit knowledge already revealed by the neural network}. The Dueling DNN introduced in Section \ref{Dueling deep neural network} serves as the action-value function $Q$ {\color{black}(with weights $\theta$)} and the target action-value function $\hat{Q}$ {\color{black}(with weights $\theta ^-$)}. {\color{black}For a mini-batch of samples, at iteration $j$, the expected Q value $y_j$ can be calculated by
\begin{align}
y_j=\left\{ \begin{matrix}
r_j& \text{if}\,\,done_{j+1}\\r_j+\gamma \hat{Q}\left( s_{j+1},arg\max_aQ\left( s_{j+1}, a; \theta \right),\theta ^- \right)& \text{otherwise}
\\
\end{matrix} \right.
\end{align}
where $done_{j+1}$ indicates whether all desired items reached their I/O point at the next iteration.
The loss function $L_j\left( \theta \right)$ measures the difference between the expected Q value and the predicted Q value and is written as
\begin{align}
L_j\left( \theta \right) =\mathbb{E}_{s,a}\left[ \left( y_j-Q\left( s_j,a_j;\theta\right) \right) ^2 \right] 
\end{align}
where $ Q\left( s_j,a_j;\theta \right) $ is the predicted Q value by function $Q$. 
The network parameter $\theta$ of function $Q$ is modified at each training step and $\theta ^-$ of function $\hat{Q}$ is updated every $C$ episodes according to a gradient descent method.}

The Double \& Dueling DQN algorithm dedicated to the PBS system is presented in detail in Algorithm \ref{algorithm2}.
	
\begin{algorithm}[H]
	\caption{Double \& Dueling DQN algorithm for the PBS system}
	\label{algorithm2}
	\begin{algorithmic}[1]
		\Require
		Initialize replay memory $D$ to capacity $N$ 
		\par
		Initialize action-value function $Q$ with random weights $\theta$
		\par
		Initialize target action-value function $\hat{Q}$ with weights $\theta ^-=\theta$  
		\par
		Initialize action choose parameters $\varepsilon$ and {\color{black}$\eta$}
		\Ensure
		\texttt{\color{blue}Trained $DNN$}
		\For {episode = 1, M}
			\State $t=0$
			\State Initialize state $s_t$ with $done_t = False$
			\State Update $\varepsilon$ and {\color{black}$\eta$} ($\varepsilon$ and {\color{black}$\eta$} gradually decrease)
			\While{not $done_t$}
				\State With probability $\varepsilon$ select action $a_t$ by using semi-random and semi-guided mechanism({\color{black}$\eta$})
				\State With probability $1-\varepsilon$ select action $a_t=arg\max _aQ\left(s_t,a;\theta \right)$
				\State Execute action $a_t$ in emulator and observe reward $r_t$, next state $s_{t+1}$ and if episode terminates $done_{t+1}$
				\State Store transition($s_t,a_t,r_t,s_{t+1},done_{t+1}$) in $D$
				\State Set $t=t+1$  
				\State Sample random mini-batch of transition ($s_j,a_j,r_j,s_{j+1},done_{j+1}$) from $D$
				\State Set $y_j=\left\{ \begin{matrix}r_j& \text{if}\,\,done_{j+1}\\r_j+\gamma \hat{Q}\left( s_{j+1},arg\max_aQ\left( s_{j+1},a;\theta \right),\theta ^- \right)& \text{otherwise}\\\end{matrix} \right.$
				\State Perform a gradient decent step on $\left( y_j-Q\left( s_j,a_j;\theta \right) \right) ^2$ with respect to the network parameters $\theta$
			\EndWhile
			\State Reset $\hat{Q} = Q$ every $C$ episodes 
		\EndFor
			
	\end{algorithmic}
\end{algorithm}

\section{Simultaneous movement consideration: a conversion algorithm}
\label{sec:Conversion}

The assumption of single-load movement wastes time and is therefore uneconomical for real warehouses. Multiple independent movements (for example, moves 1 and 2 in Figure \ref{fig:optimal path}) can be combined into a collection and then executed at the same timestamp. Several existing literature also makes the simultaneous movement assumption, e.g., \cite{Gue2013Gridstore}, \cite{Yu2016optimal}, and \cite{Bukchin2020optimal}. As seen in Figure \ref{fig:simultaneous}, the example in Figure \ref{fig:optimal path} can be completed in 8 timestamps instead of 13, reducing the processing time by about 40\%. Thus, allowing simultaneous movement is a reasonable and necessary requirement.

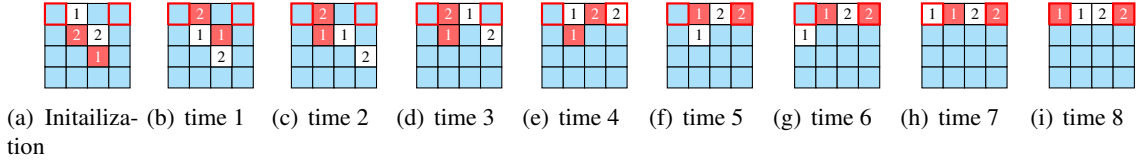
\begin{figure}[H]
\centering 
\hspace{-7mm}
	\subfigure[Initailization]{
	\tikzstyle{Point}=[circle,draw=black,fill=black,thick,
						inner sep=0pt,minimum size=2mm]
	\tikzstyle{arrow} = [thick,->,>=stealth]
	\tikzstyle{arrowonly} = [thick,-,>=stealth,line width = 2pt, color = cyan]
	\tikzstyle{arrowFrom} = [thick,<-,>=stealth]
	\begin{tikzpicture}[auto, scale=0.28]
	 \foreach \x in {1,2,3,4}
	  \foreach \y in {1,...,4}
	  {
	   \filldraw[fill=cyan!30] (\x,\y) +(-0.5,-0.5) rectangle ++(0.5,0.5);
	  }
	 \filldraw[fill=red!60] (3,2) +(-0.5,-0.5) rectangle ++(0.5,0.5);
	 \draw [white] (3,2) node{\tiny{1}}; 
	 \filldraw[fill=red!60] (2,3) +(-0.5,-0.5) rectangle ++(0.5,0.5);
	 \draw [white] (2,3) node{\tiny{2}}; 
	 \filldraw[fill=white] (2,4) +(-0.5,-0.5) rectangle ++(0.5,0.5);
	 \draw [black] (2,4) node{\tiny{1}}; 
	 \filldraw[fill=white] (3,3) +(-0.5,-0.5) rectangle ++(0.5,0.5);
	 \draw [black] (3,3) node{\tiny{2}}; 

	 \draw[-, thick, red] (0.5, 3.5) -- (1.5, 3.5) -- (1.5, 4.5) -- (0.5, 4.5) -- (0.5, 3.5);
	 \draw[-, thick, red] (3.5, 3.5) -- (3.5, 4.5) -- (4.5, 4.5) -- (4.5, 3.5) -- (3.5, 3.5);
	\end{tikzpicture}
	}\hspace{-2mm}
	\subfigure[time 1]{
	\tikzstyle{Point}=[circle,draw=black,fill=black,thick,
						inner sep=0pt,minimum size=2mm]
	\tikzstyle{arrow} = [thick,->,>=stealth]
	\tikzstyle{arrowonly} = [thick,-,>=stealth,line width = 2pt, color = cyan]
	\tikzstyle{arrowFrom} = [thick,<-,>=stealth]
	\begin{tikzpicture}[auto, scale=0.28]
	 \foreach \x in {1,2,3,4}
	  \foreach \y in {1,...,4}
	  {
	   \filldraw[fill=cyan!30] (\x,\y) +(-0.5,-0.5) rectangle ++(0.5,0.5);
	  }
	 \filldraw[fill=red!60] (3,3) +(-0.5,-0.5) rectangle ++(0.5,0.5);
	 \draw [white] (3,3) node{\tiny{1}}; 
	 \filldraw[fill=red!60] (2,4) +(-0.5,-0.5) rectangle ++(0.5,0.5);
	 \draw [white] (2,4) node{\tiny{2}}; 
	 \filldraw[fill=white] (2,3) +(-0.5,-0.5) rectangle ++(0.5,0.5);
	 \draw [black] (2,3) node{\tiny{1}}; 
	 \filldraw[fill=white] (3,2) +(-0.5,-0.5) rectangle ++(0.5,0.5);
	 \draw [black] (3,2) node{\tiny{2}}; 

	 \draw[-, thick, red] (0.5, 3.5) -- (1.5, 3.5) -- (1.5, 4.5) -- (0.5, 4.5) -- (0.5, 3.5);
	 \draw[-, thick, red] (3.5, 3.5) -- (3.5, 4.5) -- (4.5, 4.5) -- (4.5, 3.5) -- (3.5, 3.5);
	\end{tikzpicture}
	}\hspace{-2mm}
	\subfigure[time 2]{
	\tikzstyle{Point}=[circle,draw=black,fill=black,thick,
						inner sep=0pt,minimum size=2mm]
	\tikzstyle{arrow} = [thick,->,>=stealth]
	\tikzstyle{arrowonly} = [thick,-,>=stealth,line width = 2pt, color = cyan]
	\tikzstyle{arrowFrom} = [thick,<-,>=stealth]
	\begin{tikzpicture}[auto, scale=0.28]
	 \foreach \x in {1,2,3,4}
	  \foreach \y in {1,...,4}
	  {
	   \filldraw[fill=cyan!30] (\x,\y) +(-0.5,-0.5) rectangle ++(0.5,0.5);
	  }
	 \filldraw[fill=red!60] (2,3) +(-0.5,-0.5) rectangle ++(0.5,0.5);
	 \draw [white] (2,3) node{\tiny{1}}; 
	 \filldraw[fill=red!60] (2,4) +(-0.5,-0.5) rectangle ++(0.5,0.5);
	 \draw [white] (2,4) node{\tiny{2}}; 
	 \filldraw[fill=white] (3,3) +(-0.5,-0.5) rectangle ++(0.5,0.5);
	 \draw [black] (3,3) node{\tiny{1}}; 
	 \filldraw[fill=white] (4,2) +(-0.5,-0.5) rectangle ++(0.5,0.5);
	 \draw [black] (4,2) node{\tiny{2}}; 

	 \draw[-, thick, red] (0.5, 3.5) -- (1.5, 3.5) -- (1.5, 4.5) -- (0.5, 4.5) -- (0.5, 3.5);
	 \draw[-, thick, red] (3.5, 3.5) -- (3.5, 4.5) -- (4.5, 4.5) -- (4.5, 3.5) -- (3.5, 3.5);
	\end{tikzpicture}
	}\hspace{-2mm}
	\subfigure[time 3]{
	\tikzstyle{Point}=[circle,draw=black,fill=black,thick,
						inner sep=0pt,minimum size=2mm]
	\tikzstyle{arrow} = [thick,->,>=stealth]
	\tikzstyle{arrowonly} = [thick,-,>=stealth,line width = 2pt, color = cyan]
	\tikzstyle{arrowFrom} = [thick,<-,>=stealth]
	\begin{tikzpicture}[auto, scale=0.28]
	 \foreach \x in {1,2,3,4}
	  \foreach \y in {1,...,4}
	  {
	   \filldraw[fill=cyan!30] (\x,\y) +(-0.5,-0.5) rectangle ++(0.5,0.5);
	  }
	 \filldraw[fill=red!60] (2,3) +(-0.5,-0.5) rectangle ++(0.5,0.5);
	 \draw [white] (2,3) node{\tiny{1}}; 
	 \filldraw[fill=red!60] (2,4) +(-0.5,-0.5) rectangle ++(0.5,0.5);
	 \draw [white] (2,4) node{\tiny{2}}; 
	 \filldraw[fill=white] (3,4) +(-0.5,-0.5) rectangle ++(0.5,0.5);
	 \draw [black] (3,4) node{\tiny{1}}; 
	 \filldraw[fill=white] (4,3) +(-0.5,-0.5) rectangle ++(0.5,0.5);
	 \draw [black] (4,3) node{\tiny{2}};  

	 \draw[-, thick, red] (0.5, 3.5) -- (1.5, 3.5) -- (1.5, 4.5) -- (0.5, 4.5) -- (0.5, 3.5);
	 \draw[-, thick, red] (3.5, 3.5) -- (3.5, 4.5) -- (4.5, 4.5) -- (4.5, 3.5) -- (3.5, 3.5);
	\end{tikzpicture}
	}\hspace{-2mm}
	\subfigure[time 4]{
	\tikzstyle{Point}=[circle,draw=black,fill=black,thick,
						inner sep=0pt,minimum size=2mm]
	\tikzstyle{arrow} = [thick,->,>=stealth]
	\tikzstyle{arrowonly} = [thick,-,>=stealth,line width = 2pt, color = cyan]
	\tikzstyle{arrowFrom} = [thick,<-,>=stealth]
	\begin{tikzpicture}[auto, scale=0.28]
	 \foreach \x in {1,2,3,4}
	  \foreach \y in {1,...,4}
	  {
	   \filldraw[fill=cyan!30] (\x,\y) +(-0.5,-0.5) rectangle ++(0.5,0.5);
	  }
	 \filldraw[fill=red!60] (2,3) +(-0.5,-0.5) rectangle ++(0.5,0.5);
	 \draw [white] (2,3) node{\tiny{1}}; 
	 \filldraw[fill=red!60] (3,4) +(-0.5,-0.5) rectangle ++(0.5,0.5);
	 \draw [white] (3,4) node{\tiny{2}}; 
	 \filldraw[fill=white] (2,4) +(-0.5,-0.5) rectangle ++(0.5,0.5);
	 \draw [black] (2,4) node{\tiny{1}}; 
	 \filldraw[fill=white] (4,4) +(-0.5,-0.5) rectangle ++(0.5,0.5);
	 \draw [black] (4,4) node{\tiny{2}};

	 \draw[-, thick, red] (0.5, 3.5) -- (1.5, 3.5) -- (1.5, 4.5) -- (0.5, 4.5) -- (0.5, 3.5);
	 \draw[-, thick, red] (3.5, 3.5) -- (3.5, 4.5) -- (4.5, 4.5) -- (4.5, 3.5) -- (3.5, 3.5);
	\end{tikzpicture}
	}\hspace{-2mm}
	\subfigure[time 5]{
	\tikzstyle{Point}=[circle,draw=black,fill=black,thick,
						inner sep=0pt,minimum size=2mm]
	\tikzstyle{arrow} = [thick,->,>=stealth]
	\tikzstyle{arrowonly} = [thick,-,>=stealth,line width = 2pt, color = cyan]
	\tikzstyle{arrowFrom} = [thick,<-,>=stealth]
	\begin{tikzpicture}[auto, scale=0.28]
	 \foreach \x in {1,2,3,4}
	  \foreach \y in {1,...,4}
	  {
	   \filldraw[fill=cyan!30] (\x,\y) +(-0.5,-0.5) rectangle ++(0.5,0.5);
	  }
	 \filldraw[fill=red!60] (2,4) +(-0.5,-0.5) rectangle ++(0.5,0.5);
	 \draw [white] (2,4) node{\tiny{1}}; 
	 \filldraw[fill=red!60] (4,4) +(-0.5,-0.5) rectangle ++(0.5,0.5);
	 \draw [white] (4,4) node{\tiny{2}}; 
	 \filldraw[fill=white] (2,3) +(-0.5,-0.5) rectangle ++(0.5,0.5);
	 \draw [black] (2,3) node{\tiny{1}}; 
	 \filldraw[fill=white] (3,4) +(-0.5,-0.5) rectangle ++(0.5,0.5);
	 \draw [black] (3,4) node{\tiny{2}};  

	 \draw[-, thick, red] (0.5, 3.5) -- (1.5, 3.5) -- (1.5, 4.5) -- (0.5, 4.5) -- (0.5, 3.5);
	 \draw[-, thick, red] (3.5, 3.5) -- (3.5, 4.5) -- (4.5, 4.5) -- (4.5, 3.5) -- (3.5, 3.5);
	\end{tikzpicture}
	}\hspace{-2mm}
	\subfigure[time 6]{
	\tikzstyle{Point}=[circle,draw=black,fill=black,thick,
						inner sep=0pt,minimum size=2mm]
	\tikzstyle{arrow} = [thick,->,>=stealth]
	\tikzstyle{arrowonly} = [thick,-,>=stealth,line width = 2pt, color = cyan]
	\tikzstyle{arrowFrom} = [thick,<-,>=stealth]
	\begin{tikzpicture}[auto, scale=0.28]
	 \foreach \x in {1,2,3,4}
	  \foreach \y in {1,...,4}
	  {
	   \filldraw[fill=cyan!30] (\x,\y) +(-0.5,-0.5) rectangle ++(0.5,0.5);
	  }
	 \filldraw[fill=red!60] (2,4) +(-0.5,-0.5) rectangle ++(0.5,0.5);
	 \draw [white] (2,4) node{\tiny{1}}; 
	 \filldraw[fill=red!60] (4,4) +(-0.5,-0.5) rectangle ++(0.5,0.5);
	 \draw [white] (4,4) node{\tiny{2}}; 
	 \filldraw[fill=white] (1,3) +(-0.5,-0.5) rectangle ++(0.5,0.5);
	 \draw [black] (1,3) node{\tiny{1}}; 
	 \filldraw[fill=white] (3,4) +(-0.5,-0.5) rectangle ++(0.5,0.5);
	 \draw [black] (3,4) node{\tiny{2}};   

	 \draw[-, thick, red] (0.5, 3.5) -- (1.5, 3.5) -- (1.5, 4.5) -- (0.5, 4.5) -- (0.5, 3.5);
	 \draw[-, thick, red] (3.5, 3.5) -- (3.5, 4.5) -- (4.5, 4.5) -- (4.5, 3.5) -- (3.5, 3.5);
	\end{tikzpicture}
	}\hspace{-2mm}
	\subfigure[time 7]{
	\tikzstyle{Point}=[circle,draw=black,fill=black,thick,
						inner sep=0pt,minimum size=2mm]
	\tikzstyle{arrow} = [thick,->,>=stealth]
	\tikzstyle{arrowonly} = [thick,-,>=stealth,line width = 2pt, color = cyan]
	\tikzstyle{arrowFrom} = [thick,<-,>=stealth]
	\begin{tikzpicture}[auto, scale=0.28]
	 \foreach \x in {1,2,3,4}
	  \foreach \y in {1,...,4}
	  {
	   \filldraw[fill=cyan!30] (\x,\y) +(-0.5,-0.5) rectangle ++(0.5,0.5);
	  }
	 \filldraw[fill=red!60] (2,4) +(-0.5,-0.5) rectangle ++(0.5,0.5);
	 \draw [white] (2,4) node{\tiny{1}}; 
	 \filldraw[fill=red!60] (4,4) +(-0.5,-0.5) rectangle ++(0.5,0.5);
	 \draw [white] (4,4) node{\tiny{2}}; 
	 \filldraw[fill=white] (1,4) +(-0.5,-0.5) rectangle ++(0.5,0.5);
	 \draw [black] (1,4) node{\tiny{1}}; 
	 \filldraw[fill=white] (3,4) +(-0.5,-0.5) rectangle ++(0.5,0.5);
	 \draw [black] (3,4) node{\tiny{2}};  

	 \draw[-, thick, red] (0.5, 3.5) -- (1.5, 3.5) -- (1.5, 4.5) -- (0.5, 4.5) -- (0.5, 3.5);
	 \draw[-, thick, red] (3.5, 3.5) -- (3.5, 4.5) -- (4.5, 4.5) -- (4.5, 3.5) -- (3.5, 3.5);
	\end{tikzpicture}
	}\hspace{-2mm}
	\subfigure[time 8]{
	\tikzstyle{Point}=[circle,draw=black,fill=black,thick,
						inner sep=0pt,minimum size=2mm]
	\tikzstyle{arrow} = [thick,->,>=stealth]
	\tikzstyle{arrowonly} = [thick,-,>=stealth,line width = 2pt, color = cyan]
	\tikzstyle{arrowFrom} = [thick,<-,>=stealth]
	\begin{tikzpicture}[auto, scale=0.28]
	 \foreach \x in {1,2,3,4}
	  \foreach \y in {1,...,4}
	  {
	   \filldraw[fill=cyan!30] (\x,\y) +(-0.5,-0.5) rectangle ++(0.5,0.5);
	  }
	 \filldraw[fill=red!60] (1,4) +(-0.5,-0.5) rectangle ++(0.5,0.5);
	 \draw [white] (1,4) node{\tiny{1}}; 
	 \filldraw[fill=red!60] (4,4) +(-0.5,-0.5) rectangle ++(0.5,0.5);
	 \draw [white] (4,4) node{\tiny{2}}; 
	 \filldraw[fill=white] (2,4) +(-0.5,-0.5) rectangle ++(0.5,0.5);
	 \draw [black] (2,4) node{\tiny{1}}; 
	 \filldraw[fill=white] (3,4) +(-0.5,-0.5) rectangle ++(0.5,0.5);
	 \draw [black] (3,4) node{\tiny{2}};  

	 \draw[-, thick, red] (0.5, 3.5) -- (1.5, 3.5) -- (1.5, 4.5) -- (0.5, 4.5) -- (0.5, 3.5);
	 \draw[-, thick, red] (3.5, 3.5) -- (3.5, 4.5) -- (4.5, 4.5) -- (4.5, 3.5) -- (3.5, 3.5);
	\end{tikzpicture}
	}

	\caption{Optimal path with simultaneous movement}
	\label{fig:simultaneous}
\end{figure}

Since the reinforcement learning method is based on the single-load movement assumption, we develop an additional algorithm to convert the results into a simultaneous movement version. The main function of this algorithm is to detect whether the current escort's action contradicts the prior ones. When no conflict appears, escorts can move at the same timestamp. If not, the current escort must wait. 
The first type of conflict occurs when the current escort is moving to the locations the items just arrived. In other words, an item can only be moved once during the same time period. Take Figure \ref{fig:conflicts}\subref{fig:conflict1} as an example, both escorts 1 and 2 serve the desired item 1. As a result, these two processes should be done independently. Sometimes, it is possible for two escorts to exchange positions and the passively moving escort may cause the second form of conflict. As shown in Figure \ref{fig:conflicts}\subref{fig:conflict2}, the single-load movement solution is (1-right,2-up), and the movement of escort 1 (1-right) leads to the passive movement of escort 2 (2-left). Therefore, the active movement of escort 2 (2-up) should be deferred until the passive action is completed.
Based on these two forms of conflict, we summarize two principles of identification. Considering the locations of the previous escorts as forbidden places, principle 1 stipulates that the escort cannot enter forbidden areas and principle 2 states that the escort currently in forbidden areas cannot leave.

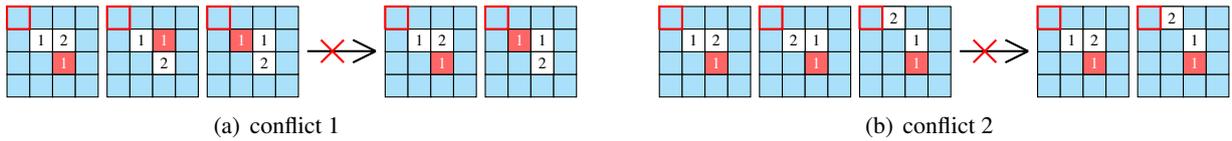
\begin{figure}[H]
\centering 
\hspace{-7mm}
	\subfigure[conflict 1]{
	\label{fig:conflict1}
	\tikzstyle{Point}=[circle,draw=black,fill=black,thick,
						inner sep=0pt,minimum size=2mm]
	\tikzstyle{arrow} = [thick,->,>=stealth]
	\tikzstyle{arrowonly} = [thick,-,>=stealth,line width = 2pt, color = cyan]
	\tikzstyle{arrowFrom} = [thick,<-,>=stealth]
	\begin{tikzpicture}[auto, scale=0.3]
	 \foreach \x in {1,2,3,4}
	  \foreach \y in {1,...,4}
	  {
	   \filldraw[fill=cyan!30] (\x,\y) +(-0.5,-0.5) rectangle ++(0.5,0.5);
	  }
	 \filldraw[fill=red!60] (3,2) +(-0.5,-0.5) rectangle ++(0.5,0.5);
	 \draw [white] (3,2) node{\tiny{1}}; 
	 \filldraw[fill=white] (2,3) +(-0.5,-0.5) rectangle ++(0.5,0.5);
	 \draw [black] (2,3) node{\tiny{1}}; 
	 \filldraw[fill=white] (3,3) +(-0.5,-0.5) rectangle ++(0.5,0.5);
	 \draw [black] (3,3) node{\tiny{2}}; 

	 \draw[-, thick, red] (0.5, 3.5) -- (1.5, 3.5) -- (1.5, 4.5) -- (0.5, 4.5) -- (0.5, 3.5);
	 \end{tikzpicture}
	\begin{tikzpicture}[auto, scale=0.3]
	 \foreach \x in {1,2,3,4}
	  \foreach \y in {1,...,4}
	  {
	   \filldraw[fill=cyan!30] (\x,\y) +(-0.5,-0.5) rectangle ++(0.5,0.5);
	  }
	 \filldraw[fill=red!60] (3,3) +(-0.5,-0.5) rectangle ++(0.5,0.5);
	 \draw [white] (3,3) node{\tiny{1}}; 
	 \filldraw[fill=white] (2,3) +(-0.5,-0.5) rectangle ++(0.5,0.5);
	 \draw [black] (2,3) node{\tiny{1}}; 
	 \filldraw[fill=white] (3,2) +(-0.5,-0.5) rectangle ++(0.5,0.5);
	 \draw [black] (3,2) node{\tiny{2}}; 

	 \draw[-, thick, red] (0.5, 3.5) -- (1.5, 3.5) -- (1.5, 4.5) -- (0.5, 4.5) -- (0.5, 3.5);
	\end{tikzpicture}
	\begin{tikzpicture}[auto, scale=0.3]
	 \foreach \x in {1,2,3,4}
	  \foreach \y in {1,...,4}
	  {
	   \filldraw[fill=cyan!30] (\x,\y) +(-0.5,-0.5) rectangle ++(0.5,0.5);
	  }
	 \filldraw[fill=red!60] (2,3) +(-0.5,-0.5) rectangle ++(0.5,0.5);
	 \draw [white] (2,3) node{\tiny{1}}; 
	 \filldraw[fill=white] (3,3) +(-0.5,-0.5) rectangle ++(0.5,0.5);
	 \draw [black] (3,3) node{\tiny{1}}; 
	 \filldraw[fill=white] (3,2) +(-0.5,-0.5) rectangle ++(0.5,0.5);
	 \draw [black] (3,2) node{\tiny{2}}; 

	 \draw[-, thick, red] (0.5, 3.5) -- (1.5, 3.5) -- (1.5, 4.5) -- (0.5, 4.5) -- (0.5, 3.5);
	\end{tikzpicture}
	\begin{tikzpicture}[auto, scale=0.3]
	 \draw[-, thick, white] (0.5, 0) -- (0.5, 1);
	 \draw[-, thick, black] (0.5, 2) -- (3.5, 2);
	 \draw[-, thick, black] (2.5, 2.5) -- (3.5, 2);
	 \draw[-, thick, black] (2.5, 1.5) -- (3.5, 2);
	 \draw[-, thick, red] (1.1, 1.5) -- (2.1, 2.5);
	 \draw[-, thick, red] (1.1, 2.5) -- (2.1, 1.5);
	\end{tikzpicture}
	\begin{tikzpicture}[auto, scale=0.3]
	 \foreach \x in {1,2,3,4}
	  \foreach \y in {1,...,4}
	  {
	   \filldraw[fill=cyan!30] (\x,\y) +(-0.5,-0.5) rectangle ++(0.5,0.5);
	    }
	 \filldraw[fill=red!60] (3,2) +(-0.5,-0.5) rectangle ++(0.5,0.5);
	 \draw [white] (3,2) node{\tiny{1}}; 
	 \filldraw[fill=white] (2,3) +(-0.5,-0.5) rectangle ++(0.5,0.5);
	 \draw [black] (2,3) node{\tiny{1}}; 
	 \filldraw[fill=white] (3,3) +(-0.5,-0.5) rectangle ++(0.5,0.5);
	 \draw [black] (3,3) node{\tiny{2}}; 

	 \draw[-, thick, red] (0.5, 3.5) -- (1.5, 3.5) -- (1.5, 4.5) -- (0.5, 4.5) -- (0.5, 3.5);
	\end{tikzpicture}
	\begin{tikzpicture}[auto, scale=0.3]
	 \foreach \x in {1,2,3,4}
	  \foreach \y in {1,...,4}
	  {
	   \filldraw[fill=cyan!30] (\x,\y) +(-0.5,-0.5) rectangle ++(0.5,0.5);
	  }
	 \filldraw[fill=red!60] (2,3) +(-0.5,-0.5) rectangle ++(0.5,0.5);
	 \draw [white] (2,3) node{\tiny{1}}; 
	 \filldraw[fill=white] (3,3) +(-0.5,-0.5) rectangle ++(0.5,0.5);
	 \draw [black] (3,3) node{\tiny{1}}; 
	 \filldraw[fill=white] (3,2) +(-0.5,-0.5) rectangle ++(0.5,0.5);
	 \draw [black] (3,2) node{\tiny{2}}; 

	 \draw[-, thick, red] (0.5, 3.5) -- (1.5, 3.5) -- (1.5, 4.5) -- (0.5, 4.5) -- (0.5, 3.5);
	\end{tikzpicture}
	}
    \hspace{0.3cm}
	\subfigure[conflict 2]{
	\label{fig:conflict2}
	\tikzstyle{Point}=[circle,draw=black,fill=black,thick,
						inner sep=0pt,minimum size=2mm]
	\tikzstyle{arrow} = [thick,->,>=stealth]
	\tikzstyle{arrowonly} = [thick,-,>=stealth,line width = 2pt, color = cyan]
	\tikzstyle{arrowFrom} = [thick,<-,>=stealth]
	\begin{tikzpicture}[auto, scale=0.3]
	 \foreach \x in {1,2,3,4}
	  \foreach \y in {1,...,4}
	  {
	   \filldraw[fill=cyan!30] (\x,\y) +(-0.5,-0.5) rectangle ++(0.5,0.5);
	  }
	 \filldraw[fill=red!60] (3,2) +(-0.5,-0.5) rectangle ++(0.5,0.5);
	 \draw [white] (3,2) node{\tiny{1}}; 
	 \filldraw[fill=white] (2,3) +(-0.5,-0.5) rectangle ++(0.5,0.5);
	 \draw [black] (2,3) node{\tiny{1}}; 
	 \filldraw[fill=white] (3,3) +(-0.5,-0.5) rectangle ++(0.5,0.5);
	 \draw [black] (3,3) node{\tiny{2}}; 

	 \draw[-, thick, red] (0.5, 3.5) -- (1.5, 3.5) -- (1.5, 4.5) -- (0.5, 4.5) -- (0.5, 3.5);
	 \end{tikzpicture}
	\begin{tikzpicture}[auto, scale=0.3]
	 \foreach \x in {1,2,3,4}
	  \foreach \y in {1,...,4}
	  {
	   \filldraw[fill=cyan!30] (\x,\y) +(-0.5,-0.5) rectangle ++(0.5,0.5);
	  }
	 \filldraw[fill=red!60] (3,2) +(-0.5,-0.5) rectangle ++(0.5,0.5);
	 \draw [white] (3,2) node{\tiny{1}}; 
	 \filldraw[fill=white] (3,3) +(-0.5,-0.5) rectangle ++(0.5,0.5);
	 \draw [black] (3,3) node{\tiny{1}}; 
	 \filldraw[fill=white] (2,3) +(-0.5,-0.5) rectangle ++(0.5,0.5);
	 \draw [black] (2,3) node{\tiny{2}}; 

	 \draw[-, thick, red] (0.5, 3.5) -- (1.5, 3.5) -- (1.5, 4.5) -- (0.5, 4.5) -- (0.5, 3.5);
	\end{tikzpicture}
	\begin{tikzpicture}[auto, scale=0.3]
	 \foreach \x in {1,2,3,4}
	  \foreach \y in {1,...,4}
	  {
	   \filldraw[fill=cyan!30] (\x,\y) +(-0.5,-0.5) rectangle ++(0.5,0.5);
	  }
	 \filldraw[fill=red!60] (3,2) +(-0.5,-0.5) rectangle ++(0.5,0.5);
	 \draw [white] (3,2) node{\tiny{1}}; 
	 \filldraw[fill=white] (3,3) +(-0.5,-0.5) rectangle ++(0.5,0.5);
	 \draw [black] (3,3) node{\tiny{1}}; 
	 \filldraw[fill=white] (2,4) +(-0.5,-0.5) rectangle ++(0.5,0.5);
	 \draw [black] (2,4) node{\tiny{2}}; 

	 \draw[-, thick, red] (0.5, 3.5) -- (1.5, 3.5) -- (1.5, 4.5) -- (0.5, 4.5) -- (0.5, 3.5);
	\end{tikzpicture}
	\begin{tikzpicture}[auto, scale=0.3]
	 \draw[-, thick, white] (0.5, 0) -- (0.5, 1);
	 \draw[-, thick, black] (0.5, 2) -- (3.5, 2);
	 \draw[-, thick, black] (2.5, 2.5) -- (3.5, 2);
	 \draw[-, thick, black] (2.5, 1.5) -- (3.5, 2);
	 \draw[-, thick, red] (1.1, 1.5) -- (2.1, 2.5);
	 \draw[-, thick, red] (1.1, 2.5) -- (2.1, 1.5);
	\end{tikzpicture}
	\begin{tikzpicture}[auto, scale=0.3]
	 \foreach \x in {1,2,3,4}
	  \foreach \y in {1,...,4}
	  {
	   \filldraw[fill=cyan!30] (\x,\y) +(-0.5,-0.5) rectangle ++(0.5,0.5);
	    }
	 \filldraw[fill=red!60] (3,2) +(-0.5,-0.5) rectangle ++(0.5,0.5);
	 \draw [white] (3,2) node{\tiny{1}}; 
	 \filldraw[fill=white] (2,3) +(-0.5,-0.5) rectangle ++(0.5,0.5);
	 \draw [black] (2,3) node{\tiny{1}}; 
	 \filldraw[fill=white] (3,3) +(-0.5,-0.5) rectangle ++(0.5,0.5);
	 \draw [black] (3,3) node{\tiny{2}}; 

	 \draw[-, thick, red] (0.5, 3.5) -- (1.5, 3.5) -- (1.5, 4.5) -- (0.5, 4.5) -- (0.5, 3.5);
	\end{tikzpicture}
	\begin{tikzpicture}[auto, scale=0.3]
	 \foreach \x in {1,2,3,4}
	  \foreach \y in {1,...,4}
	  {
	   \filldraw[fill=cyan!30] (\x,\y) +(-0.5,-0.5) rectangle ++(0.5,0.5);
	  }
	 \filldraw[fill=red!60] (3,2) +(-0.5,-0.5) rectangle ++(0.5,0.5);
	 \draw [white] (3,2) node{\tiny{1}}; 
	 \filldraw[fill=white] (3,3) +(-0.5,-0.5) rectangle ++(0.5,0.5);
	 \draw [black] (3,3) node{\tiny{1}}; 
	 \filldraw[fill=white] (2,4) +(-0.5,-0.5) rectangle ++(0.5,0.5);
	 \draw [black] (2,4) node{\tiny{2}}; 

	 \draw[-, thick, red] (0.5, 3.5) -- (1.5, 3.5) -- (1.5, 4.5) -- (0.5, 4.5) -- (0.5, 3.5);
	\end{tikzpicture}
	}

	\caption{Two conflict situations of simultaneous movement}
	\label{fig:conflicts}
\end{figure}

Now, we are ready to present the conversion algorithm in Algorithm \ref{Conversion_algorithm}. Based on the single-load movement solution, check whether it violates principles 1 and 2 for each escort. If it does not, it can be combined with the previous escorts, otherwise, it must wait. Iterate this process until the conversion is complete. The list \texttt{move\_timestamps} stores all move timestamps for each escort, the list 
\texttt{move\_actions} stores the action at each timestamp and the list \texttt{escort\_locations} stores the locations of all escorts for each timestamp.

\section{Solving large-scale instances: a decomposition framework}
\label{sec:decompositionframework}

Reinforcement learning is capable of providing solutions for small-sized instances (e.g., grid size {\color{black}$8\times 8$}) in milliseconds. As the problem size increases, the training time required for reinforcement learning becomes longer. Consider a $20\times 20$ PBS system with two desired items and three escorts, the state space of which is $A_{20\times 20}^{2+3}-A_{20\times 20-2}^{3} = 9.98617\times 10^{12}$, 
implying an unaffordable training time. Thus, reinforcement learning is too time-consuming to be applied directly in real-world industrial scenarios. To extend it to large-scale cases, we construct a framework (as seen in Figure \ref{fig:DecompositionFramework}) to decompose large-scale instances into smaller ones, {\color{black}called sub-PBS,} and solve them one by one. 
With such an approach, the advantages of reinforcement learning are fully exploited. Note that this framework can also integrate existing exact and heuristic algorithms.

\begin{algorithm}[t]
	\caption{Conversion algorithm from single-load to simultaneous movement}
	\label{Conversion_algorithm}
	\begin{algorithmic}[1]
		\Require
		Lists \texttt{move\_timestamps}, \texttt{move\_actions} and \texttt{escort\_locations}
		\Ensure
		List \texttt{simultaneous\_moves}
		\While {True}
            \State Save the next move timestamp for each escort as \texttt{available} according to \texttt{move\_timestamps}
            \If {\texttt{available} is empty}
				\State break
			\EndIf
            \State Sort \texttt{available} according to the timestamps
			\State (\textbf{\color{blue} The first move in \texttt{available} is absolutely allowed})
            \State Add the first escort in \texttt{available}, its current location, and action into \texttt{moved}, \texttt{prohibited}, and \texttt{simultaneous\_moves\_temp} respectively
            \State Delete timestamp 1 in \texttt{move\_timestamps}
            \State Delete the first escort in \texttt{available}
            \State Let from denotes timestamp 1
			\For {move\_info in \texttt{available}}
				\State Get the escort and let to denotes the new timestamp
				\For {time in range(from+1, to)}
					\State Get the temp\_escort which is moved at timestamp time
					\State Add the current location of temp\_escort into \texttt{prohibited}
				\EndFor
				\State (\textbf{\color{blue} Conflicts 1 and 2})
				\If {the next or current location of the escort is in \texttt{prohibited}}
					\State Continue
				\Else
					\State Add the action of the escort into \texttt{simultaneous\_moves\_temp}
					\State Delete to in \texttt{move\_timestamps}
				\EndIf
				\State Add the escort and its current loaction into \texttt{moved} and \texttt{prohibited} respectively
				\State from = to
			\EndFor
			\State Add \texttt{simultaneous\_moves\_temp} into \texttt{simultaneous\_moves}
		\EndWhile
	\end{algorithmic}
\end{algorithm}

\begin{figure}[H]
\centering
\includegraphics[width=0.45\textwidth] {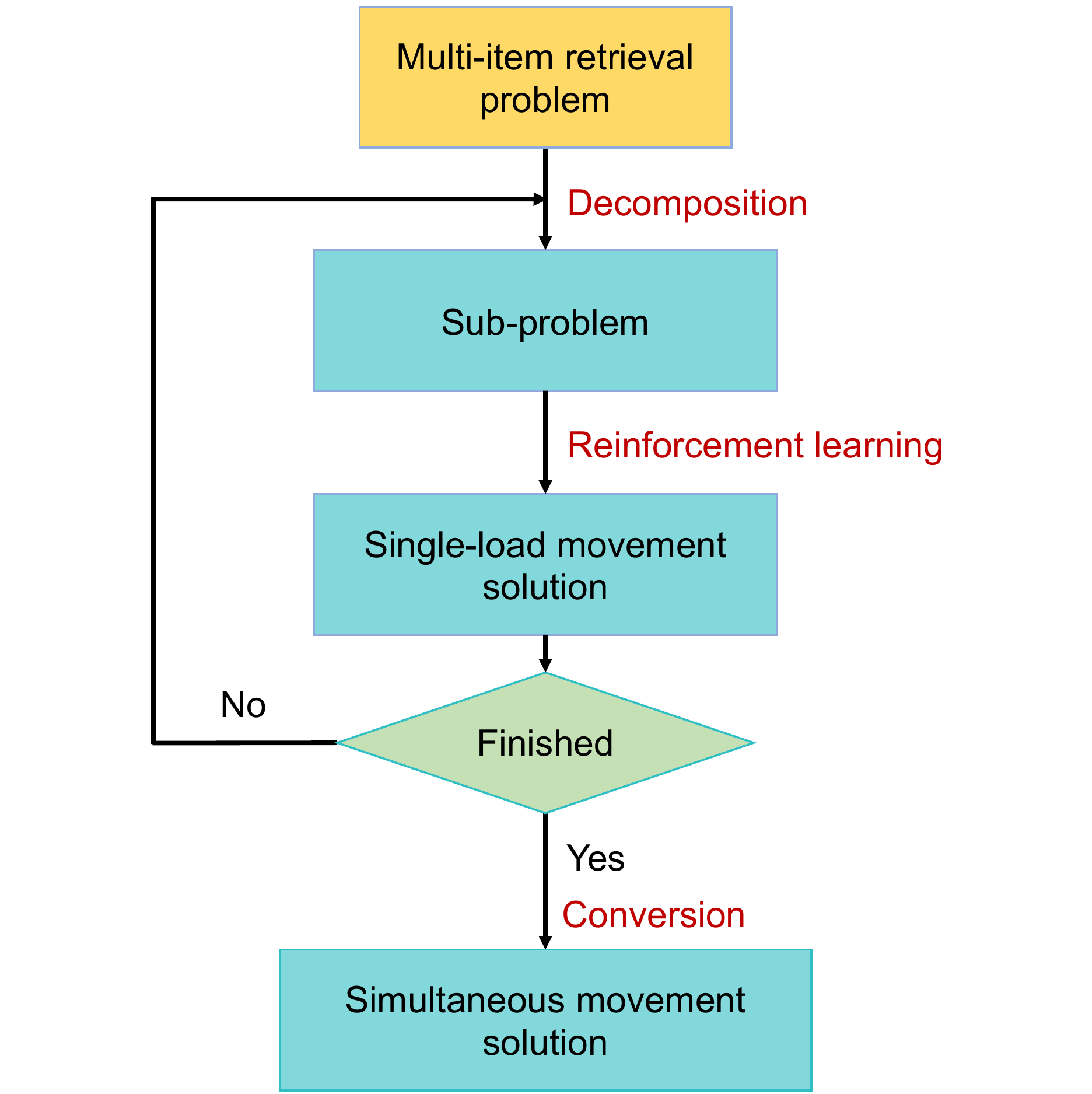}
\caption{Decomposition algorithm framework}
\label{fig:DecompositionFramework}
\end{figure}

{\color{black}The decomposition framework works as follows.}
\begin{itemize}
\item {\color{black}First, determine the grid size of the sub-PBS. Similar to the filter in a convolutional neural network (CNN), the sub-PBS region will slide through the PBS system, thus generating the small-sized problems to be solved. And there are several points to note: (i) Generally speaking, since the decomposition process usually has a negative impact on the solution, the sub-PBS should be as large as possible. At the extreme, if the sub-PBS is the same size as the PBS, i.e., if the original problem is solved directly, the error caused by the decomposition process will not occur. In our experiments, we used a dimension of $5\times5$. (ii) The sliding of the sub-PBS regions should be overlapping, i.e., the stride should be smaller than the corresponding edge length, thus ensuring that the desired items have a chance to reach any location in the PBS system. In addition, a small stride allows choosing from more sub-problems, but it may lead to increased computation burden at the same time. With a long stride, the computational effort will decrease, but it is likely to result in a limited choice of sub-problems. In short, there is no clear set of criteria for choosing the stride. In our experiments, we randomly use a stride of 1 or 2 to balance the speed and accuracy.

\item Then, slide the sub-PBS region from bottom to top and from left to right in the PBS system until there is at least one desired item in the sub-PBS region.
\item Next, generate a solvable sub-PBS problem according to the sub-PBS region.} We need to determine where to relocate the desired items within the sub-PBS region. {\color{black}Specifically, the (virtual) I/O point of an item is assigned to an ambient location of the region which is mostly toward the final I/O point.} Moreover, we assume that there should be at least $e$ ($e\ge1$) escorts in the sub-PBS region. 
If there are less than $e$ escorts, we must transfer the available escorts from outside to the current sub-PBS region; to distinguish them, we refer to these escorts as transferred escorts.

\item Following, execute the reinforcement learning algorithm (can be other exact and heuristic algorithms) to solve the sub-PBS problem. Note that some transferred escorts may not be utilized in the solution of the current sub-PBS problem, resulting in a waste of the total number of movements. These invalid escorts would return to their previous locations, and the resulted additional movements would not be recorded. 
\item Finally, convert the single-load movement solution into a simultaneous movement version with Algorithm \ref{Conversion_algorithm} after completing the iterative process.
\end{itemize}
}

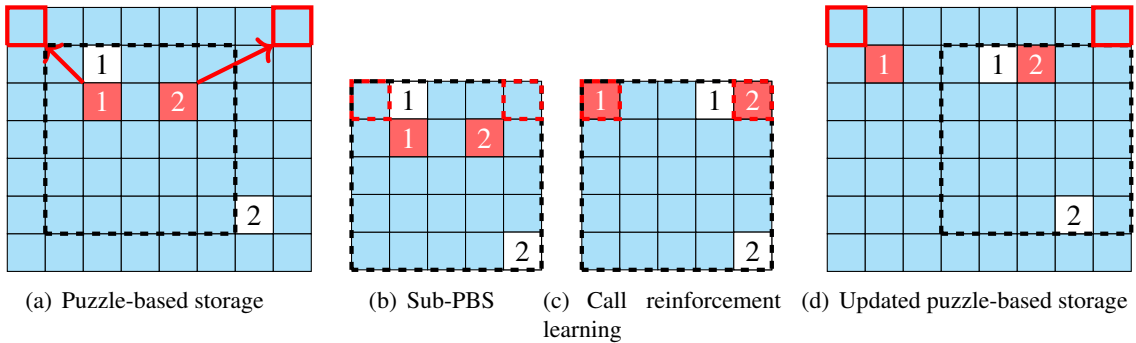
\begin{figure}[H]
\centering
\hspace{-7mm}
	\subfigure[Puzzle-based storage]{
	\label{fig:Puzzle-based storage}
	\tikzstyle{Point}=[circle,draw=black,fill=black,thick,
						inner sep=0pt,minimum size=2mm]
	\tikzstyle{arrow} = [thick,->,>=stealth]
	\tikzstyle{arrowonly} = [thick,-,>=stealth,line width = 2pt, color = cyan]
	\tikzstyle{arrowFrom} = [thick,<-,>=stealth]
	\begin{tikzpicture}[auto, scale=0.5]
	 \foreach \x in {1,...,8}
	  \foreach \y in {1,...,7}
	  {
	   \filldraw[fill=cyan!30] (\x,\y) +(-0.5,-0.5) rectangle ++(0.5,0.5);
	  }
	 \filldraw[fill=white] (3,6) +(-0.5,-0.5) rectangle ++(0.5,0.5);
	 \draw [black] (3,6) node{1}; 
	 \filldraw[fill=white] (7,2) +(-0.5,-0.5) rectangle ++(0.5,0.5);
	 \draw [black] (7,2) node{2}; 
	 \filldraw[fill=red!60] (5,5) +(-0.5,-0.5) rectangle ++(0.5,0.5);
	 \draw [white] (5,5) node{2}; 
	 \filldraw[fill=red!60] (3,5) +(-0.5,-0.5) rectangle ++(0.5,0.5);
	 \draw [white] (3,5) node{1}; 

	 \draw[->, ultra thick, red] (5.5,5.5) -- (7.5,6.5);
	 \draw[->, ultra thick, red] (2.5,5.5) -- (1.5,6.5);
	 \draw[dashed, ultra thick, black] (1.5, 1.5) -- (1.5, 6.5) -- (6.5, 6.5) -- (6.5, 1.5) -- (1.5, 1.5);

	 \draw[-, ultra thick, red] (8.5, 6.5) -- (8.5, 7.5) -- (7.5, 7.5) -- (7.5, 6.5) -- (8.5, 6.5);
	 \draw[-, ultra thick, red] (0.5, 6.5) -- (0.5, 7.5) -- (1.5, 7.5) -- (1.5, 6.5) -- (0.5, 6.5);
	\end{tikzpicture}
	}\hspace{-2mm}
	\subfigure[Sub-PBS]{
	\label{fig:sub-PBS}
	\tikzstyle{Point}=[circle,draw=black,fill=black,thick,
						inner sep=0pt,minimum size=2mm]
	\tikzstyle{arrow} = [thick,->,>=stealth]
	\tikzstyle{arrowonly} = [thick,-,>=stealth,line width = 2pt, color = cyan]
	\tikzstyle{arrowFrom} = [thick,<-,>=stealth]
	\begin{tikzpicture}[auto, scale=0.5]
	 \foreach \x in {1,...,5}
	  \foreach \y in {1,...,5}
	  {
	   \filldraw[fill=cyan!30] (\x,\y) +(-0.5,-0.5) rectangle ++(0.5,0.5);
	  }
	 \filldraw[fill=white] (2,5) +(-0.5,-0.5) rectangle ++(0.5,0.5);
	 \draw [black] (2,5) node{1}; 
	 \filldraw[fill=white] (5,1) +(-0.5,-0.5) rectangle ++(0.5,0.5);
	 \draw [black] (5,1) node{2}; 
	 \filldraw[fill=red!60] (4,4) +(-0.5,-0.5) rectangle ++(0.5,0.5);
	 \draw [white] (4,4) node{2}; 
	 \filldraw[fill=red!60] (2,4) +(-0.5,-0.5) rectangle ++(0.5,0.5);
	 \draw [white] (2,4) node{1}; 


	 \draw[dashed, ultra thick, black] (0.5, 0.5) -- (0.5, 5.5) -- (5.5, 5.5) -- (5.5, 0.5) -- (0.5, 0.5);

	 \draw[dashed, ultra thick, red] (0.5, 4.5) -- (0.5, 5.5) -- (1.5, 5.5) -- (1.5, 4.5) -- (0.5, 4.5);
	 \draw[dashed, ultra thick, red] (4.5, 4.5) -- (4.5, 5.5) -- (5.5, 5.5) -- (5.5, 4.5) -- (4.5, 4.5);
	\end{tikzpicture}
	}\hspace{-2mm}
	\subfigure[Call reinforcement learning]{
	\label{fig:reinforcement}
	\tikzstyle{Point}=[circle,draw=black,fill=black,thick,
						inner sep=0pt,minimum size=2mm]
	\tikzstyle{arrow} = [thick,->,>=stealth]
	\tikzstyle{arrowonly} = [thick,-,>=stealth,line width = 2pt, color = cyan]
	\tikzstyle{arrowFrom} = [thick,<-,>=stealth]
	\begin{tikzpicture}[auto, scale=0.5]
	 \foreach \x in {1,...,5}
	  \foreach \y in {1,...,5}
	  {
	   \filldraw[fill=cyan!30] (\x,\y) +(-0.5,-0.5) rectangle ++(0.5,0.5);
	  }
	 \filldraw[fill=white] (4,5) +(-0.5,-0.5) rectangle ++(0.5,0.5);
	 \draw [black] (4,5) node{1}; 
	 \filldraw[fill=white] (5,1) +(-0.5,-0.5) rectangle ++(0.5,0.5);
	 \draw [black] (5,1) node{2}; 
	 \filldraw[fill=red!60] (5,5) +(-0.5,-0.5) rectangle ++(0.5,0.5);
	 \draw [white] (5,5) node{2}; 
	 \filldraw[fill=red!60] (1,5) +(-0.5,-0.5) rectangle ++(0.5,0.5);
	 \draw [white] (1,5) node{1}; 


	 \draw[dashed, ultra thick, black] (0.5, 0.5) -- (0.5, 5.5) -- (5.5, 5.5) -- (5.5, 0.5) -- (0.5, 0.5);

	 \draw[dashed, ultra thick, red] (0.5, 4.5) -- (0.5, 5.5) -- (1.5, 5.5) -- (1.5, 4.5) -- (0.5, 4.5);
	 \draw[dashed, ultra thick, red] (4.5, 4.5) -- (4.5, 5.5) -- (5.5, 5.5) -- (5.5, 4.5) -- (4.5, 4.5);
	\end{tikzpicture}
	}
	\subfigure[Updated puzzle-based storage]{
	\label{fig:Updated}
	\tikzstyle{Point}=[circle,draw=black,fill=black,thick,
						inner sep=0pt,minimum size=2mm]
	\tikzstyle{arrow} = [thick,->,>=stealth]
	\tikzstyle{arrowonly} = [thick,-,>=stealth,line width = 2pt, color = cyan]
	\tikzstyle{arrowFrom} = [thick,<-,>=stealth]
	\begin{tikzpicture}[auto, scale=0.5]
	 \foreach \x in {1,...,8}
	  \foreach \y in {1,...,7}
	  {
	   \filldraw[fill=cyan!30] (\x,\y) +(-0.5,-0.5) rectangle ++(0.5,0.5);
	  }
	 \filldraw[fill=white] (5,6) +(-0.5,-0.5) rectangle ++(0.5,0.5);
	 \draw [black] (5,6) node{1}; 
	 \filldraw[fill=white] (7,2) +(-0.5,-0.5) rectangle ++(0.5,0.5);
	 \draw [black] (7,2) node{2}; 
	 \filldraw[fill=red!60] (2,6) +(-0.5,-0.5) rectangle ++(0.5,0.5);
	 \draw [white] (2,6) node{1}; 
	 \filldraw[fill=red!60] (6,6) +(-0.5,-0.5) rectangle ++(0.5,0.5);
	 \draw [white] (6,6) node{2}; 

	 \draw[dashed, ultra thick, black] (3.5, 1.5) -- (3.5, 6.5) -- (8.5, 6.5) -- (8.5, 1.5) -- (3.5, 1.5);

	 \draw[-, ultra thick, red] (8.5, 6.5) -- (8.5, 7.5) -- (7.5, 7.5) -- (7.5, 6.5) -- (8.5, 6.5);
	 \draw[-, ultra thick, red] (0.5, 6.5) -- (0.5, 7.5) -- (1.5, 7.5) -- (1.5, 6.5) -- (0.5, 6.5);
	\end{tikzpicture}
	}\hspace{-2mm}
	\caption{Iterative example of the decomposition algorithm}
	\label{fig:decomposition}
\end{figure}

We present an example to illustrate the process of a single iteration. As shown in Figure \ref{fig:decomposition}\subref{fig:Puzzle-based storage}, there is a sub-PBS region marked by black dotted lines. In Figure \ref{fig:decomposition}\subref{fig:sub-PBS}, we extract the sub-PBS and mark the virtual I/O points of the desired items 1 and 2 with red dotted lines. Let $e = 2$, unfortunately, there are not enough escorts. Therefore we transfer escorts 1 and 2 into the region. In Figure \ref{fig:decomposition}\subref{fig:reinforcement}, we apply the reinforcement learning algorithm to solve the sub-PBS issue, indicating that transferred escort 2 is not used. Finally, in Figure \ref{fig:decomposition}\subref{fig:Updated}, escort 2 returns to its previous position, while the sub-PBS region advances to the next iteration.

\section{Numerical Experiments}
\label{sec:Numerical}
We conduct extensive numerical experiments to investigate the performance of the reinforcement learning approach and the decomposition framework (in which the conversion algorithm is embedded). We first compare the solution of the reinforcement learning approach to the optimal solution obtained by the IP model. After which, we compare our approach with existing algorithms in their corresponding problem settings. Then, we explore the boundary of the proposed method via increasing the grid size and construct a regression model to estimate the retrieval effort. Finally, we demonstrate the effectiveness of the decomposition framework by examining medium- and large-scale instances. Our purpose is to demonstrate that (i) the reinforcement learning solutions are extremely close to optimal for the majority of small-scale instances, (ii) reinforcement learning can adapt to a variety of cases, and {{\color{black}outperforms} state-of-the-art algorithms in the corresponding problem settings, (iii) the proposed approach can handle small-scale instances eﬀiciently, and (iv) the decomposition framework {{\color{black}is capable of solving large-scale problems} and the conversion algorithm is beneficial. 

{\color{black}
In reinforcement learning-related training, the neural networks contain an input layer, 6 to 10 hidden layers with 64 or 128 neurons (determined by the complexity), and an output layer. Our model is trained with the Adam optimizer using a learning rate of $10^{-4}$. The capacity $N$ of the replay memory $D$ is $10^6$ and we use mini-batches of $10^3$ sequences. As for the two coefficients related to exploration and exploitation, $\varepsilon$ decays uniformly from 1 to 0.1 and $\eta$ from 0.7 to 0. To prevent vanishing gradient, the decay coefficient $\gamma$ should be large enough, while it may cause exploding gradient. To ensure successful and stable training, we let $\gamma$ gradually increase from 0.8 to 0.98 during the training process.}
All reinforcement learning-related neural network training is performed on a single GPU (2080Ti), and other numerical experiments are carried out on a Windows Operating System with a 2.90-GHz CPU and 16 GB of RAM (IP models are solved by Gurobi 9.0.1). Detailed numerical experimental results can be viewed via the GitHub link \footnote{\url{https://github.com/hsinglukLiu/Puzzle-based_storage_appendix}}.

\subsection{Experiment setting} 

We randomly generate a series of benchmark instances for various problem settings, including single-item and multi-item retrieval problems. The detailed information of the benchmark instance set is summarized in Table \ref{longtable:instanceSummary}.

\renewcommand\arraystretch{1.05}
\begin{table}[h]
	\centering
	\caption{Comparision between RL and IP} 
	\label{longtable:instanceSummary}   
	\begin{footnotesize}
			\begin{tabular}{
					p{2.5cm}<{\centering}|
					p{1cm}<{\centering}
					p{1cm}<{\centering}
					p{1.5cm}<{\centering}
					p{3cm}<{\centering}|
					p{1.5cm}<{\centering}|
					p{3cm}<{\centering}
				}
				\hline 
		{\color{blue} Instance Series ID}  &
		{\color{blue} Grid size}  &
		{\color{blue} Desired items}  &
		{\color{blue} Escorts}  &
		{\color{blue} I/O points}  &
		{\color{blue} Instance Number}  &
		{\color{blue} Purpose}
		\\ \hline 
R422	& $4 \times 4$	& 2	&  2 & 2 ([[0, 0], [0, 3]])  & 1000 & \multirow{2}{*}{Compare IP and RL} \\
R622	& $6 \times 6$ & 2	&  2 & 2 ([[0, 0], [0, 5]]) 	& 1000 & \\
\hline  

F611 & $6 \times 6$		& 1	&  1 ([0, 0]) & 1 ([0, 0])& 35 & Compare with \cite{Gue2007PuzzleBased}\\
 \hline
F621 & $6 \times 6$	& 2&  1 ([0, 0]) & 2 ([[0, 0], [0, 1]])& 200 & Compare with \cite{Mirzaei2017modelling}\\
\hline 
{\color{black}R611} & {\color{black}$6 \times 6$}	& {\color{black}1}	&  {\color{black}1} & {\color{black}1 ([0, 0])} 	& {\color{black}1000} & {\color{black} \multirow{3}{*}{\begin{tabular}{p{3cm}<{\centering}} Compare with \cite{Yalcin2019} \end{tabular}} }\\
R612 & $6 \times 6$	& 1	&  2 & 1 ([0, 0]) 	& 1000 & \\
R613 & $6 \times 6$& 1	&  3 & 1 ([0, 0])	& 1000 & \\
\hline 
{\color{black}R621} & {\color{black}$6 \times 6$ }& {\color{black}2}	& {\color{black} 1} &{\color{black} 2 ([[0, 0], [0, 5]])}	& {\color{black}1000} & {\color{black}\multirow{3}{*}{\begin{tabular}{p{3cm}<{\centering}} Compare with \cite{zou2021heuristic} \end{tabular}}}\\
R622 & $6 \times 6$& 2	&  2 & 2 ([[0, 0], [0, 5]])	& 1000 & \\
{\color{black}R623} & {\color{black}$6 \times 6$}&{\color{black} 2}	& {\color{black} 3} & {\color{black}2 ([[0, 0], [0, 5]])}	& {\color{black} 1000} & \\
\hline 
R422	& $4 \times 4$	& 2	&  2 & 2 ([[0, 0], [0, 3]])  & 1000 & \multirow{5}{*}{\begin{tabular}{p{3cm}<{\centering}} {\color{black}Sensitivity analysis on the RL approach} \end{tabular}} \\
R522	& $5 \times 5$	&2	&  2 & 2 ([[0, 0], [0, 4]])   & 1000 & \\
R622	& $6 \times 6$ & 2	&  2 & 2 ([[0, 0], [0, 5]]) 	& 1000 & \\
R722	& $7 \times 7$ & 2	&  2 & 2 ([[0, 0], [0, 6]])	& 1000 & \\
R822	& $8 \times 8$ & 2	&  2 & 2 ([[0, 0], [0, 7]])	& 1000 & \\
 \hline
R-6$\times$37-1-22 & $6 \times 37$ & 1	&  22 & 1([0,18]) 	& 200 & \multirow{4}{*}{\begin{tabular}{p{3cm}<{\centering}} Test the performance of the decomposition framework \end{tabular}}\\ 
R-6$\times$37-13-22	& $6 \times 37$ & 13	&  22 & 13	& 100 & 
 \\ 
R-10$\times$61-1-61 & $10 \times 61$ & 1	&  61 & 1([0,30]) 	& 200 & \\
R-10$\times$61-21-61 & $10 \times 61$ & 21	&  61 & 21 	& 100 & \\ 
\hline 

Total 	& 	&   &  &	& 10835  & \\  \hline 			
		\multicolumn{7}{l}{$^1$Escorts: [0, 0] denotes that the escort location is fixed at location [0, 0]. If not particularly marked, {\color{black}the escorts} are placed randomly.}	  
				
			\end{tabular}
	\end{footnotesize}
\end{table}

\vspace{-0.3cm}

For readability, we generate instance series IDs for each dataset according to its characteristics.
We assume that small instances (where the grid size is no more than $8 \times 8$) have a square grid shape, i.e., $m=n$. Therefore, we refer to the grid size as $n$ in our instance series ID. Specifically, we label these instances as $\mathbb{P}nde$, where $\mathbb{P}$ has two optional values: R (escorts are randomly distributed) and F (only one escort is considered and placed at [0,0]), and $n,d,e$ are the grid size, the number of desired items and escorts respectively. For medium and large instances, we intuitively denote the instance ID as R-$m\times n$-$d$-$e$. {\color{black}For instances with $d > 1$, we simply assign the $i^{th}$desired item to the $i^{th}$ I/O point where $i=1,2,...d$.}
All {\color{black}10835} instances are classified into four series for different purposes. 
More specifically, the R422 and R622 series aim to verify the solution quality of the reinforcement learning approach by comparing its results with those of the IP formulation. Then, {\color{black}8 groups of instances} are used to compare the proposed method to existing algorithms. After that, experiments on R$n$22 ($n \in \{ 4, 5, 6, 7,8\}$) instance sets are conducted to demonstrate the effectiveness of the reinforcement learning approach on handling small-scale instances and conduct 
relevant sensitivity analysis.
Finally, we test the performance of the decomposition framework employing warehouses of size $6 \times 37$ and $10 \times 61$ as medium and large instances, respectively.

Based on our experiments, the reinforcement learning method performs well on R$n$11 (100\% optimal actually). However, it is difficult to ensure optimality when training complex scenarios. Occasionally, the model does not learn how to solve a complicated instance, i.e., the solution obtained is invalid and the number of actions may be unreasonably large. Therefore, we need to eliminate such outliers. The elimination rule is: since the minimum number of moves for R$n$11 is $8n-11$ (according to our experiments and proved by \cite{Gue2007PuzzleBased}), the upper bound for R$nde$ instances is $(8n-11)d$. If the reinforcement learning solutions exceed $(8n-11)d$, we think the method fails to solve the instances, otherwise, the instances are considered successful.
The elimination rule is also applicable to heuristic algorithms.

\subsection{Solution quality for small instances: comparison between reinforcement learning and integer programming}\label{sec: RL vs IP}


To validate the {\color{black}solution quality} of the reinforcement learning approach, we compare its results with the IP solutions. Since the IP model is difficult to solve as the problem size increases, we only test {\color{black} R422 and R622 instances}. To avoid wasting too much time on hard instances, we set the CPU time limit to {\color{black}14400} {\color{black}seconds (4h)}. The results are summarized in Table \ref{longtable:RLvsIPR622andR422}.

 \renewcommand\arraystretch{1}
\begin{table}[h]
	\centering
	\caption{Comparision between RL and IP} 
	\label{longtable:RLvsIPR622andR422}   
	\begin{footnotesize}
		{\color{black}
			\begin{tabular}{
					>{\arraybackslash}p{5.3cm} |
					>{\centering\arraybackslash}p{1.8cm} | 
					>{\centering\arraybackslash}p{1.8cm}
					>{\centering\arraybackslash}p{1.8cm}
					>{\centering\arraybackslash}p{1.8cm}
					>{\centering\arraybackslash}p{1.8cm}
				}
				\hline 
				Instance series & \textbf{R422} & \multicolumn{4}{c}{\textbf{R622}} \\
				\hline 
				\multirow{1}{*}{Total number of instances} & \multirow{1}{*}{1000} & \multicolumn{4}{c}{1000}     \\ \hline  
				RL Obj range  & [1, 30]	 & $[0, 20]$  & $[21, 29]$ & $[30, 32]$  & $[33, 74]$ \\ 
				Number of tested instances  & 1000 & 110  & 275 & 124 & 491 
				\\ 
				\hline \hline 
				%
				%
				\multirow{2}{*}{Average RL Obj}	 & \multirow{2}{*}{15.503}  & \multicolumn{4}{c}{31.615} \\ \cline{3-6}
				& 	 & 15.591 & 25.807 & 31.008 & 39.648 
				\\ 
				\hline 
				\multirow{2}{*}{Average RL CPU time (s)}	 & \multirow{2}{*}{0.014}  & \multicolumn{4}{c}{0.031} \\ \cline{3-6}
				& & 0.015 & 0.025  & 0.031 & 0.038 
				\\ 
				\hline \hline 
				\multirow{2}{*}{\makecell[l]{Average IP incumbent Obj}}	 & \multirow{2}{*}{15.461}  & \multicolumn{4}{c}{27.011*} \\ \cline{3-6}
				& & 15.582 & 25.712  & 30.860 & 34.946 
				\\ 
				\cline{2-6} 
				\multirow{2}{*}{Average IP best LB}	 & \multirow{2}{*}{15.461}  & \multicolumn{4}{c}{25.601} \\ \cline{3-6}
				& & 15.582 & 25.567  & 28.323 & 27.177 
				\\ 
				\cline{2-6} 
				\makecell[l]{Average IP incumbent Gap (\%)} & \textbf{0 \%}  & \textbf{0 \%} & \textbf{0.486\%}  & 7.781\% & 27.177\% 
				\\ 
				\cline{2-6} 
				Average IP CPU time (s) & 33.621 & 17.346 & 2116.403 & 10964.640 & 14232.863 
				\\ 
				\hline \hline
				Average RL Gap to IP incumbent (\%)	 & \textbf{0.271\%}	 & \textbf{0.061\%} & \textbf{0.320\%}  & \textbf{0.427\%}  & \textbf{1.072\%}  \\ 
				\cline{2-6}  
				Total Success (RL)	 & 992/1000 & 110/110 & 275/275  & 124/124  & 479/491  \\ 
				\cline{2-6}  
				Total Optimal (RL)	 & 933/1000 & 109/110 & 243/275  & 43/124  & 8/491  
				\\
				\hline \hline 
				Total Success (IP)	 &1000/1000 & 110/110 & 274/275  & 114/124  & 148/491  \\ 
				\cline{2-6}  
				Total Optimal (IP)	 &1000/1000 & 110/110 & 262/275  & 57/124  & 13/491  
				\\
				\hline \hline 
				\multicolumn{6}{l}{
					\makecell[l]{
						*: Note that only the instances with at least one found integer feasible solution are included when computing this \\metric. This is exactly why the average IP incumbent objective on R622 instance set (646 included instances) is\\ significantly smaller 
						than that of RL approach (i.e., 31.615, and 988 included instances). However, it does not mean\\ that RL solutions are far from the optimal solution. 
					}
				}
				\\
				\hline 
				
			\end{tabular}
		}
	\end{footnotesize}
\end{table}

The indices in Table \ref{longtable:RLvsIPR622andR422} are defined as follows:
\begin{itemize}
	\setlength{\itemsep}{0pt}
	\item 
	{\color{black}\textit{Average IP best LB}
		$ = \sum_{i=1}^{N}{\left( \text{IP best LB}_i \right)/N  
		}$}
	, where $i$ is the instance ID, $N$ is the number of {\color{black}tested} instances.
	{\color{black}\texttt{IP best LB} is the minimum lower bound reported by the solver within the CPU time limit (all tested instances are included).}
	
	\item {\color{black}\textit{Average IP incumbent Gap} } (\%) $ = 100 \times \sum_{i=1}^{N}{\left(
		\frac{\text{IP incumbent Obj}_i - {\color{black}\text{IP best LB}_i}}{\text{IP incumbent Obj}_i}\right)/N  
	}$ \%, representing the optimality gap of the best solution obtained by the solver. 
	
	\item {\color{black}\textit{Average RL Gap to IP incumbent}} (\%)
	$ = 100 \times \sum_{i=1}^{N}{\left(
		\frac{\text{RL Obj}_i - \text{IP incumbent Obj}_i}{{\color{black}\text{RL Obj}_i}}\right)/N   
	}$ \%, indicating the gap between the reinforcement learning solutions and the incumbents obtained by the solver. 
\end{itemize}

{\color{black}
	For the set of R422 instances, all 1000 
	instances are solved to optimality by the IP formulation. It is gratifying that the RL approach obtains optimal solution 
	on 933 instances, achieving a quite tight average optimality gap (only 0.271\% actually). 
	With respect to R622 instances, the overall results become more complicated because of the increased problem size. 
	To illustrate the comparison results more clearly, we categorize the instances into four groups by their RL objective values, which reflects the computational complexity of instances, as those with higher RL objective values are often more difficult to solve using the IP formulation, more specifically, [0, 20], [21, 29], [30, 32], [33, 74]. 
	Basically, the first two sets of instances are relatively easy to solve and the performance of the RL approach is well verified according to the reported results. Specifically, for the first group of instances (110 in total), the IP model obtains optimal solutions on all of them. 
	Additionally, the RL approach achieves 
	an impressive performance on these instances, where 109 out of 
	110 instances' solutions are proved to be optimal as well as 
	the corresponding overall average 
	optimality gap hits 0.061\%. And then, for the second set of instances, the RL approach solves 243 out of 275 instances to optimality, with an average gap of 0.320\%. 
	However, as the RL objective value rises, the computational burden of the IP model increases dramatically, resulting in more instances where no feasible solution can be found within the time limit. 
	To be more specific, 10 and 343 instances fail to yield available feasible solutions by solving the IP model for the third and fourth groups of instances respectively. As a result, the optimality gap of RL solutions cannot be effectively evaluated on the aforementioned instances. 
	Fortunately, the 
	results indicate that these two sets of instances' RL solutions are very near to their IP solutions, 
	with corresponding overall average gaps of 0.427\% and 1.072\%. 
	It is also worth noting that the needed computational time of the trained RL models for all tested instances is extremely short (in milliseconds). 
	In contrast, the IP model is usually highly time-consuming, especially for those hard-to-solve instances.
	Based on the above comparisons, the reinforcement learning approach seems to be powerful on small-scale instances, both in solution quality and computational time.
	
	We can also look into the failed instances by the RL method and figure out how to tackle them. It is easy to find that instances with larger RL objectives are more likely to fail. These instances require more moves to finish the retrieval tasks, i.e., more rounds of action-taking decisions. In each round, the RL method has the possibility to give an inappropriate action due to the noise of the training process. Consequently, the more rounds there are, the more likely the task will fail. In practice, we further suggest two options to solve the failed instances. First, notice that RL fails because the route length surpasses the given limit; for example, the item is tracked and moving in a local area. Therefore, we can try to find out such one or a few ``weak'' points and randomly or heuristically select a move other than RL's suggestion. We leave this for future work. Second, if, unfortunately, the first one fails again, we can adopt any algorithm which supports single-item multi-escort retrieval in PBS, such as \cite{Yalcin2019}, to route the items separately, and then use our conversion method to transfer the solution to a simultaneous movement plan.
}

\begin{scriptsize} 
	\begin{longtable}{p{2.5cm}<{\centering}p{3cm}<{\centering}|p{1.5cm}<{\centering}p{2.5cm}<{\centering}p{2cm}<{\centering}|p{1.5cm}<{\centering}}
		\caption{Comparision of RL and other algorithms}   
		\label{longtable:RLvsHeuristic}    \\  
		
		\hline 
		{\color{blue} Instance series ID}  &
		{\color{blue} Methodology}  &
		{\color{blue} Ave. Obj}  &
		{\color{blue} Ave. CPU time (s)}  &
		{\color{blue} Total Success} &
		{\color{blue} Diff (\%)
		}  
		\\
		\hline
		\endfirsthead
		
		\hline 
		{\color{blue} Instance series ID}  &
		{\color{blue} Methodology}  &
		{\color{blue} Ave. obj}  &
		{\color{blue} Ave. CPU time (s)}  &
		{\color{blue} Total Success} &
		{\color{blue} Diff (\%)
		}    \\
		\hline
		\endhead
		
		\endfoot 	
		
		\multirow{2}{*}{F611} &RL  & 	19.857	  &0.014& 	35/35& \multirow{2}{*}{0} \\
		\cline{2-5}  
		&\cite{Gue2007PuzzleBased}  & 	19.857	 &0.000004 & 35/35&  \\							
		\hline

		\multirow{2}{*}{F621} &RL  & 	32.462  &0.028& 	199/200& \multirow{2}{*}{$-19.278$} \\
		\cline{2-5}  
		&\cite{Mirzaei2017modelling}  & 	38.720 &0.0000052 & 200/200&  \\							
		\hline

		\multirow{2}{*}{\color{black}R611} & \color{black}RL  & 	\color{black}19.369	 & \color{black} 0.024& \color{black} 	1000/1000& \multirow{2}{*}{\color{black}$-0.005$} \\
		\cline{2-5}  
		&\cite{Yalcin2019}  & \color{black}	19.463	 & \color{black} 0.00032  & \color{black} 1000/1000&  \\
		\cline{2-6}  

		\multirow{2}{*}{R612} &RL  & 	17.505	  &0.015& 	999/1000& \multirow{2}{*}{$-3.959$} \\
		\cline{2-5}  
		&\cite{Yalcin2019}  & 	18.198	 &0.00099  & 1000/1000&  \\
		\cline{2-6}  							

		\multirow{2}{*}{R613} &RL  & 	16.931  &0.015& 	995/1000& \multirow{2}{*}{$-3.562$} \\
		\cline{2-5}  
		&\cite{Yalcin2019}  & 	17.534	 &0.00123  & 1000/1000&  \\							
		\hline

		\multirow{2}{*}{\color{black}R621} &\color{black}RL  & 	\color{black}34.72	  &\color{black}0.030& 	\color{black}1000/1000& \multirow{2}{*}{\color{black}$-21.483$} \\
		\cline{2-5}  
		&\cite{zou2021heuristic}  & \color{black}	42.179	  &\color{black}0.346 & \color{black}900/1000&  \\
		\cline{2-6} 							

		\multirow{2}{*}{R622} &RL  & 	31.615	  &0.028& 	988/1000& \multirow{2}{*}{$-6.864$} \\
		\cline{2-5}  
		&\cite{zou2021heuristic}  & 	33.785	  &0.346 & 948/1000&  \\							
		\cline{2-6}

		\multirow{2}{*}{\color{black}R623} &\color{black}RL  & \color{black}	29.545	  &\color{black}0.028& 	\color{black}974/1000& \multirow{2}{*}{\color{black}$-4.813$} \\
		\cline{2-5}  
		&\cite{zou2021heuristic}  & \color{black}	30.967	  &\color{black}0.556 & \color{black}976/1000&  \\							
		\hline
		\multicolumn{6}{l}{$^1$Diff (\%): $\frac{100(\text{RL Obj}-\text{Other Obj})}{\text{RL Obj}}$}	 

	\end{longtable}
\end{scriptsize} 

\subsection{Extension to several reduction versions}\label{sec: RL vs Other}

Since existing researches rarely consider multiple desired items and multiple escorts at the same time, it is difficult for us to verify the effectiveness of our method. However, by slightly changing the parameters, reinforcement learning can solve the reduction versions, i.e., \textit{single-item retrieval with a single escort}, \textit{single-item retrieval with multiple escorts}, \textit{multi-item retrieval with a single escort}. We conducted several numerical experiments to show that reinforcement learning outperforms state-of-the-art algorithms. The results are summarized in Table \ref{longtable:RLvsHeuristic} and the details of the experimental design and analysis of the results are described in the following sections.

\subsubsection{Comparison between \cite{Gue2007PuzzleBased} and reinforcement learning}\label{sec:Gue}

For F$n$11, \cite{Gue2007PuzzleBased} presented a closed-form formula (optimal) to indicate the minimum number of moves. To retrieve an item located at $(i, j)$, the needed number of moves is
\begin{align}
&6i+2j-13, &&\text{ if } i>j, \label{eq:Gue1} \\
&6j+2i-13, &&\text{ if } j>i, \label{eq:Gue2} \\
&8i-11, &&\text{ if } i=j. \label{eq:Gue3}
\end{align}

This work starts from 0 instead of 1 (\cite{Gue2007PuzzleBased}) to index the grid cells. To be uniform, we must execute a transfer step before running. More specifically, assuming that the coordinates are ($i',j'$) in our generated instances, let $i=i'+1,\ j=j'+1$ to use the formula. As the escort must initially locate at the I/O point in \cite{Gue2007PuzzleBased}, the number of available instances is limited (only 35 instances in total). We compare RL with \cite{Gue2007PuzzleBased} by testing F611 instances.  

For F611, all instances are solved to optimality by the RL approach. As for the CPU time, the RL approach terminates within 0.1 seconds. Since the closed-form formula can offer the number of moves without considering state transitions, its CPU time is extremely short.

\subsubsection{Comparison between \cite{Mirzaei2017modelling} and reinforcement learning}

Another variant is F$n21$, where the initial position of the single escort is fixed at [0,0]. \cite{Mirzaei2017modelling} proposed a heuristic for F$n21$, which we found does not fully consider the boundary conditions. We refined Theorem 1 of \cite{Mirzaei2017modelling} by filling in the missing cases: the minimum number of moves to retrieve two horizontally aligned adjacent desired items in a puzzle system is
\begin{align}
&3i+7j-11,    &&i<j \text{ and }  i \text{ even, }  \label{eq:Mir1} \\ 
&10i-4,   &&j-i=1  \text{ and }    i \text{ odd, } \label{eq:Mir2}\\
&3i+7j-13,  &&j-i \geqslant 2  \text{ and }    i \text{ odd, }  \label{eq:Mir3}\\
&10i-7, &&i=j  \text{ and }    i \text{ even, }  \label{eq:Mir4}\\
&10i-9, &&i=j  \text{ and }    i \text{ odd, }  \label{eq:Mir5}\\
&7i+3j-7, &&i>j  \text{ and }    j \text{ even, } \label{eq:Mir6}\\
&7i+3j-9,  &&i>j  \text{ and }    j \text{ odd. }  \label{eq:Mir7}
\end{align} 
where $(i, j)$ is the position of the desired item closer to the I/O point and the escort is positioned to the right of the two desired items. Note that the two desired items must be gathered together before applying the formula.

Similar to the procedure in Section \ref{sec:Gue}, the index transformation is required before running the algorithm, which we will not describe here. To make the setup of this paper consistent with that of \cite{Mirzaei2017modelling}, we specify that the retrieval task is completed when the two desired items are moved to the positions [0,0] and [0,1], respectively. We ignore the final move (i.e., move the second desired item from [0,1] to [0,0]) in \cite{Mirzaei2017modelling}. Hence, the result offered by the revised heuristic method is one less than that calculated by the original version. In addition, the initial position of the escort is fixed at [0,0]. These two settings result in fewer instances (200 instances are selected).

The results indicate that RL outperforms the heuristic. RL obtained valid solutions in 99.5 percent of all instances. Moreover, RL provides a 19.28\% reduction in the average number of moves compared to the heuristic algorithm. The main reason is that gathering the two desired items requires too many moves and, therefore, impairs the heuristic method's performance. The CPU time is not comparable because the heuristic results are calculated by directly employing formulas.

\subsubsection{Comparison between \cite{Yalcin2019} and reinforcement learning}

A more complex variant of F$n$11 is to consider multiple randomly placed escorts, which we call R$n1e$. The reinforcement learning method is capable of dealing with R$n1e$ naturally. Based on the $A^*$ algorithm, \cite{Yalcin2019} proposed a heuristic for R$n1e$. For each move of the desired item, an escort is selected according to a certain rule. The chosen escort helps the desired item to achieve the movement. The process repeats until the retrieval operation is completed. There are three specific search guiding estimates: manhattan distance, occupation estimate, and escorts estimate, and we {\color{black}implement their heuristic by using} the first one. We compare RL with \cite{Yalcin2019} by conducting {\color{black}R611,} R612, and R613 instances.

{\color{black}For R611, RL can successfully solve all 1000 instances, and the difference with the heuristic of \cite{Yalcin2019} is neglectable (0.005\%) in terms of the objective value.} For R612 and R613, 99.9\% and 99.5\% instances are successfully solved by RL respectively. The average number of moves provided by RL {\color{black}is approximately 4.0\% and } 3.5\% less than the results of the heuristic. {\color{black}The CPU time of the heuristic is very short for both RL and the heuristic}.

\subsubsection{Comparison between \cite{zou2021heuristic} and reinforcement learning}

By designing a series of estimation rules to evaluate the score of each action under the state, \cite{zou2021heuristic} proposed a heuristic algorithm for R$nde$, which has the identical problem setting of reinforcement learning. However, for a new instance, the heuristic algorithm must be re-executed, whereas reinforcement learning can solve it efficiently using the experience stored in the neural network. Not to mention that the rules in \cite{zou2021heuristic} have some incompleteness, leading to a lower success rate. {\color{black}R621, }R622{\color{black}, and R623} instances are used to compare RL with \cite{zou2021heuristic}.

For the results of {\color{black}all the tested instances}, it is obvious that RL outperformed the heuristic method in all three aspects: average number of moves, CPU time, and success rate. {\color{black}We can also observe that the performance advantage of RL over \cite{zou2021heuristic} diminishes as escort number increases; for example, a large difference (21.483\%) can be observed when escort number is equal to one, but the difference decreases to 6.864\% and 4.813\% when escort number is 2 and 3, respectively. Future efforts can be devoted to further increasing RL's ability to handle instances with more escorts, e.g., increasing the training time when there are more escorts.}

\vspace{-0.3cm}
\begin{figure}[H]
\centering
\hspace{-7mm}
	\subfigure[R422 instance set]{
	\label{fig:R422}
\includegraphics[width=0.3\textwidth,height=0.2\textheight]{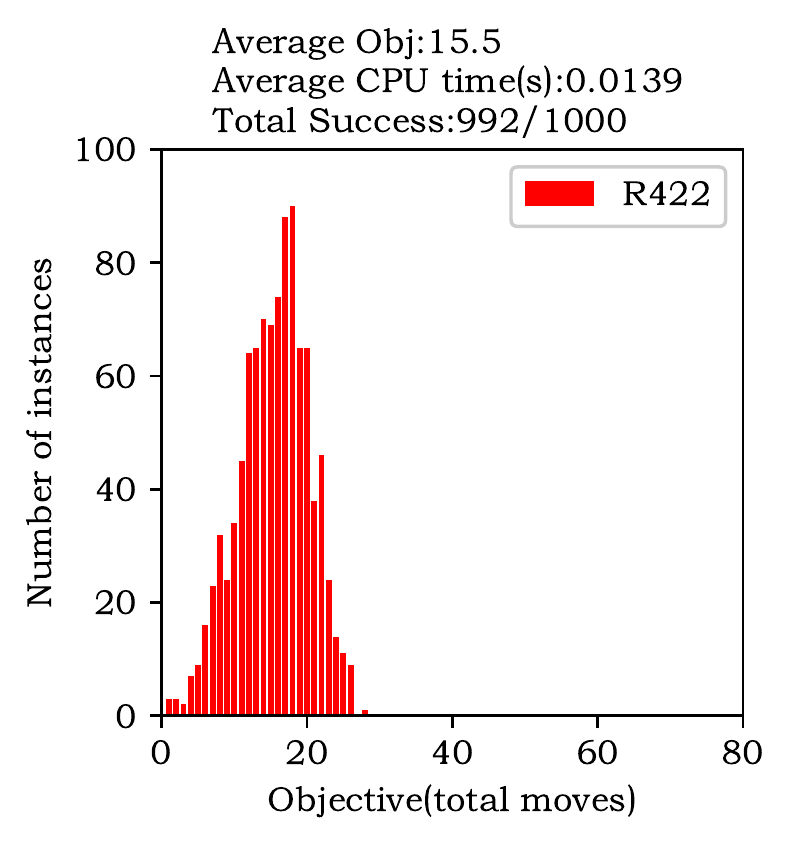}
	}
	\subfigure[R522 instance set]{
	\label{fig:R522}
\includegraphics[width=0.3\textwidth,height=0.2\textheight]{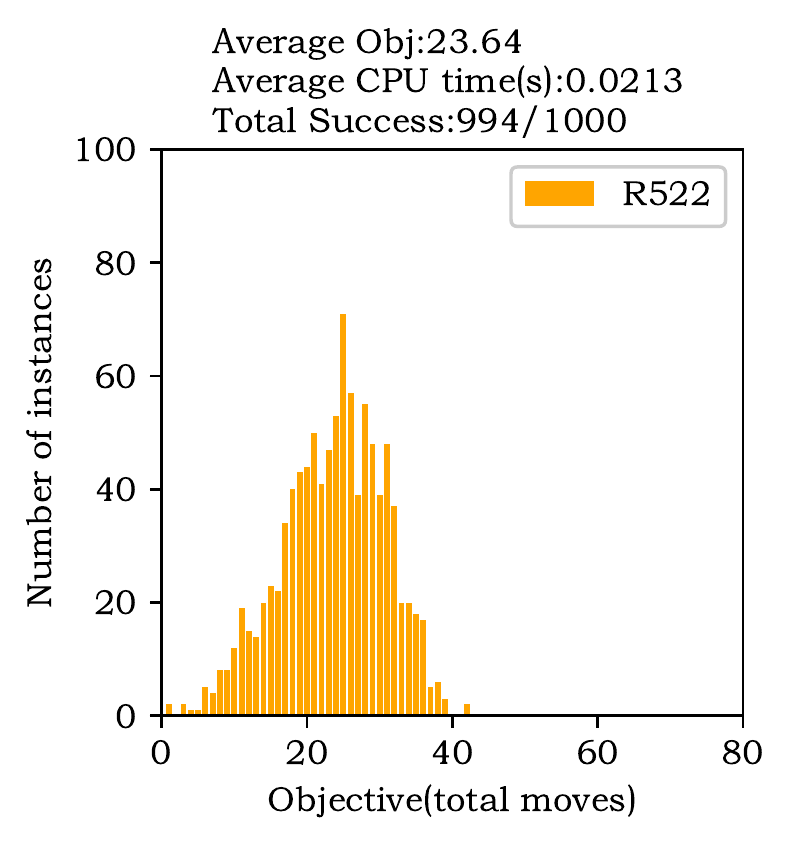}
	}
	\subfigure[R622 instance set]{
	\label{fig:R622}
\includegraphics[width=0.3\textwidth,height=0.2\textheight]{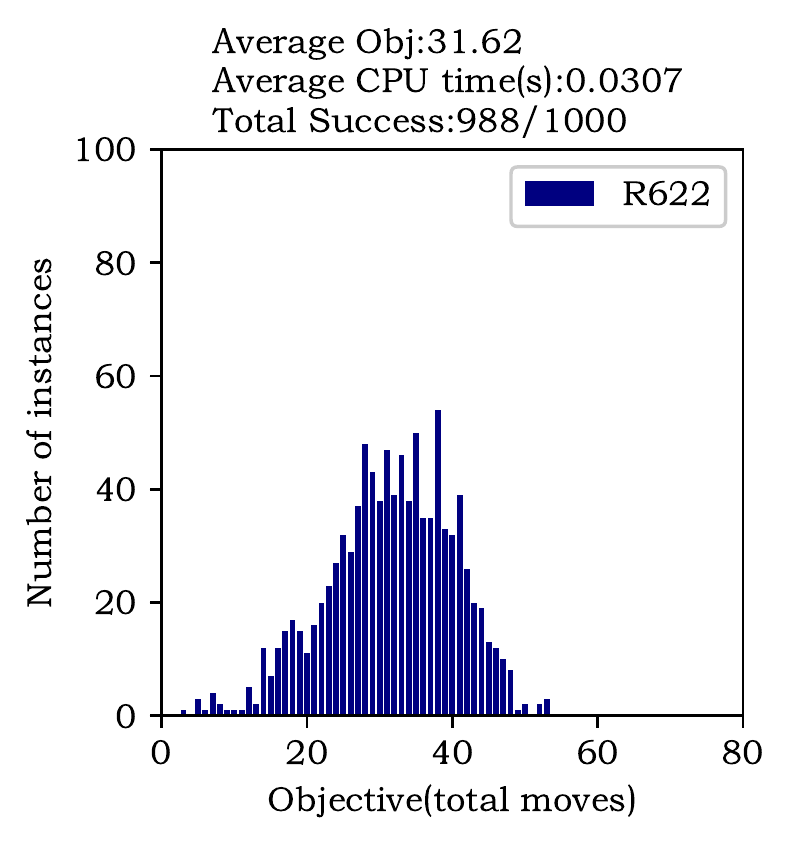}
	}
	\subfigure[R722 instance set]{
	\label{fig:R722}
\includegraphics[width=0.3\textwidth,height=0.2\textheight]{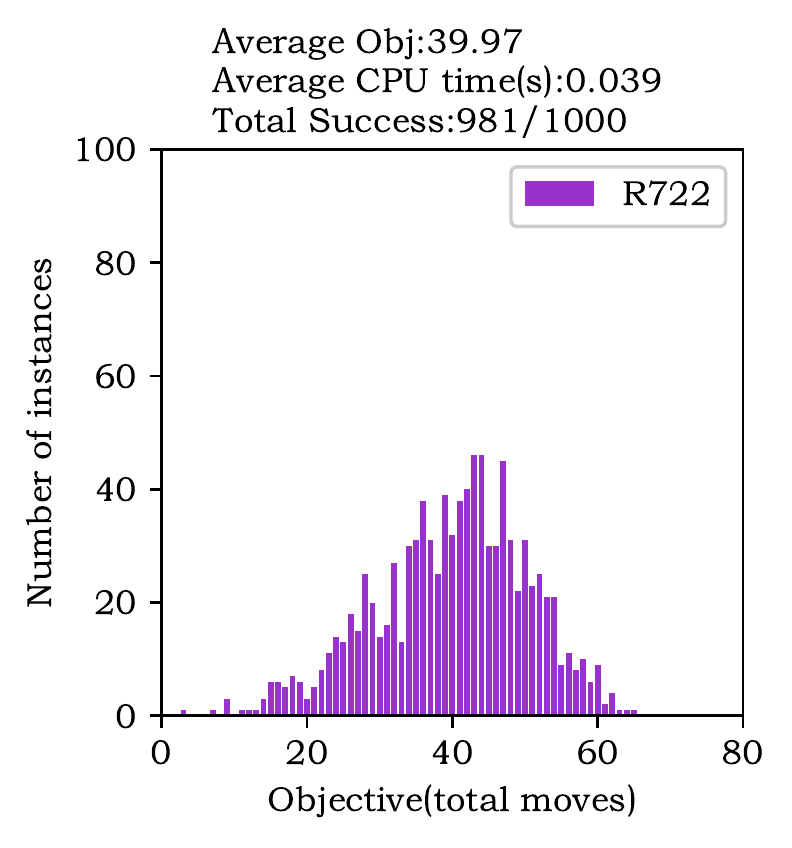}
	}
	\subfigure[R822 instance set]{
	\label{fig:R822}
\includegraphics[width=0.3\textwidth,height=0.2\textheight]{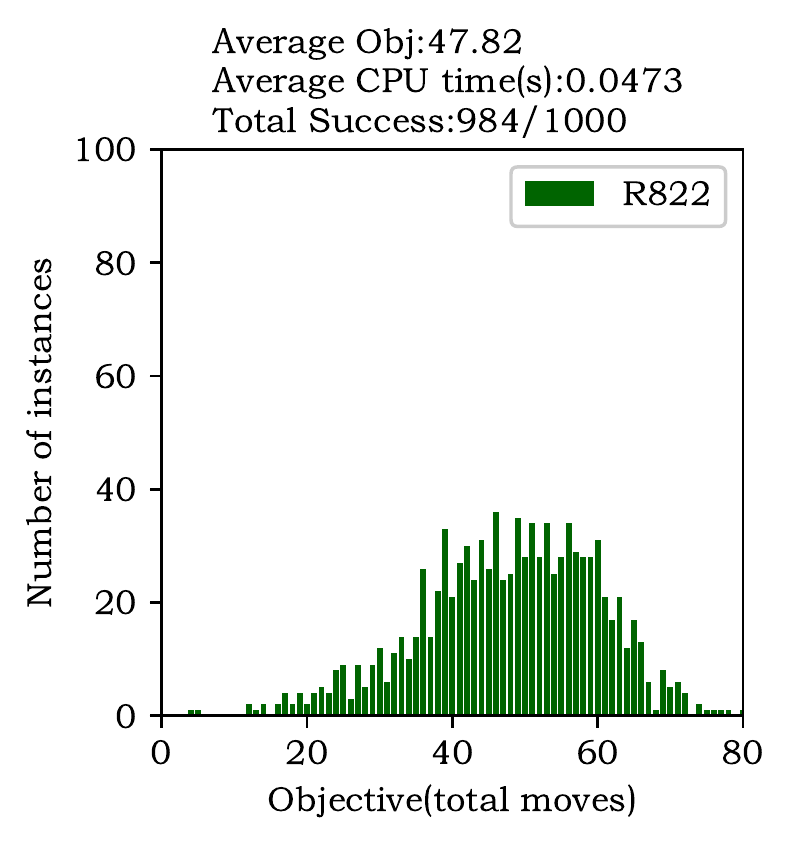}
	}
	\caption{\color{black}Results of RL on instances with different grid size}
	\label{fig:RLresults}
\end{figure}

\subsection{\color{black}Sensitivity analysis on the reinforcement learning approach}\label{sec: RL self}
\label{sec:regressionModel}
Occasionally, managers need to estimate the workload of different locations to schedule goods.
That is, given the locations of the desired items 
{\color{black} and the number of available escorts}, predict the number of moves required to complete the task in a given grid. Therefore, we attempt to analyze the relationship between $\sum \,\,d_{item\_manhattan}$ {\color{black}(defined in Section \ref{sec:usefulPoint}), $e$ (the number of escorts), and the average number of needed moves under various gird sizes. }

{\color{black}We first take experiments on R$n$22 ($n \in \{ 4, 5, 6, 7,8\}$) instances to observe the effects of grid size.} As shown in Figure \ref{fig:RLresults}, a moderate amount of movement is required in most cases for a given grid, with a low incidence of too little or too much movement.
Moreover, the average number of moves seems to show a linear growth as the grid size increases. The CPU time depends on the average number of moves, which represents the number of state transitions and neural network calls. Despite this, the CPU time is still short enough ($<0.1$ sec).

\begin{figure}[H]
\centering
\hspace{-7mm}
	\subfigure[The influence of grid size]{
	\label{fig:fig_422-822}
\includegraphics[width=0.45\textwidth]{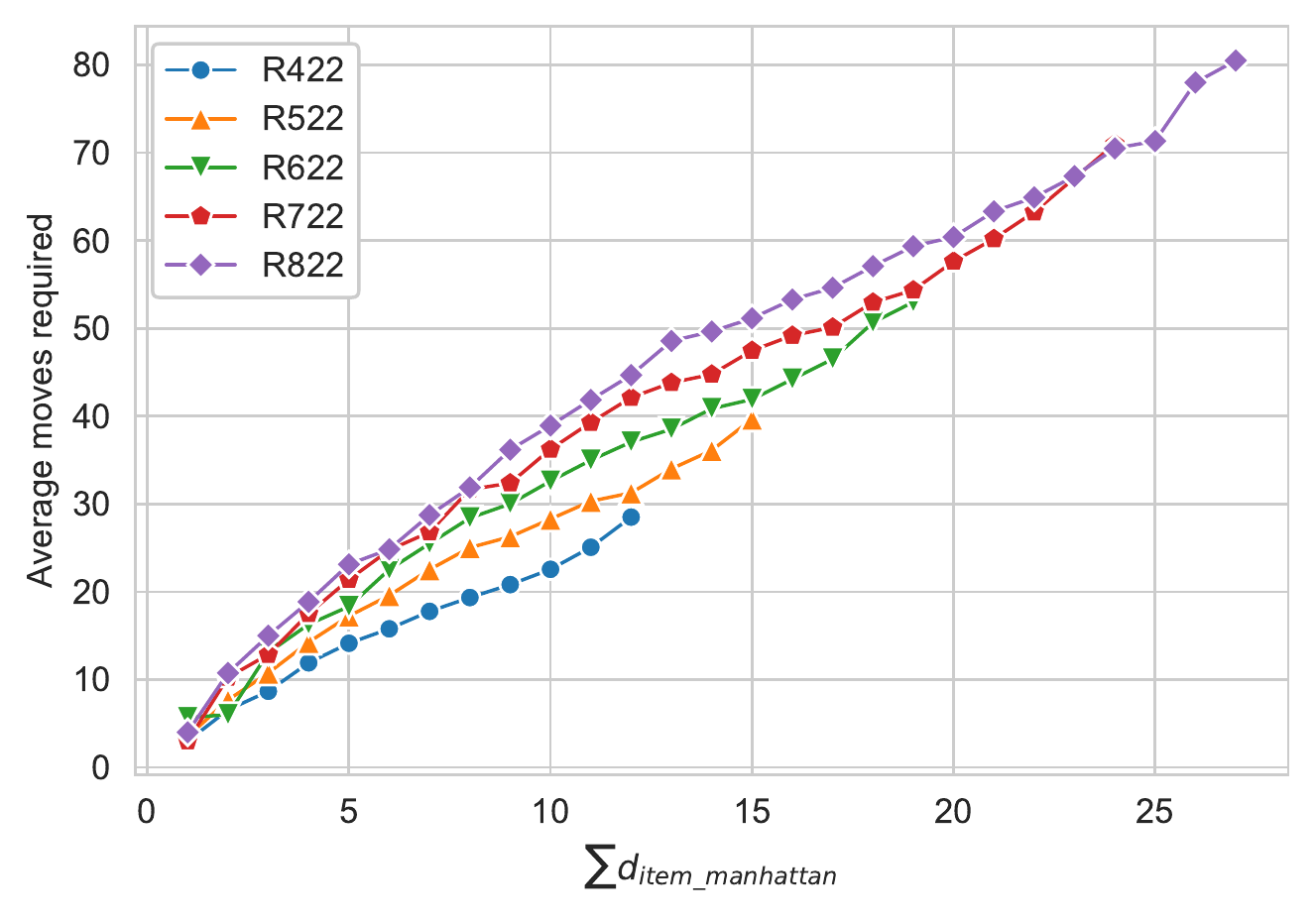}
	}
	\subfigure[The influence of escort number]{
	\label{fig:fig-62-123}
\includegraphics[width=0.45\textwidth]{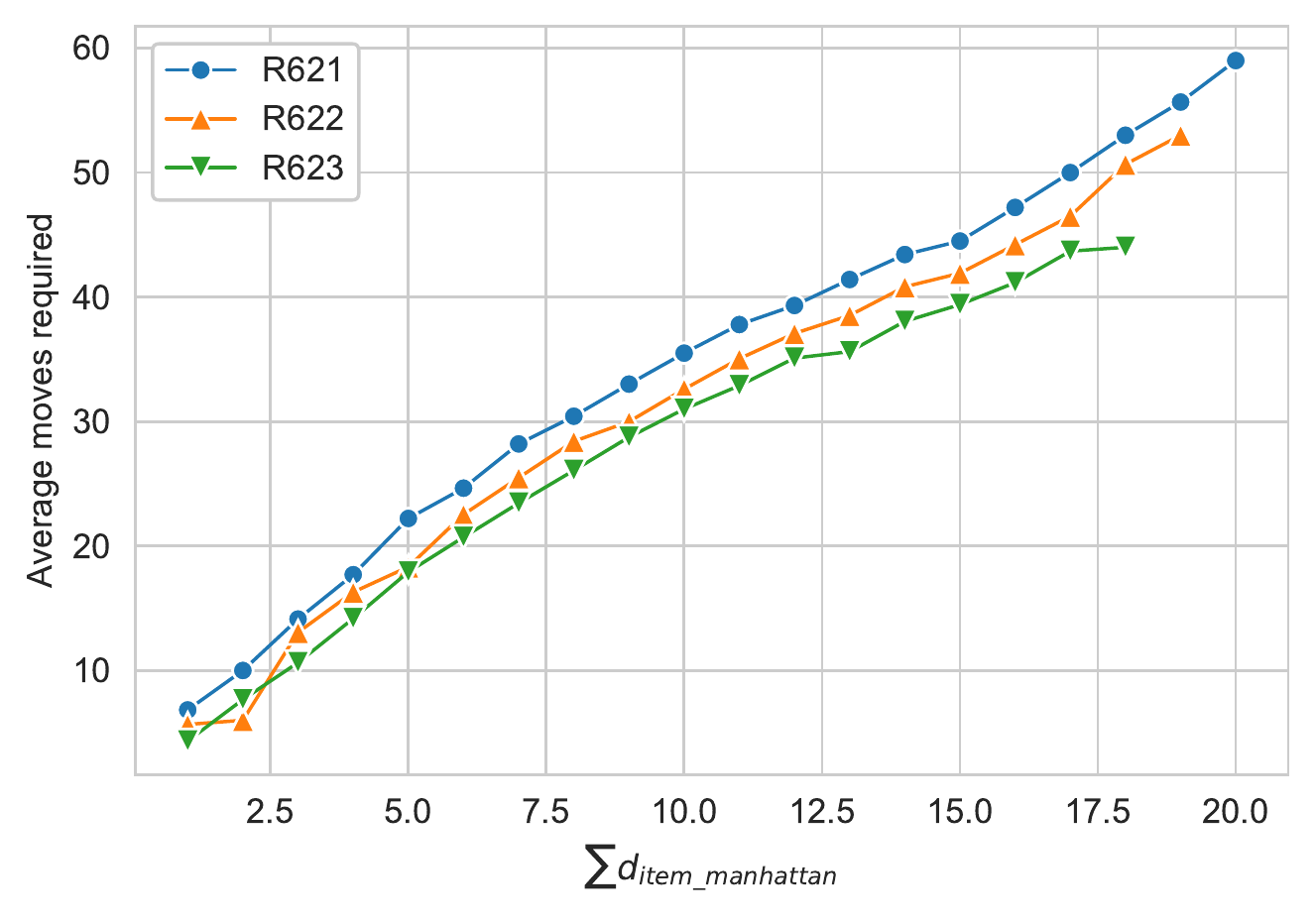}
	}
	\caption{The average number of moves with given desired items and grid size}
	\label{fig:regressionModel}
\end{figure}
 
Figure \ref{fig:regressionModel}\subref{fig:fig_422-822} {\color{black} presents the statistical results over various 
instance sets, which concerns} only the successful instances.
Notably, {\color{black}for any tested instance set,} the average minimum number of moves shows a strong linear relationship with the $\sum \,\,d_{item\_manhattan}$, which makes intuitive sense. Additionally, larger-scale instances {\color{black} often} require more moves on average with the same $\sum \,\,d_{item\_manhattan}$ {\color{black}due to the increase in the average distance between the desired items and the escorts. Based on this observation, we introduce the grid size $n$ as a variable as well. To take sensitivity analysis on the number of escorts, we use the experimental results on R621, R622, and R623 as summarized in Table \ref{longtable:RLvsHeuristic}. As Figure \ref{fig:regressionModel}\subref{fig:fig-62-123} shows, the number of moves can be saved when given more escorts; however, the marginal saving decreases.}

{\color{black}With these observations, we propose the following empirical prediction model for the average minimum number of moves $d_{pred}$
\begin{align}
d_{pred}= \alpha_0 +\alpha_1 \sum \,\,d_{item\_manhattan}+\alpha_2 \ln(e) + \alpha_3 n. 
\label{eq:pred}
\end{align}
Note that we use a nonlinear ($\ln$) term for escort number $e$ due to its dimishing marginal saving nature. Here $\alpha_0, \alpha_1, \alpha_2, \alpha_3$ are the coefficients obtained by regression. We take data from R422 $\sim$ R822 and R621 $\sim$ R623 for regression, and the results are
$\alpha_0=-9.517, \alpha_1=2.351, \alpha_2= -4.142, \alpha_3 = 3.445$.
The R-square $R^2=0.916$ confirms a good fitting. The first term $\alpha_0$ is a constant term. The second term, $\alpha_1 \sum \,\,d_{item\_manhattan}$, provides an approximated number of needed moves to deliver the desired items from their dwell positions to the corresponding I/O points. Then, $\alpha_2 \ln(e)$ ($\alpha_2 < 0$) can be interpreted as the number of saved moves due to the introduction of escorts. The last term, $\alpha_3 n$, reflects the effect of grid size.  
Equation (\ref{eq:pred}) can also be used to obtain a tight $K$ for the IP model: first estimate an average number of moves, that is, $d_{pred}$, then we can gradually increase the value if the solver 
returns an infeasible message.} 

\subsection{Performance of decomposition framework (conversion algorithm) on medium and large instances}

To demonstrate the effectiveness of the decomposition framework, we execute the approach on medium- and large-scale cases. {\color{black}We mainly use the RL method to solve the sub-PBS problems; however, for the IP-based decomposition, the computational time grows significantly as the number of iterations increases, so we only tried IP on R-6×37-1-22 instances to further evaluate the performance of our method.} Here, the escort number is set to 10\% of the total number of cells, as the PBS system is most applicable when the required density exceeds about 90\% (\cite{Gue2007PuzzleBased}). 
The only comparable approach is \cite{Yalcin2019}, {\color{black}whose method solves single-item retrieval problems with multiple escorts in a heuristic manner that can be applied directly to large-scale cases.} The other three highly relevant pieces of literature are not quite suitable. 
\cite{Gue2007PuzzleBased} aims to provide exact solutions for single-item instances with a single escort, and \cite{Mirzaei2017modelling} focuses on two-load and three-load with a single escort. These two settings may be less practical in medium- and large-scale warehouse environments. 
For \cite{zou2021heuristic}, their algorithm can not provide a feasible solution within 0.5h, which is unacceptable for a heuristic method, so we do not conduct comparative experiments with them. The results are summarized in Table \ref{table:largescale}. {\color{black}Note that due to the randomness in the iterative process of the decomposition algorithm, we take the best value among the results of three runs. And multiple runs are acceptable because reinforcement learning takes a very short time.}

\begin{scriptsize} 
	\begin{longtable}{p{3cm}@{}ccccc@{}}
		\caption{Results of decomposition framework on medium- and large-scale instances}  
		\label{table:largescale} 
		  \\  

		\hline 
		{\color{blue} Instance series ID}  &
		{\color{blue} Methodology}  &
		{\color{blue} Total success}  &
		{\color{blue} Ave. Obj (lm)}  &
		{\color{blue} Ave. Obj (sm)} &
		{\color{blue} Average CPU time (s)
		}  
		\\
		\hline
		\endfirsthead
		
		\hline 
		{\color{blue} Instance series ID}  &
		{\color{blue} Methodology}  &
		{\color{blue} Total success}  &
		{\color{blue} Ave. Obj (lm)}  &
		{\color{blue} Ave. Obj (sm)} &
		{\color{blue} Average CPU time (s)
		}      \\
		\hline
		\endhead
		
		\endfoot 	
		
		\multirow{3}{*}{R-6$\times$37-1-22}
		& {\color{black}Decompositio(IP) }  & {\color{black}200/200}                        & {\color{black}40.39 }            & -           & {\color{black}59.95 }
		\\
		& {\color{black}Decompositio(RL)}   & {\color{black}199/200}               & {\color{black}40.55}            & {\color{black}\textbf{30.45} }          & {\color{black}0.084   }
		\\
		&\cite{Yalcin2019}                 & 200/200                       & 40.60               & -                       & 0.004       
		\\ \cline{2-6}  	
		\multirow{2}{*}{R-10$\times$61-1-61}
		    & Decomposition(RL)   & 200/200                 & 58.48               & \textbf{45.45}           & 0.105       

		\\
		 
		 & \cite{Yalcin2019}                 & 200/200                    & 58.18               & -                        & 0.056     
		\\ \hline 
		R-6$\times$37-13-22 & Decomposition(RL) & 100/100                         & 255.54              & \textbf{64.81}           & 0.437      \\ \cline{2-6}  
		R-10$\times$61-21-61 & Decomposition(RL) & 100/100                 & 631.71              & \textbf{91.79}           & 1.250       \\							
		\hline
		\multicolumn{6}{l}{Ave. Obj (lm): the average number of moves over tested instances under single-load movement assumption}	 
        \\
		\multicolumn{6}{l}{Ave. Obj (sm): the average retrieval time over tested instances under simultaneous movement assumption}

	\end{longtable}
\end{scriptsize} 

According to the results, our approach is comparable with that of \cite{Yalcin2019} for the single-item retrieval scheme. {\color{black}For example, the IP-based decomposition, the RL-based decomposition, and \cite{Yalcin2019} require 40.39, 40.55, and 40.60 moves on average for R-6×37-1-22, respectively, with very subtle differences. Meanwhile, the IP-based decomposition algorithm performs slightly better than the RL-based decomposition one, which is intuitive because reinforcement learning is not guaranteed to yield the optimal solutions. In the large-scale setting, 58.48 and 58.18 moves are needed on average according to the RL-based decomposition and \cite{Yalcin2019}.} 
Moreover, retrieval efficiency improves significantly when simultaneous movements are taken into account. For medium-scale instances, the average retrieval time under the single-load movement assumption is 3.94 ($=255.54/64.81$) {{\color{black}times longer than} that under the simultaneous movement setting. And the figure for large-scale cases is 6.88 ($=631.71/91.79$). Additionally, we present the results for medium- and large-scale multi-item retrieval cases. 
In conclusion, the decomposition framework performs comparably to state-of-the-art algorithms in single-item retrieval cases and achieves high efficiency in multi-item retrieval problems.

\section{Conclusion and future research}

In this work, we address the multi-item retrieval problem in the PBS system with general settings. Under the single-load movement assumption, a RL algorithm (double \& dueling DQN) is developed to provide a near-optimal retrieval path.
A semi-random and semi-guided action selection mechanism is proposed to balance exploration and exploitation and accelerate convergence.
{\color{black}Furthermore, we develop a general compact IP model for the studied problem that achieves a considerable reduction in the number of decision variables and constraints when compared to the literature. 
It is then utilized to evaluate the solution quality of the RL approach.} Following that, a conversion algorithm is proposed to enable the obtained solutions to handle simultaneous movement considerations.  
Finally, to solve large-scale instances, we design a decomposition framework, in which each iteration extracts a sub-area and then solves the subproblem using the RL method. 

We generate {\color{black}10835} instances in the absence of a comparable benchmark, serving as a benchmark dataset for this and future relevant research. According to the experimental results presented in Section \ref{sec: RL vs IP}, the solutions  yielded by the RL method are very close to optimal. For simple instances (R422), 
{\color{black}
RL obtains the optimal solution over 93.3\% of tested instances, and achieves 
a quite tight average optimality gap of 0.271\% over full dataset. 
For more difficult instances (R622), the results reveal that the solutions of the RL approach are very close to those of the IP model as well, and the average gap over four groups of instances are 
0.061\%, 0.320\%, 0.427\% and 1.072\%, respectively.} 
Moreover, the results reported in Section \ref{sec: RL vs Other} indicate that the RL method always outperforms the three highly relevant heuristic algorithms after extending the proposed approach to the corresponding problem variants, not to mention that the RL method can offer specific retrieval paths within in a very short computational time.
{\color{black}Then, we conduct the sensitivity analysis based on the results of small-scale benchmark instances and propose a regression model that can be adopted to approximate the minimum number of moves required for a given condition.}
In addition, the decomposition framework performs well in terms of solution quality and processing time for both medium- and large-scale instances. This implies that our approaches have a great potential 
{\color{black}to be applied }
in real manufacturing environments and (semi-)automated warehouses, providing high-quality and executable retrieval solutions in real-time for companies.

The main limitation of our method is that the optimality gap for medium- and large-scale instances are not provided, owing to the high complexity of the IP model. 
{\color{black}In addition, due to the way the state is designed, the neural networks need to be trained separately for problems with different settings. And because of the value-based property of DQN, the dimension of the action depends on the number of escorts.
Therefore, there are several possible directions for future research.} First, effective algorithms can be developed to accelerate the exact model and to offer tight bounds. {\color{black}Second, apply transfer learning policies to help the input representation be independent of the numbers of desired items and escorts, or even the grid size. Then, policy-based RL algorithms can be considered to weaken the relationship between the action and the number of escorts.} Moreover, retrieval problems in three-dimensional warehouses would be a potential challenge that remains to be investigated.

\section*{Acknowledgments}

This work is supported by the National Natural Science Foundation of China (Grant No. 71772100 and 71971127), and the Guangdong Pearl River Plan (2019QN01X890).

	\bibliographystyle{model5-names} 
	\biboptions{authoryear}
	\bibliography{reference.bib}
	
	

	
	%
\end{document}